\documentclass[lettersize,journal]{IEEEtran}
\usepackage{amsmath,amsfonts,algorithmic,array,textcomp,stfloats,url,verbatim,graphicx,color,multirow,hhline}
\usepackage[noadjust]{cite}
\usepackage{xspace}
\usepackage{comment}
\usepackage{hyperref}
\usepackage{caption}
\usepackage{graphicx}
\usepackage{booktabs}
\usepackage[table,xcdraw]{xcolor}
\usepackage{tikz}
\usepackage{flushend}
\usepackage{tabularx}

\newcommand{\myparagraph}[1]{\hspace{\fill}\linebreak\noindent\textbf{#1}}

\newcommand{\ITLogisticShort}{ITS\&L\xspace}
\def\tcheck{\tikz\fill[scale=0.4](0,.35) -- (.25,0) -- (1,.7) -- (.25,.15) -- cycle;} 

\newcommand{\dataops}{\textit{DataOps}\xspace}
\newcommand{\mlops}{\textit{MLOps}\xspace}
\newcommand{\llmops}{\textit{LLMOps}\xspace}
\newcommand{\agentops}{\textit{AgentOps}\xspace}

\newcommand{\TAI}{\textit{TAI}\xspace}
\newcommand{\fmops}{\textit{FMOps}\xspace}

\let\orgautoref\autoref

\renewcommand{\autoref}
        {\def\equationautorefname{Equation}%
         \def\figureautorefname{Figure}%
         \def\subfigureautorefname{Figure}%
         \def\Itemautorefname{item}%
         \def\tableautorefname{Table}%
         \def\sectionautorefname{Section}%
         \def\subsectionautorefname{Section}%
         \def\subsubsectionautorefname{Section}%
         \def\chapterautorefname{Section}%
         \def\partautorefname{part}%
         \orgautoref}

\usepackage[textwidth=1cm,textsize=tiny]{todonotes}

\newcommand{\xmark}{\texttimes}

\newcolumntype{C}[1]{>{\centering\arraybackslash}p{#1}}
\begin{document}
\title{Trustworthy Data- and ML-Ops for \\
Intelligent Transportation Systems and Logistics}
\author{
Antonio Emanuele Cinà~\IEEEmembership{Member, IEEE},
Giovanni Scodeller~\IEEEmembership{Member, IEEE},\\
Cecilia Caterina Pasquale~\IEEEmembership{Member, IEEE},
Silvia Siri~\IEEEmembership{Senior, IEEE},\\
Davide Anguita~\IEEEmembership{Senior, IEEE},
Fabio Roli~\IEEEmembership{Fellow, IEEE},\\
Simona Sacone~\IEEEmembership{Senior, IEEE},
Luca Oneto~\IEEEmembership{Senior, IEEE}
}
\markboth{Accepted at IEEE Transactions on Intelligent Transportation Systems. DOI: \url{10.1109/TITS.2026.3711756}}%
{How to Use the IEEEtran \LaTeX \ Templates}

\maketitle

\begin{abstract}
The rapid evolution of Intelligent Transportation Systems and Logistics (\ITLogisticShort) has become a cornerstone of the modern social economy, relying heavily on the integration of Data, Artificial Intelligence (AI), and, more specifically, Machine Learning (ML). 
This paper provides a comprehensive review of Trustworthy Data and Machine Learning Operations (\dataops and \mlops) in the \ITLogisticShort domain, underscoring their importance in improving efficiency, reliability, and decision-making precision within transportation and logistics services. 
We begin by identifying gaps in current literature, offering clear context for our contribution. 
Subsequently, we explore the complexities of \dataops and \mlops, discussing their necessity, key components, available tools, practical insights, and case studies relevant to \ITLogisticShort. 
Additionally, we address the critical issue of Trustworthiness in AI applications, examining methods and tools designed to strengthen confidence in AI systems - especially in real-world \ITLogisticShort scenarios. 
The paper concludes with a discussion of persisting challenges and future prospects in this rapidly advancing field, aiming to serve as a vital resource for researchers, industry practitioners, and policy makers. 
Overall, this work not only establishes a foundational understanding of \dataops and \mlops in \ITLogisticShort but also charts a path for further research and innovation in developing more efficient, sustainable, and trustworthy intelligent transportation and logistics systems.
\end{abstract}


\section{Introduction}
\IEEEPARstart{I}{intelligent} Transportation Systems and Logistics (\ITLogisticShort) are at the backbones of the contemporary social economy~\cite{iyer2021ai}.
These systems play a crucial role in ensuring the efficiency and functionality of modern services~\cite{torre2018big,Yin2021DeepLO}. They support industries, such as transporting raw materials~\cite{sahal2021blockchain}, facilitating e-commerce deliveries~\cite{barenji2019intelligent}, and ensuring the timely distribution of medical supplies~\cite{caird2004vehicle}.

From the digitalization advancement of the late 90s', which enabled the complete digitization of tracking and monitoring systems~\cite{bhimani2014digitisation}, to the most recent Internet of Things (IoT) devices that enable real-time monitoring of the status of the goods~\cite{chen2012research}, \ITLogisticShort is nowadays rapidly producing a large amount of heterogeneous data~\cite{torre2018big,d2017real}.
Such a data explosion has driven the adoption of Artificial Intelligence (AI), particularly Machine Learning (ML) and advanced data analytics, to improve operational efficiency, diagnostics, and predictive capabilities, moving toward more automated \ITLogisticShort operations~\cite{haghighat2020applications,Sathyanarayana2023DeepLO}.
The integration of Data and AI in \ITLogisticShort empowers organizations to make more informed decisions and deliver more competitive services, improving efficiency and performance across various domains such as traffic management~\cite{ouallane2022fusion}, vehicle control~\cite{qureshi2013survey}, and warehousing~\cite{boysen2021last}.
For example, in traffic management, AI-driven predictive models optimize signal timings, reduce congestion, and decrease travel times~\cite{ouallane2022fusion}.
In vehicle control, AI algorithms enable adaptive cruise control and autonomous navigation, enhancing safety and reducing driver workload~\cite{qureshi2013survey}. 
In warehousing, intelligent systems streamline inventory management by automating stock tracking and optimizing space utilization, boosting operational efficiency and reducing costs~\cite{boysen2021last}.
For a more comprehensive collection of \ITLogisticShort applications that benefit from Data and AI, we refer to \autoref{sec:its-areas} and~\cite{haghighat2020applications}.
These systems have in common the use of Data and AI to process complex data patterns and provide solutions that support real-time decision-making and system optimization, becoming the core of the business for companies operating within \ITLogisticShort.

However, despite the inevitable hype accompanying Data and AI-based systems~\cite{sladkowski2016intelligent,haghighat2020applications}, unlocking their full potential within the \ITLogisticShort domain brings multiple challenges~\cite{Atwal2019PracticalDD}. \color{black}
In particular, data-related challenges in \ITLogisticShort applications reflect the Volume, Velocity, and Variety dimensions of the traditional 3V model of Big Data~\cite{dautov2017quantifying}.
Volume stems from the large number of entities that must be continuously monitored, including vehicles, drivers, passengers, infrastructures, and carriers~\cite{Barrachina2012CAOVAAC}.
Velocity arises from the need for real-time control, tracking, monitoring, and decision-making, since delayed information may result in outdated route recommendations, inefficient traffic control, missed delivery windows, or late emergency responses~\cite{zhu2018big}.
Variety derives from the heterogeneous nature of the collected data, which may refer to highly different quantities as traffic flows, average speeds, individual trajectories, intermodal transport units. In addition, data can be measured with different devices, techniques, and formats, which can include tabular records, videos, images, logs, graphs representing mobility networks, and data generated at different time scales and frequencies~\cite{amini2017big}.
For these reasons, systems are needed that can efficiently and effectively collect, store, and curate such data~\cite{iyer2021ai,torre2018big,zhu2018big}.
\color{black}

Furthermore, beyond data management, AI systems demand continuous monitoring of their performance both with technical metrics~\cite{bharadiya2023machine} (e.g., accuracy~\cite{novakovic2017evaluation}, non-regressiveness~\cite{yan2021positive}, and sustainability~\cite{spring2017scalable}) and human-related metric~\cite{sperrle2021survey} (e.g., robustness~\cite{cina2023wild}, fairness~\cite{caton2024fairness}, and privacy~\cite{liu2021machine}).
Fulfilling these needs becomes fundamental to preserve adequate system behavior and inducing trust in AI use in the area of \ITLogisticShort applications~\cite{chowdhury2020trustworthiness,tao2019trustdata}.
This is especially true when it comes to the adoption of state-of-the-art tools like Generative AI~\cite{fui2023generative} (e.g., Large Language Models~\cite{chang2023survey} or Multimodal Language Models~\cite{chang2023survey}) to bridge the gap between results of AI and their actual actionability (e.g., to support the operators in \ITLogisticShort in natural language or with images/videos~\cite{cui2024survey,sha2023languagempc}).
The need to face this challenge is also becoming mandatory due to the recent effort of regulatory bodies to enforce regulations such as the AI-Act~\cite{ai-act-council,eu2021ai-act,european_ai_act_2021}.
Consequently, it becomes essential to combine cultural philosophies, practices, and tools that enable \ITLogisticShort companies to deliver applications and services with high velocity~\cite{Liu2019IntelligentRP}, empowering them to better serve their customers and compete more effectively in the market~\cite{Atwal2019PracticalDD}.

Similar challenges have also been faced in the software development industry, leading to the introduction of Development Operations (\textit{DevOps}) to enhance both the speed and quality of software engineering~\cite{sen2021devops}.
\textit{DevOps} comprises two main components: Development (\textit{Dev}), which includes software planning, development, and testing, and Operations (\textit{Ops}), which is responsible for the continuous integration, deployment, and monitoring of the software~\cite{sen2021devops}.
When it comes to encompassing Data-driven software development and AI, researchers have extended traditional \textit{DevOps} practices to include Data Operations~\cite{Khalajzadeh2018DataOps,Rodrguez2020GoodPF} (\dataops) for managing data workflows, and ML Operations~\cite{treveil2020introducing,Testi2022MLOpsAT} (\mlops) for optimizing AI and ML pipelines.

\dataops defines a collection of practices and tools tailored to the design, deployment, and management of data pipelines in an agile, efficient, and usable manner~\cite{Atwal2019PracticalDD}.
\color{black}
\dataops establishes processes for addressing evolving and real-time data requirements by ensuring that large, fast, and heterogeneous data streams can be reliably collected, organized, processed, and made available when needed, across diverse sources and under rapidly changing operational conditions~\cite{Munappy2020FromAD,chowdhury2024data}. 
\color{black}

\mlops defines a set of practices and tools for deploying and maintaining ML models, which are essential components of AI-enabled systems, in production environments~\cite{Das2023WhatDM}.
\mlops provides developers with functionalities for integrating ML models into scalable and automated infrastructures, enabling continuous learning from new data through updated techniques and thereby preserving reliability and performance over time~\cite{makinen2021needs}.
Moreover, \mlops enables developers to seamlessly monitor different technical and human-related metrics, supporting the development and operationalization of Trustworthy AI (\TAI)-based systems~\cite{zeller2024toward}.
\color{black}
In this way, \mlops directly addresses the AI lifecycle challenges outlined above: it supports the automation of model retraining pipelines to cope with continuously evolving data distributions, manages the complexity of large-scale ML deployments through reproducible and versioned workflows, and operationalizes the continuous monitoring of both technical metrics, such as accuracy and non-regressiveness, and human-related metrics, such as robustness, fairness, and privacy, thereby supporting compliance with regulatory frameworks such as the AI Act~\cite{ai-act-council,eu2021ai-act,european_ai_act_2021}.
\color{black}

\begin{figure}
    \centering
    \includegraphics[width=\linewidth]{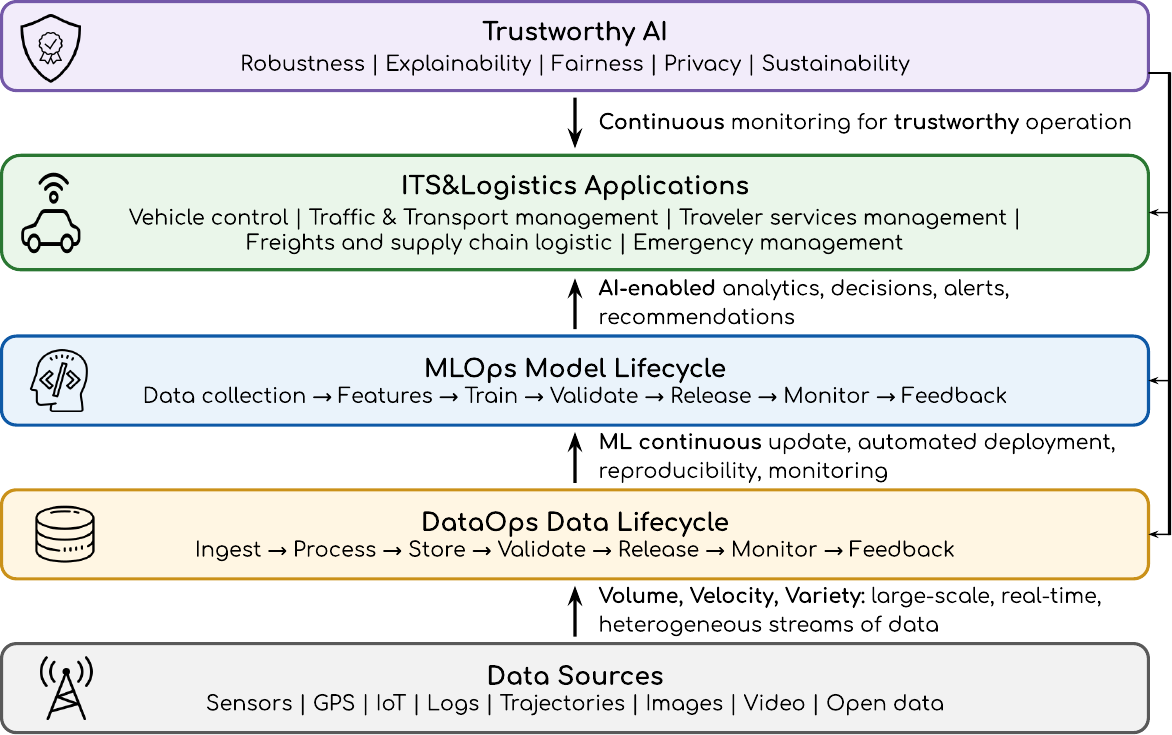}
\caption{\color{black}\dataops and \mlops in \ITLogisticShort. Heterogeneous data sources are governed through the \dataops lifecycle and subsequently support the \mlops lifecycle for model development, deployment, monitoring, and updating. The resulting AI-enabled services operate across the main \ITLogisticShort application domains, while \TAI provides a governance layer grounded in robustness, explainability, fairness, privacy, and sustainability.\color{black}}
\label{fig:dataops_mlops_tai_framework}
\label{fig:placeholder}
\end{figure}

\color{black}
Together, \dataops and \mlops provide a comprehensive and complementary response to the challenges inherent to the \ITLogisticShort domain. \dataops governs the data layer by automating the ingestion, processing, and storage of high-volume, high-velocity, and heterogeneous data streams, while \mlops governs the AI model lifecycle - from training and validation to deployment and monitoring - ensuring continuous performance, reproducibility, and compliance with trustworthiness requirements.
They are therefore indispensable for ensuring the reliability and effectiveness of data-driven solutions, particularly in applications where user safety, operational efficiency, and decision-making precision are of paramount importance~\cite{Atwal2019PracticalDD,makinen2021needs}.
Figure~\ref{fig:dataops_mlops_tai_framework} provides an end-to-end view of how \dataops, \mlops, and \TAI jointly support trustworthy \ITLogisticShort applications.
\color{black}

In this work, we present a review of Trustworthy \dataops and \mlops for \ITLogisticShort able to
\begin{itemize}
\item conduct a review of the current surveys related to the content of our work to pinpoint gaps in previously published surveys on this topic;
\item provide the motivations and detailed description for \dataops and \mlops surveying both available tools and their current adoption within \ITLogisticShort application;
\item provide motivations and a detailed description of \TAI, surveying both the key metrics and their current adoption within \ITLogisticShort applications;
\item underline the synergy between \TAI, \dataops, and \mlops in the context of \ITLogisticShort;
\item outline the unresolved challenges and future prospects in this burgeoning field of research.
\end{itemize}
\color{black}
Today, this review is timely and relevant, and we believe it can serve as a valuable resource for different audiences. It primarily targets early-career researchers seeking a structured entry point into the intersection of \dataops, \mlops, \TAI, and \ITLogisticShort. At the same time, it may support industrial practitioners interested in understanding the operational implications, tools, and case studies associated with these technologies, as well as senior researchers seeking to identify research gaps, future directions, and opportunities for real-world impact.
\color{black}

The rest of this review is organized as follows.
~\autoref{sec:review} conducts a search for current reviews related to the content of our work aimed at clarifying our contribution.
\autoref{sec:its-areas} focuses on presenting the main application areas in \ITLogisticShort and the diverse challenges they need to face demanding Trustworthy \dataops and \mlops solutions. 
~\autoref{sec:dataops} delves into the \dataops, focusing on the reason for its need, its essential parts, the available tools (open source and proprietary), some practical insights, and the current case studies in \ITLogisticShort.
~\autoref{sec:mlops} delves into the \mlops, following the same structure of ~\autoref{sec:dataops}, with the addition of an in-depth analysis of a specific case of \mlops and \llmops, which is the customization of \mlops for Large Language Models (LLMs) that are currently emerging as the next evolution of AI for \ITLogisticShort.
~\autoref{sec:trustAI} focuses on the \TAI, outlining methods and tools to integrate into the \mlops pipeline to enhance trust in AI-empowered systems, accompanied by examples of use cases in \ITLogisticShort.
~\autoref{sec:discussion} discusses open challenges and future perspectives for Trustworthy Data- and ML-Ops in \ITLogisticShort. Finally, ~\autoref{sec:Conclusions} concludes the paper.

\section{Literature Review}
\label{sec:review}
In the literature, it is possible to find many different reviews partially dealing with the topics covered in our work.
For this reason, this section scrutinizes the current reviews to allow the reader to trace these works and better understand our novel contribution.
In particular, we describe the search and selection criteria followed in this review and then we focus on the gaps in the current reviews that will be addressed in our work.

We searched for reviews in a specific time window using a series of keywords in the most important academic databases: Scopus\footnote{\url{https://www.scopus.com/}} and Google Scholar\footnote{\url{https://scholar.google.com/}}.
In terms of time window, we restricted the search to the last $10$ years.
For the keywords to be searched, we applied the following criteria
\begin{itemize}
\item papers containing in the title, abstract, or keywords one of the following strings:
``DataOps'',
``Data Models'',
``Data Collection'',
``MLOps'',
``LLMOps'',
``FMOps'',
``AIOps'',
``Trustworthy AI'',
``Trustworthy ML'',
``Transportation Systems'';
``Intelligent Transportation'';
``Logistics''.
\item papers containing in the title, abstract, or keywords one of the following strings:
``review'',
``sok'',
``overview'', and
``survey''.
\end{itemize}
Moreover, papers cited in relevant papers found using the previous criteria have also been included.
\begin{table*}
\caption{Meta-review of existing surveys in the \ITLogisticShort, \dataops, \mlops, and \TAI research fields.}
\label{table:meta-review}
\renewcommand{\arraystretch}{.95}
\setlength{\tabcolsep}{.1cm}
\centering
\begin{tabular}{|c|c|c|c|c|c|p{12cm}|}
\hline
\hline
\textbf{Ref.} & \textbf{Year} & \textbf{\ITLogisticShort} & \textbf{\dataops} & \textbf{\mlops} & \textbf{\TAI} & \textbf{Keywords} \\
\hline
\hline
\rowcolor[HTML]{EFEFEF}\multicolumn{7}{|c|}{TrustworthyAI Surveys} \\
\hline
\hline

\multirow{2}{*}{\cite{biggio18wild}} & \multirow{2}{*}{2018} &  &  &  & \multirow{2}{*}{\tcheck} & Adversarial machine learning, Evasion attacks, Poisoning attacks, Adversarial examples, Secure learning, Deep learning \\

\hline

\multirow{2}{*}{\cite{Li2021TrustworthyAF}} & \multirow{2}{*}{2021} &  &  &  & \multirow{2}{*}{\tcheck} & Trustworthy AI, robustness, generalization, explainability, transparency, reproducibility, fairness, privacy protection, accountability \\

\hline

\cite{liu2021machine} & 2021 &  &  &  & \tcheck & Machine learning, privacy, deep learning, differential privacy \\

\hline

\multirow{2}{*}{\cite{Oneto22TrustGraph}} & \multirow{2}{*}{2022} &  &  &  & \multirow{2}{*}{\tcheck} & Learning on graphs, Graph neural networks, Kernels for graphs, Trustworthy machine learning, Fairness, Privacy, Robustness, Explainability, Learning automatically Learning with guarantees \\

\hline

\cite{cina2023wild} & 2023 &  &  &  & \tcheck & Poisoning attacks, backdoor attacks, Machine learning, Computer vision, Computer security \\

\hline

\cite{caton2024fairness} & 2023 &  &  &  & \tcheck & Fairness, Accountability, Transparency, Machine learning \\

\hline

\multirow{2}{*}{\cite{chander2024toward}} & \multirow{2}{*}{2024} &  &  &  & \multirow{2}{*}{\tcheck} & Artificial Intelligence, Machine Learning, Deep Learning, Trustworthy AI, Explainability, Robustness, Decision making, Fairness, Data privacy. \\

\hline

\cite{albahri2024systematic} & 2024 &  &  &  & \tcheck & Artificial intelligence, Natural disasters, Explainability, Data fusion, Taxonomy, Trustworthy \\

\hline

\cite{wei2024trustworthy} & 2024 &  &  &  & \tcheck & Security and privacy, Computing methodologies, Distributed algorithms.\\

\hline

\cite{bailoni2025review} & 2025 &  &  &  & \tcheck & Artificial Intelligence, Trustworthy, Digital Health, Digital Assistants\\

\hline
\hline
\rowcolor[HTML]{EFEFEF}\multicolumn{7}{|c|}{\ITLogisticShort Surveys} \\
\hline
\hline

\multirow{2}{*}{\cite{Vlahogianni2015ComputationalIA}} & \multirow{2}{*}{2015} & \multirow{2}{*}{\tcheck} &  &  &  & Computational Intelligence, Transportation Problem, Swarm Intelligence, Intelligent Transportation System, Modeling Paradigm \\

\hline

\multirow{2}{*}{\cite{sladkowski2016intelligent}} & \multirow{2}{*}{2016} & \multirow{2}{*}{\tcheck} &  &  &  & Personalized mobility, Logistics for industry grid/industry 4.0, Smart cities, Individual/personalized transport, Personalized medicine, Individual products/mass customization \\

\hline

\cite{amini2017big} & 2017 & \tcheck & \tcheck &  &  & Intelligent Transportation System, Big Data, Kafka, Real-time traffic control \\

\hline

\multirow{2}{*}{\cite{Laa2018RoadTF}} & \multirow{2}{*}{2018} & \multirow{2}{*}{\tcheck} &  &  &  & Forecasting, Predictive models, Road traffic, Data models, Time series analysis, Traffic control, Big Data, Logistics \\

\hline

\cite{zhu2018big} & 2018 & \tcheck & \tcheck &  &  & Big Data, Transportation, Data analysis, Smart cards, Global Positioning System \\

\hline

\cite{torre2018big} & 2018 & \tcheck & \tcheck &  & \tcheck & No keywords provided \\

\hline

\cite{hireche2020machine} & 2020 & \tcheck &  &  &  & Machine Learning, Automatic incident detection systems, Detection Rate, False alarm rates\\

\hline

\cite{haghighat2020applications} & 2020 & \tcheck &  &  &  & Deep learning, ITS, Survey, Transportation systems \\

\hline

\cite{chowdhury2020trustworthiness} & 2020 & \tcheck &  &  & \tcheck & Roads, Global Positioning System, Peer-to-peer computing, Industries, Vehicular ad hoc networks, Safety \\

\hline

\cite{Yin2021DeepLO} & 2021 & \tcheck &  &  &  & Deep learning, Correlation, Predictive models, Data models, Convolution, Roads, Learning systems \\

\hline

\multirow{2}{*}{\cite{iyer2021ai}} & \multirow{2}{*}{2021} & \multirow{2}{*}{\tcheck} &  &  &  & Public transport, Safety management, Logistics, Intelligent transportation systems, Traffic management, Transport management systems \\

\hline

\multirow{2}{*}{\cite{bevsinovic2021artificial}} & \multirow{2}{*}{2021} & \multirow{2}{*}{\tcheck} &  &  & \multirow{2}{*}{\tcheck} & Artificial intelligence,
railway transport, machine learning, computer vision, traffic management, predictive maintenance \\

\hline

\cite{Zhou2022ACS} & 2022 & \tcheck &  &  &  & No keywords provided \\

\hline

\multirow{2}{*}{\cite{elallid2022comprehensive}} & \multirow{2}{*}{2022} & \multirow{2}{*}{\tcheck} &  &  &  & Autonomous driving, Deep learning, Reinforcement learning, Motion planning, Decision making, Vehicle control \\

\hline

\cite{guo2023sustainability} & 2022 & \tcheck &  &  & {\tcheck} &  Connected autonomous vehicles, artificial intelligence, traffic routing, sustainability, ethics\\

\hline

\multirow{2}{*}{\cite{Gaur2022ITSXAI}} & \multirow{2}{*}{2022} & \multirow{2}{*}{\tcheck} &  &  &  & Explainable Artificial Intelligence, AI, Machine Learning, Transportation System, Intelligent Transportation System, Autonomous Vehicle, Ethics in transportation \\

\hline
\cite{Ahmad2023DeepLM} & 2023 & \tcheck &  &  &  & Deep learning, Cloud computing, Edge computing, Fog computing, IoT, Models \\

\hline

\cite{Sathyanarayana2023DeepLO} & 2023 & \tcheck &  &  &  & Deep learning, Correlation, Predictive models, Data models, Convolution, Roads, Learning systems \\

\hline

\multirow{2}{*}{\cite{li2024survey}} & \multirow{2}{*}{2024} & \multirow{2}{*}{\tcheck} &  &  & \multirow{2}{*}{\tcheck} & Intelligent transportation system, Vehicular ad-hoc networks, Intelligent traffic lights, Virtual traffic lights, Mobility prediction, Sustainability, Security challenges, Case Studies \\

\hline

\multirow{2}{*}{\cite{khalil2024advanced}} & \multirow{2}{*}{2024} & \multirow{2}{*}{\tcheck} &  &  & \multirow{2}{*}{\tcheck} & Transportation, Surveys, Intelligent transportation systems, Autonomous vehicles, Vehicle dynamics, Large language models, Traffic control, Deep learning \\
\hline

\multirow{2}{*}{\textcolor{black}{\cite{mahmud2025integrating}}} &
\multirow{2}{*}{\textcolor{black}{2025}} &
\multirow{2}{*}{\textcolor{black}{\tcheck}} &
& & &
\textcolor{black}{Intelligent transportation systems, large language models, traffic flow optimization, autonomous driving, traffic management} \\

\hline

\textcolor{black}{\cite{yan2025generative}} &
\textcolor{black}{2025} &
\textcolor{black}{\tcheck} &
& & &
\textcolor{black}{Intelligent transportation system, generative AI, autonomous driving, traffic flow} \\

\hline
\hline
\rowcolor[HTML]{EFEFEF}\multicolumn{7}{|c|}{Big Data and \dataops Surveys} \\
\hline
\hline

\cite{dautov2017quantifying} & 2017 &  & \tcheck &  &  & Big Data, Measurement, Social network services, Tools, Business, Software \\

\hline

\cite{Khalajzadeh2018DataOps} & 2018 &  & \tcheck &  &  & Tools, Data visualization, Data analysis, Software, Machine learning, Business, Data models \\

\hline

\multirow{2}{*}{\cite{Atwal2019PracticalDD}} & \multirow{2}{*}{2019} &  & \multirow{2}{*}{\tcheck} & \multirow{2}{*}{\tcheck} &  & DataOps, Data Science, Data Engineering, Machine Learning, Artificial Intelligence, Advanced Analytics, Cloud Analytics, Agile Data Science, Lean Analytics, MLOps, Big Data \\

\hline

\cite{tao2019trustdata} & 2019 &  & \tcheck &  & \tcheck & Data Collection, Monitoring, Event detection, Reliability, Informatics, Data privacy, Security \\

\hline

\cite{Munappy2020FromAD} & 2020 &  & \tcheck &  &  & DataOps, Data Pipelines, Continuous Monitoring, DevOps, Data technologies, Agile  \\

\hline

\cite{Rodrguez2020GoodPF} & 2020 &  & \tcheck &  &  & No keywords provided \\

\hline

\cite{chowdhury2024data}& 2024 &  & \tcheck &  &  & No keywords provided \\
\hline
\hline
\rowcolor[HTML]{EFEFEF}\multicolumn{7}{|c|}{\mlops Surveys} \\
\hline
\hline

\multirow{2}{*}{\cite{tamburri2020sustainable}} & \multirow{2}{*}{2020} &  &  & \multirow{2}{*}{\tcheck} &  & Scientific computing, Decision making, Machine learning, Market research, Software systems, Sustainable development, Middleware \\

\hline

\cite{sen2021devops} & 2021 &  &  & \tcheck &  & DevOps, DevSecOps, AIops 
\\

\hline

\cite{makinen2021needs} & 2021 &  &  & \tcheck &  & Training, Computational modeling, Time series analysis, Focusing, Production, Tools, Data models \\

\hline

\multirow{2}{*}{\cite{Testi2022MLOpsAT}} & \multirow{2}{*}{2022} &  &  & \multirow{2}{*}{\tcheck} &  & MLOps, Continuous monitoring, Continuous integration, Continuous delivery, Continuous training, XAI, Sustainability. \\

\hline

\multirow{2}{*}{\cite{Das2023WhatDM}} & \multirow{2}{*}{2023} &  &  & \multirow{2}{*}{\tcheck} &  & MLOps, ML operations, Machine learning, Artificial intelligence, ML monitoring, ML observability, Information systems \\

\hline
\textcolor{black}{\cite{black2023toward}} &
\textcolor{black}{2023} &
& &
\textcolor{black}{\tcheck} &
\textcolor{black}{\tcheck} &
\textcolor{black}{No keywords provided} \\
\hline
\cite{eken2024multivocal}  & 2024 & &  & \tcheck &  &  MLOps, Machine Learning Operations, DevOps, CI/CD \\ 

\hline
\cite{zeller2024toward} & 2024 & \tcheck &  & \tcheck &  & Safety, Machine learning, MLOps, Continuous assessment, Automated driving systems, Railway \\ 

\textcolor{black}{\cite{zarour2025mlops}} &
\textcolor{black}{2025} &
& &
\textcolor{black}{\tcheck} &
&
\textcolor{black}{MLOps, DevOps, ML, Best practices, Automation, Deployment} \\
\hline

\multirow{2}{*}{\textcolor{black}{\cite{bayram2025towards}}} &
\multirow{2}{*}{\textcolor{black}{2025}} &
& &
\multirow{2}{*}{\textcolor{black}{\tcheck}} &
\multirow{2}{*}{\textcolor{black}{\tcheck}} &
\textcolor{black}{Artificial intelligence, machine learning, Trustworthy AI, robustness, MLOps systems, DataOps, ModelOps, model performance} \\

\hline
\hline
\end{tabular}
\end{table*}
\color{black}
To reduce the risk of missing relevant surveys that adopt different terminology, the screening process also considered broader terms used in adjacent research areas, such as AI safety and governance, since relevant surveys may vary in terminology depending on their research community.
However, because these broader terms often retrieve heterogeneous literature beyond the scope of this paper, we report only the keywords that are most prominent and directly related to operational AI lifecycles and \ITLogisticShort applications.
In cases where multiple reviews address overlapping topics, we report the most recent or comprehensive one.
\color{black}
The search results are summarized in Table~\ref{table:meta-review}, where each row represents a survey paper. A checkmark (\tcheck) indicates coverage of the four topics considered in this paper: \ITLogisticShort, \dataops, \mlops, and \TAI. 
Several observations can be derived from Table~\ref{table:meta-review} regarding the current reviews on the topics addressed in this work and our contribution.

Unlike existing surveys in the literature, which generally provide a broad overview of the entire spectrum of \dataops, \mlops, and \TAI for general-purpose applications, our work focuses explicitly on the practical needs within the \ITLogisticShort sector. 
Existing works often treat \TAI, \dataops, and \mlops as separate entities, overlooking their potential synergy. 
Some works, such as \cite{zhu2018big, amini2017big}, analyze the relevance of \dataops in \ITLogisticShort applications and the associated challenges but do not delve into the role of \TAI and \mlops development. 
Tao et al.\cite{tao2019trustdata} investigate the role of data quality in maintaining trustworthiness in data-driven systems but lack a focus on \ITLogisticShort. 
\color{black}
More recent studies have begun to link operational ML pipelines with trustworthiness.
For example, Bayram and Ahmed~\cite{bayram2025towards} provide a comprehensive survey of robustness in \mlops systems, while Black et al.~\cite{black2023toward} examine fairness from a pipeline-aware perspective.
However, these works do not address multiple \TAI principles in an integrated manner, nor do they jointly analyze \dataops and \mlops within transportation and logistics applications.
\color{black}
The most comprehensive work, Zhu et al.\cite{zhu2018big}, covers three out of the four research domains investigated in this survey, excluding \mlops. Moreover, their discussion on \TAI focuses solely on the robustness and privacy principles.

\color{black}
In this survey, we jointly focus on five trustworthiness dimensions: robustness, interpretability, fairness, sustainability, and privacy.
We also review tools that support the assessment of these requirements, helping practitioners evaluate whether data pipelines, ML models, and deployed services satisfy these principles throughout their lifecycle.
\color{black}
\section{Intelligent Transportation Systems and Logistics}\label{sec:its-areas}
In this section, we provide an overview of the main areas of \ITLogisticShort, drawing on several notable reviews in the field~\cite{alam2016introduction,iyer2021ai,qureshi2013survey,haghighat2020applications,zhang2011data,Zhou2022ACS}.
This overview aims to frame the problems and opportunities related to Trustworthy Data- and ML-Ops, which are examined in detail in the following sections.
\color{black}
Building on these existing works, we organize \ITLogisticShort applications into an operational taxonomy tailored to the scope of this survey.
The objective is to group applications that face similar challenges in terms of \dataops, \mlops, and \TAI, rather than to provide a purely descriptive classification of the domain.
These applications share common needs, such as handling large and heterogeneous data, managing task complexity, and ensuring service trustworthiness.
However, each area also presents specific operational requirements, for example in terms of real-time processing, scalability, safety, privacy, fairness, sustainability, or explainability.
This taxonomy is therefore used in the following sections to organize the discussion of case studies and to connect each application area with its most relevant requirements.
\color{black}

\myparagraph{Vehicle Control Systems.}
\color{black}
Vehicle control systems involve vehicle-level technologies designed to improve vehicle performance and safety through automation and driver assistance~\cite{qureshi2013survey}.
\color{black}
Examples of these systems are: advanced driver-assistance systems~\cite{martinez2015driving}, vehicle-to-vehicle or vehicle-to-everything~\cite{tong2019artificial}, self driving cars~\cite{grigorescu2020survey}, electronic stability control~\cite{ferguson2007effectiveness}, and incident management~\cite{singh2021highway}.
These systems are developed and integrated into self-driving cars to increase efficiency, improve vehicle handling, and support humans to correct potential errors or improve their usage experience~\cite{grigorescu2020survey}. 
Furthermore, if connected together in a network~\cite{tong2019artificial}, these systems can communicate with each other and the surrounding infrastructure for rapid detection, response, and mitigation of traffic incidents~\cite{singh2021highway}. 
Vehicle technologies, including ML components, are rigorously tested, validated, and monitored in real time to detect and correct potential issues before they cause harm.
Furthermore, from an operational perspective, these systems collect and process vast amounts of data, including sensitive information about drivers and their behaviors, thus demanding protection mechanisms (e.g., data encryption) to avoid unauthorized access and ensure compliance with safety and privacy regulations. 
Lastly, they must be able to scale in real time to accommodate varying traffic conditions, an increasing number of connected vehicles, and the dynamic nature of road environments~\cite{singh2021highway}.

\myparagraph{Traffic Management.} 
It focuses on optimizing traffic flow, reducing congestion, and improving overall road network efficiency~\cite{ouallane2022fusion}.
For instance, these systems can manage traffic lights~\cite{avatefipour2018traffic} and road diversions~\cite{alomari2019road} to enhance road traffic and benefit drivers. 
Another example is dynamic toll pricing systems~\cite{lombardi2021model}, which use traffic data to adjust toll rates~\cite{guerreiro2016architecture}. 
These systems promote the use of highways when state roads are heavily congested, thereby balancing the traffic load across different road networks.
Traffic management, however, does not only deal with road and city traffic.
It also encompasses services and implementations in navigation systems~\cite{zhilenkov2018problems}, air transport systems~\cite{degas2022survey}, water transport systems~\cite{nti2022environmental}, and rail systems~\cite{tang2022literature}. 
In traffic management applications, we find the need for real-time data processing to provide timely and accurate recommendations, which directly impacts the efficiency of traffic management and congestion reduction. 
Practically, and from an operational perspective, they require continuous processing of network-level traffic data to support congestion control, routing efficiency, pricing decisions, and environmental objectives. 
At the same time, these systems face with privacy, security, fairness, and sustainability principles: privacy and security because of the sensitive nature of user data involved in the systems and their integrity~\cite{de2017traffic}, fairness because the inner algorithms and models should not favor users or areas in recommendations and pricing~\cite{loginextsolutions}, and sustainability because of the environmental impact these systems can have on traffic pollution~\cite{musa2023sustainable}.

\myparagraph{Transportation Management Systems.}
\color{black}
Unlike traffic management systems, which primarily focus on regulating traffic flows and network conditions, transportation management systems aim to model travelers' journey patterns across multimodal transportation networks, including car, bus, rail, and metro systems. These systems use such information to plan, optimize, and improve mobility services for users, including public transit routes, ride-sharing services, and transportation infrastructure~\cite{zhu2018big,fiore2019integrated,alic2019bigsea,ushakov2022artificial,saleem2022smart}.
\color{black}
These applications strongly rely on information collected from multiple, widespread, and heterogeneous data sources (e.g., social networks and smartphones) and on the ability to extract actionable insights from them.
For instance, these systems can leverage large-scale data from social media to understand public opinions on transportation services or to detect traffic events (e.g., accidents, breakdowns, and road closures) and consequently improve planningpolicies~\cite{qi2020framework,ruiz2016social}.
Decisions made based on these data have a strong societal and business impact. 
Starting from privacy, given the usage of sensitive data (e.g., personal data from transit cards), they must implement robust security measures to protect users data from unauthorized access, breaches, and misuse.
Furthermore, they have to handle peak traffic times, where transportation demand increases, and future expansions without performance degradation, and allocate transportation resources in a way that benefits all communities fairly. 

\myparagraph{Traveler Information and Experience Systems.} 
\color{black}
This category comprises \emph{user-facing systems} that leverage algorithmic and AI-based methods to search for, simulate, plan, and recommend routes, destinations, or mobility options tailored to individual travelers' needs and preferences, thereby enhancing the overall travel experience~\cite{qureshi2013survey,cepeda2022deep,al2021hybrid,levinson2003value,wang2018will,siripanpornchana2016travel,chi2021design}.
\color{black}
For example, these systems can offer drivers the fastest route to their destination, continuously updated in real-time to accommodate regular and unexpected traffic congestion, thus reducing overall travel costs and improving user satisfaction~\cite{wang2018will,siripanpornchana2016travel}.
\color{black}
Although these systems may rely on traffic data and routing algorithms, their objective differs from that of traffic management systems: they primarily seek to optimize the information, guidance, and recommendations provided to individual users, rather than directly controlling traffic flows or coordinating network-level operations.
\color{black}
The challenge is to deliver the service in real time to a large number of users, using the most recent data from a large number of sources~\cite{asaithambi2023thematic,chi2021design}.
The focus in these applications is not on security and safety, as the division mainly impacts user comfort or other metrics such as sustainability.
However, they need to integrate explainable models to provide clear and understandable explanations for recommendations (e.g., route deviations), help users make informed decisions, and build trust in the system~\cite{gao2023towards}.

\color{black}
\myparagraph{Freight and Supply Chain Logistics.}
These systems support the full supply-chain continuum, from managing cargo shipments across multiple transportation modes, including sea, road, air, inland waterways, and rail, to coordinating urban freight policies and optimizing warehousing and last-mile delivery operations~\cite{qureshi2013survey,tsolaki2023utilizing,boysen2021last}. 
At the freight level, they can be used to reduce pollutant emissions, optimize energy consumption~\cite{de2021simulation}, and coordinate information exchange among seaport stakeholders~\cite{sarabia2019seaport}. 
At the warehousing level, smart systems leverage real-time data from RFID tags and IoT sensors to track goods, manage inventory levels, forecast demand, reduce waste, and optimize storage layouts~\cite{pandian2019artificial,lee2018design}. 
At the delivery level, last-mile systems integrate GPS data, real-time traffic updates, and customer feedback to dynamically adjust routes, reduce fuel consumption, and improve delivery reliability~\cite{borgi2017big}.

Across this spectrum, constraints such as energy efficiency, operational resilience, and business and environmental sustainability are of primary importance~\cite{marchet2014environmental}. 
Disruptions, data compromises, or delays in data processing can propagate along the supply chain, causing logistics errors and potentially significant financial losses. 
Therefore, continuous data monitoring is essential to assess data integrity, while scalable implementations are needed to accommodate variable cargo volumes and adapt to changing demand patterns and regulatory requirements. 
These systems should also be designed to mitigate biases that may lead to unequal treatment of customers or regions based on location, accessibility, or socioeconomic status~\cite{lyu2022towards,chen2023same}. 
For example, last-mile delivery algorithms may unintentionally prioritize neighborhoods over less affluent areas because of disparities in road infrastructure, demand density, or service accessibility, resulting in poorer service quality for underserved communities. 
Addressing these risks requires incorporating fairness-aware constraints to support a more equitable distribution of logistics services across regions~\cite{lyu2022towards}.
\color{black}

\myparagraph{Emergency Management.}
These systems are essential to coordinate responses to unforeseen events and minimize their impact on communities~\cite{thibaud2018internet,huang2021systematic}. 
Largely applied in large-scale disaster contexts, they aim to ensure the timely delivery of emergency resources and rescue services into the affected regions so as to assist in rescue activities~\cite{jiang2019emergency}.
These systems connect more data sources together to collect data and use them to prevent and mitigate crises such as natural disasters~\cite{arinta2019natural}, terrorist attacks~\cite{huamani2020machine}, 
or public health emergencies~\cite{garriga2022machine}.
The challenge lies in the timely collection and processing of data to rapidly extract information that can prevent or better manage emergencies~\cite{aringhieri2022fairness}. 
Concurrently, they aim to ensure the integrity of collected data to avoid misinformation, which could lead to ineffective or harmful responses. 
Lastly, these systems must preserve fairness in their decisions and operations, e.g., equity and fairness in humanitarian relief services~\cite{aringhieri2022fairness}.

\section{\dataops}
\label{sec:dataops}

Data has become crucial in \ITLogisticShort applications, driving the need for efficient process streamlining and operational efficiency~\cite{alexakis2022distributed,torre2018big}.
This involves seamless data collection, transfer, storage, and accessibility with minimal human intervention, facilitating complete automation of various logistical tasks~\cite{zhu2018big,chowdhury2024data}.
However, managing data in the \ITLogisticShort sector poses challenges~\cite{wang2023technical}.
These applications generate large volumes of data~\cite{torre2018big}, which are important for scientific progress and process optimization. 
However, their architecture, involving a variety of interconnected components-such as vehicles, sensors, and digital machinery-creates a complex data environment characterized by high velocity and continuous streams~\cite{alexakis2022distributed}. 
As a result, advanced and scalable data management solutions are required~\cite{wang2023technical,Atwal2019PracticalDD}, and \dataops~\cite{DataOpsEnterpise,ereth2018dataops} provides a promising solution within the \ITLogisticShort sector.
It integrates practices and technologies to streamline the entire data lifecycle, from collection to analysis, ensuring efficient processing and availability of data generated~\cite{Munappy2020FromAD}. 
\color{black}
In the end-to-end view presented in Figure~\ref{fig:dataops_mlops_tai_framework}, this section corresponds to the \dataops layer, where heterogeneous \ITLogisticShort data sources are transformed into reliable, monitored, and reusable data assets for downstream analytics and ML workflows.
\color{black}

In the following, we start by revisiting the underlying motivations for adopting \dataops, as seen in~\autoref{sec:its-areas} and further detailed in ~\autoref{subsec:dataops:Motivations}.
Subsequently, we delineate the various stages encompassed within a \dataops pipeline, outlined in ~\autoref{subsec:dataops:Life}.
This is followed by an examination of the principal open-source (\autoref{subsec:dataops:openTools}) and proprietary \dataops tools (\autoref{subsec:dataops:proprietaryTools}).
A comparative analysis of these tools is presented in ~\autoref{subsec:dataops:comparison}, offering practical insights and implications.
The section culminates with a presentation of current case studies in the \ITLogisticShort domain, elaborated in ~\autoref{subsec:dataops:casestudy}.
This structured approach provides a comprehensive overview of the \dataops landscape, highlighting its practical applications and theoretical underpinnings.
\subsection{Motivations}
\label{subsec:dataops:Motivations}
%
As seen in \autoref{sec:its-areas}, the adoption of advanced and scalable data management solutions from \dataops has been catalyzed by the proliferation of three primary factors~\cite{guerreiro2016architecture,zhu2018big}: Volume, Variety, and Velocity. 
Although the services and applications developed behind the different areas of \ITLogisticShort are different in nature, they all must address these three factors during their development and service life cycle. 
For example, while volume is a common feature, variety and velocity are together particularly important when it comes to traffic and emergency management applications. 
These applications must continuously collect data from different sources and process it in real time to maintain quality of service and usefulness for users.
In the following paragraphs we describe where these factors originate and some examples of their incorrect management.

\myparagraph{Volume.}
The massive volume of data within the \ITLogisticShort sector is experiencing exponential growth, a trend further accelerated by the emergence of new data sources~\cite{guerreiro2016architecture}.
Presently, \ITLogisticShort applications are leveraging data derived from three primary sources:
\begin{itemize}
\item Novel Sensors~\cite{borgi2017big}: this category encompasses a range of devices such as Inertial Measurement Units (IMUs), On-Board Units (OBUs), GPS navigators, LIDARs, and various environmental sensors. 
These instruments are pivotal in gathering high-fidelity, real-time data pertinent to logistics operations;

\item Social Data~\cite{borgi2017big}: this source includes Geo-tagged information obtained from social media platforms.
Such data is instrumental in extracting user trajectories and behavioral patterns, offering valuable insights into consumer behavior and logistics demands;

\item Open Data~\cite{torre2018big}: this refers to datasets related to national public transportation services, which have become increasingly available due to the growing trend towards openness and transparency among government institutions.
These datasets are crucial for understanding and optimizing public transportation logistics.
\end{itemize}
The integration of these diverse data sources has undeniably inaugurated a new epoch of possibilities, services, and applications in the \ITLogisticShort field~\cite{zhu2018big,zhang2011data}.
However, this rapid influx of data simultaneously poses significant challenges.
It requires the development of scalable solutions for the collection, management, processing (including cleaning and cleansing), and dissemination of these data~\cite{DataOpsEnterpise,chowdhury2024data}.
Inefficient handling of large data volumes can lead to significant delays in data processing, resulting in outdated or inaccurate information being fed into decision-making processes. 
For example, in traffic management, outdated data can lead to incorrect traffic light timings or inaccurate route recommendations, exacerbating congestion and increasing travel times. 
Moreover, poor data volume management can overwhelm storage systems and lead to data loss or corruption, further compromising the reliability of the service. This can erode user trust and satisfaction, ultimately impacting the operational efficiency and financial viability of businesses reliant on these applications. 
This situation underscores the inadequacies of traditional methodologies in handling the complexity and volume of current data streams in the \ITLogisticShort sector~\cite{zhu2018big,borgi2017big}.

\myparagraph{Velocity.}
Regarding velocity, the \ITLogisticShort industry heavily depends on real-time data processing for critical decision-making processes~\cite{zhu2018big}. 
Timely data processing is crucial for several operations, including route optimization~\cite{wang2014real}, which seeks to minimize distances and travel times in real time; delivery time reduction~\cite{nagpal2021use}, which tracks delivery schedules to improve customer satisfaction; traffic updates~\cite{amini2017big}, which requires collecting real-time traffic information to avoid delays and ensure on-time deliveries; and the overall efficiency of logistics operations~\cite{borgi2017big}. 
When the data processing mechanisms fail to match this high velocity, the efficacy of the entire system is compromised, undermining its capability to support the intended applications~\cite{zhu2018big}, or leading to system crashes.
For instance, in route optimization, delays in processing traffic updates can cause vehicles to follow suboptimal paths, increasing travel time and fuel consumption. 
In delivery logistics, slow data processing can disrupt scheduling, leading to missed delivery windows and decreased customer satisfaction. 
Over time, the increasing velocity of data generation and the demands of associated services have highlighted the inadequacies of traditional methods and tools in addressing these challenges.
This revelation underscores the need for innovative approaches in data processing within the \ITLogisticShort sector~\cite{borgi2017bigsecond}.

\myparagraph{Variety.}
Finally, regarding variety, the \ITLogisticShort sector is witnessing an ever-increasing diversification in its data sources and types~\cite{zhu2018big}.
This trend marks a significant shift from the reliance on traditional tabular data formats to a more complex array of data types such as graph data~\cite{zhou2020variational} (utilized for representing complex relationships and network structures, essential in logistics for route optimization and network analysis), sequences~\cite{gupta2020early} (sequential data, such as time series, are crucial for forecasting and trend analysis in logistics), logs~\cite{ji2014cloud} (log files provide detailed records of events, transactions, and operations, invaluable for auditing and process optimization), 
audio~\cite{parineh2024acoustic} (emerging as a data source for voice-based commands and communications in logistics operations) and natural language~\cite{putri2021intelligent} (text data, including customer feedback and communication, are increasingly analyzed for insights into logistics performance and customer satisfaction).
These diverse data types necessitate sophisticated management strategies, including versioning (along with their metadata), processing, and linking~\cite{warren2015big,storey2017big}.
Integrating these varied data forms without losing any critical information is crucial, as such data can be pivotal for future uses and the development of innovative services.
Failure to properly integrate and manage diverse data types can lead to incomplete or inaccurate analyses, negatively impacting decision-making processes. 
For instance, neglecting the integration of graph data might result in suboptimal route planning, missing out on critical network relationships~\cite{guo2020optimized}. 
Additionally, ignoring log files may prevent the identification of operational inefficiencies or security breaches~\cite{ye2020cyber}. 
Furthermore, the lack of a unified advanced approach to manage these varied data types can result in fragmented data silos~\cite{holmstrom2024stalled}, impeding comprehensive analysis and innovation.
In conclusion, managing data variety represents one of the most formidable challenges for contemporary data pipelines in the \ITLogisticShort sector.
It exposes the limitations of traditional methods and tools, underscoring the need for advanced solutions capable of handling the complexity and diversity of modern logistics data~\cite{borgi2017bigsecond}.
\subsection{DataOps Lifecycle}
\label{subsec:dataops:Life}
\begin{figure}
\centering
\includegraphics[width=\columnwidth]{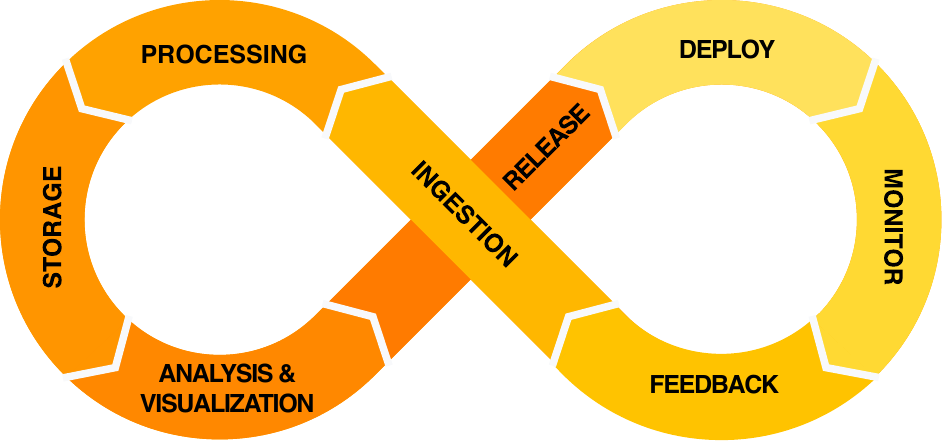}
\caption{\dataops abstract lifecycle.}
\label{fig:dataops_design}
\end{figure}
Although there is widespread agreement within the data management community regarding the initiation (data acquisition) and termination points (analysis and visualization) of the data lifecycle, consensus on the intermediate stages is limited~\cite{Atwal2019PracticalDD,Munappy2020FromAD}. 
Nonetheless, various proposed models of the \dataops lifecycle consistently depict an iterative workflow compliant with agile principles~\cite{Larman2003IterativeAI}, which facilitates continual refinement, adaptation, and enhancement of data processes to address evolving application requirements and challenges~\cite{Rodrguez2020GoodPF}.
Within this paper, we propose a generalized \dataops process, depicted in \autoref{fig:dataops_design}, designed to reflect the principles of DevOps for efficient software development.
The proposed schema abstracts the \dataops lifecycle into eight distinct steps: ingestion, processing, storage, analysis and visualization, release, deployment, monitoring, and feedback.

\myparagraph{Data Ingestion.}
Data ingestion in the \dataops pipeline involves planning, acquiring, and loading data (along with its metadata) from various sources into a centralized storage system~\cite{Munappy2020FromAD}. 
It requires precise data requirement planning, which entails a thorough understanding of application-specific needs, potential data sources and acquisition methods, expected data volume, and relevant attributes. 
This stage corresponds to the well-known \textit{Plans} phase in DevOps~\cite{sen2021devops}, where project objectives and requirements are defined.

\myparagraph{Data Processing.}
The acquired data enters into a processing stage which entails the implementation of processing~\cite{Munappy2020FromAD}, cleaning~\cite{Munappy2020FromAD}, filtering~\cite{Munappy2020FromAD}, and refinement~\cite{Munappy2020FromAD} procedures to ensure adherence to the required format and structure defined by the metadata~\cite{Munappy2020FromAD}.
Any inaccuracies or inconsistencies are tracked and possibly addressed, resulting in a dataset that is accurate and aligned with its metadata~\cite{Munappy2020FromAD}.
This stage aligns with the Code phase in DevOps~\cite{sen2021devops}, where developers write and integrate code for new features or improvements.

\myparagraph{Data Storage.}
The data storage phase emerges as the central pillar for managing the data volume, velocity, and variety generated by data ingestion and processing~\cite{thusoo2017creating,Munappy2020FromAD}. 
For this purpose, data storage architectures have gradually evolved over the years (i.e., relational database~\cite{harrington2016relational}, data warehouse~\cite{krishnan2013data}, data lake~\cite{nargesian2019data}, and data lakehouse~\cite{armbrust2021lakehouse}), underscoring the dynamic nature of data storage solutions. 
Currently, \textit{data lakehouse}~\cite{armbrust2021lakehouse} are the standard solution as they leverage the strengths of both data warehouses~\cite{krishnan2013data} and both data lakes~\cite{harby2022data}, providing a centralized repository to store structured, semi-structured, and unstructured data.
This solution ensures flexibility, scalability, and cost-effectiveness. The storage phase in \dataops aligns with the Build phase of DevOps~\cite{sen2021devops}, where a developer commits their code to a shared code repository.

\myparagraph{Analysis and Visualization.}
The analysis and visualization phases align with the Test stage of DevOps pipelines~\cite{sen2021devops}.
Its role relies on inspecting the stored data to test the correct operativity of the \dataops pipeline.
This step is of paramount importance since it allows to actually inspect and provide data to higher level services (e.g., \mlops), which leverage these data to generate value~\cite{grover2018creating}.
In fact, the primary role of the \dataops substrate is to enable data-driven services, which can span from simple real-time tracking of a service to the delivery of the most advanced AI-based services.

\myparagraph{Release, Deploy, Monitoring, and Feedback.}
The concluding stages of \dataops pipelines align exactly with the ones of the DevOps~\cite{sen2021devops} pipeline, namely the \textit{Ops}.
First, the release phase ensures data consistency and verifies the proper functioning of the \dataops pipeline for deployment~\cite{Atwal2019PracticalDD}. 
After release, the deployment phase involves the deployment of developed data pipelines into the designated testing or production environments~\cite{Munappy2020FromAD}. 
Post-deployment, monitoring tools and frameworks are utilized to observe the health and performance of data pipelines~\cite{ereth2018dataops}. 
Metrics such as data throughput, processing time, error rates, and alerting mechanisms can be integrated to assess system performance and promptly notify teams of any issues or anomalies. 
Finally, the \dataops feedback loop involves the extraction of insights on data quality and performance from monitoring tools, coupled with the assimilation of feedback from end-users or downstream systems to correct, improve, or better monitor the other steps of the \dataops pipeline~\cite{Munappy2020FromAD}.

\subsection{Open-Source Solutions}
\label{subsec:dataops:openTools}
This section reviews the most common and adopted open-source tools providing free access for implementing the \dataops pipeline, rendering them advantageous choices for development projects. 
These tools empower developers to construct and customize their data architectural framework while facilitating seamless integration with various third-party libraries and come with the support of the open-source community, documentation, and discussion groups~\cite{boulanger2005open}.
It is important to note that a single tool often does not implement all components of the \dataops pipeline, and developers typically combine multiple tools to achieve the complete pipeline~\cite{Atwal2019PracticalDD,amini2017big,gillani2023machine}.

\myparagraph{Apache Kafka.} 
Apache Kafka~\cite{garg2013apache,apacheApacheKafka} is a distributed streaming platform widely acclaimed as a messaging system in real-time data processing. It supports the ingestion and monitoring phases of the \dataops pipeline. Its versatility and seamless ability to ingest data from diverse sources, including web applications and databases, is well known in both industry and research with more than $29$ thousand stars on GitHub as of January 2025.
The design of the data processing pipeline not only empowers Kafka to maintain high throughput and low latency while handling substantial data volumes but also enables smooth integration with a diverse array of Apache tools for future data processing.
Finally, Kafka offers robust security measures, encompassing data encryption during transmission, authentication, and authorization.

\myparagraph{Apache Hadoop Distributed File System.}
Apache Hadoop Distributed File System (HDFS)~\cite{borthakur2007hadoop,apacheApacheHadoop} is a distributed file system that stores vast volumes of data across multiple cluster nodes. It supports the development of the storage phase of the \dataops pipeline, being the most used storage tool in data lakehouse~\cite{harby2022data,armbrust2021lakehouse}.
HDFS is a foundational component within the Apache framework, boasting more than $14$ thousand stars on GitHub as of January 2025. 
The architecture of HDFS is designed to provide high reliability (through data replication) and scalability (through the possibility of adding nodes) in storing and maintaining large-scale data.
HDFS natively deals with variety thanks to its ability to handle unstructured data, which includes tables, text, audio, video, images, and more.

\myparagraph{Apache Spark.}
Apache Spark~\cite{zaharia2016apache,apacheApacheSparkx2122} is an open-source framework for scalable in-memory data processing.
With over $37$ thousand stars on GitHub as of January 2025 it has proven to be the golden standard solution for implementing the processing phase of the \dataops pipeline.
A key feature of Apache Spark is the distributed (i.e., ability to leverage multiple nodes) in-memory processing capabilities reducing the number of disk accesses and leading to a significant acceleration in data processing. 
Finally, Apache Spark seamlessly integrates with various programming languages and cloud platforms, not limited to Apache's offerings but extending to third-party tools. 

\myparagraph{Apache Flink.}
Apache Flink~\cite{carbone2015apache, apacheApacheFlink} is an alternative open-source framework to Apache Spark.
Although it shares many similarities with Apache Spark, Apache Flink focuses on stream processing.
Stream processing revolves around continuous data management which is crucial when the immediate processing of data streams is essential. 
With more than $24$ thousand stars on GitHub as of January 2025, Flink is the golden standard in this setting.
Stream processing reduces storage requirements compared to batch processing, but the choice between stream and batch processing may vary depending on specific task requirements, considering precision, velocity, and efficiency factors.

\myparagraph{Ops Tools.}
The Ops tools are related to the release, deploy, monitor, and feedback steps in the \dataops lifecycle (see Figure~\ref{fig:dataops_design}), namely the ones that are entirely inherited from DevOps.
Of course, in the \dataops lifecycle, some of them are the same used in DevOps while others are tailored toward the specific context.
dbt (Data Build Tool)~\cite{getdbtLabsTransform} is a command-line tool, focused on the release step, that empowers analysts and engineers to execute rigorous testing of data models to assess their reliability and accuracy before deployment.
Docker~\cite{dockerHome} and Kubernetes~\cite{kubernetesProduction} jointly form an indispensable duo for deployment~\cite{dessalk2020scalable}. 
Docker provides container solutions for encapsulating applications and dependencies. 
Kubernetes orchestrates containerized applications, furnishing scalability and adept management capabilities. 
Docker and scalability integration facilitates the deployment and scaling of containerized data processing applications~\cite{dessalk2020scalable}.
Prometheus~\cite{prometheusPrometheusMonitoring}, Apache Airflow~\cite{apacheAirflow}, and Jenkins~\cite{jenkinsJenkins} are open-source tools designed for monitoring and alerting, providing real-time insights into pipeline performance and functionality. 
On the other hand, proprietary tools like Power BI~\cite{powerBI} and Tableau~\cite{tableau} offer user-friendly integration, robust support, and advanced data visualization features.
Finally, version control systems (e.g., Git~\cite{gitscm} and SVN~\cite{apacheApacheSubversion}) represent the final piece in the \dataops puzzle. They enable practitioners to establish an iterative feedback loop, fostering continuous improvement in data-related operations.
%
\subsection{Proprietary Solutions}
\label{subsec:dataops:proprietaryTools}
This section is the counterpart of ~\autoref{subsec:dataops:openTools} focused on the proprietary tools implementing all the \dataops steps of Figure~\ref{fig:dataops_design} in a unique user-friendly solution.
These proprietary solutions provide dedicated customer support and additional features specifically tailored to address the unique needs of enterprises, a single point of contact for any issues or concerns, and extensive documentation and training resources facilitating the learning curve~\cite{boulanger2005open}.
However, it is important to consider potential drawbacks, such as licensing costs, potential vendor lock-in, and limited customization options compared to open-source solutions~\cite{caulkins2013make}. 

\myparagraph{Google Cloud Dataflow (GCD).}
GCD~\cite{GoogleDataFlow} is a powerful tool that shares functional similarities with Apache Spark and Apache Flink~\cite{carbone2015apache} while introducing its own distinctive features and architectural solutions. 
Similarly to its Apache counterparts, GCD has been meticulously designed to provide an efficient and low-latency approach to data manipulation while maximizing resource utilization. 
Contrarily to its Apache counterparts, it operates as a fully managed service on Google servers, integrating all the necessary components for building an efficient \dataops pipeline, which is a unique, easy-to-use solution.
It represents a valid solution for users from a non-programming background having a user-friendly graphical user interface.
GCD automatically provides cloud resources and automatically upscales or downscales the number of worker instances executing the job based on the data traffic.
However, GCD can not be served on-premise, limiting its application to certain applications, i.e., enterprises with stringent privacy needs. 

\myparagraph{Amazon Elastic MapReduce (EMR).}
EMR~\cite{amazonDataPlatform} is an Amazon cloud-based data processing service, featuring technologies and functionalities similar to the Apache suite (e.g., Apache Hadoop and Apache Spark). 
Like GCD, EMR is a fully managed framework committed to simplifying and streamlining the data processing path that automatically manages the scalability and efficiency of the \dataops pipeline for various data processing tasks. 
For example, the Amazon S3 (Simple Storage Service)~\cite{amazonCloudObject} is a data storage infrastructure, an alternative to HDFS, to retrieve and manage large amounts of data. 
Amazon Kinesis~\cite{amazonDataStream} is an Amazon service, an alternative to Apache Flink, designed to handle and process streams of large amounts of data. 
EMR offers fine-grained control over cluster size, allowing users to adjust the amount of data processed automatically and scale resources according to workload needs. 
EMR, as the GCP, can not be served on-premise.

\myparagraph{Azure Data Factory (ADF).}
ADF~\cite{microsoftAzureData} stands out as a robust cloud-based data integration service provided by Microsoft. 
Similarly to GCD and EMR, it offers a scalable and efficient cloud data processing solution, all while ensuring a user-friendly experience. 
This platform empowers users to collect data from various supported sources seamlessly and enables them to create, schedule, and manage data pipelines with the added benefit of automated workflow orchestration. 
Furthermore, it allows users to leverage the capabilities of the Apache Hadoop and Spark ecosystems within their data pipelines. 
This means that users can incorporate the power and versatility of these popular open-source frameworks directly into their workflows, enhancing the overall capabilities and efficiency of the data processing pipelines in the Azure Data Factory environment.
ADF, contrary to GCD and EMR, allows for an in-premise deployment of the platform.

\begin{table*}[t]
\color{black}
\centering
\renewcommand{\arraystretch}{1.25}
\setlength{\tabcolsep}{0.10cm}
\caption{Comparison of representative \dataops tools.}
\label{tab:dataops_tools_comparison}
\begin{tabular}{|p{2.15cm}||p{1.55cm}|p{2.2cm}|C{1.75cm}|C{1.25cm}|p{3.7cm}|p{3.7cm}|}
\hline
\hline
\textbf{Tool} & \textbf{Source} & \textbf{\dataops phase(s)} & \textbf{Processing} & \textbf{On-premise} & \textbf{Scalability} & \textbf{Limitation} \\
\hline
\hline

\multicolumn{7}{|l|}{\textbf{Open-source -- core data processing tools}} \\
\hline

Apache Kafka & Open-source & Ingestion, Monitoring & Stream & \tcheck & High throughput and low latency for large-scale streams & Requires expertise to configure, operate, and maintain distributed streaming clusters \\
\hline

Apache HDFS & Open-source & Storage & -- & \tcheck &  Horizontal scalability through data replication and node addition & Primarily designed for storage and batch-oriented workloads, requiring complementary tools for real-time analytics \\
\hline

Apache Spark & Open-source & Processing & Batch & \tcheck &  Distributed in-memory processing across multiple nodes, reducing disk access & Cluster configuration and performance tuning can be complex in production environments \\
\hline

Apache Flink & Open-source & Processing & Stream & \tcheck & Reduces storage overhead by processing continuous streams & Operationally more complex than batch-oriented frameworks and often requires strong stream-processing expertise \\
\hline

\multicolumn{7}{|l|}{\textbf{Open-source -- Ops tools, from release to feedback}} \\
\hline

dbt & Open-source & Release & -- & \tcheck &  -- & Focused on data transformation and testing, rather than complete pipeline orchestration \\
\hline

Docker + Kubernetes & Open-source & Deploy & -- & \tcheck & Horizontal scalability through container orchestration & Requires DevOps expertise and careful management of containerized infrastructures \\
\hline

Prometheus / Airflow / Jenkins & Open-source & Monitor, Feedback & -- & \tcheck & -- & Usually need to be integrated with other tools to cover the full \dataops lifecycle \\
\hline

Git / SVN & Open-source & Feedback & -- & \tcheck & -- & Support versioning and collaboration, but do not directly manage data processing or deployment \\
\hline

\multicolumn{7}{|l|}{\textbf{Proprietary -- end-to-end integrated platforms}} \\
\hline

Google Cloud Dataflow & Proprietary & End-to-end & Batch + Stream & \xmark & Auto-scaling through automatic worker instance provisioning & Limited infrastructure control and possible vendor lock-in within the Google Cloud ecosystem \\
\hline

Amazon EMR & Proprietary & End-to-end & Batch + Stream & \xmark & Automatic and fine-grained cluster sizing & Strong dependence on the AWS ecosystem and cloud-based deployment model \\
\hline

Azure Data Factory & Proprietary & End-to-end & Batch + Stream & \tcheck &  Cloud and on-premise hybrid scalability & Less customizable than self-managed open-source pipelines \\
\hline

Power BI / Tableau 
& Proprietary 
& Analysis, Visualization 
& --
& \tcheck
& Scalable dashboarding and reporting for large operational audiences 
& Focused on visualization and business intelligence, rather than complete data pipeline management \\
\hline
\hline
\end{tabular}
\end{table*}

\subsection{Solutions Comparison}
\label{subsec:dataops:comparison}
The choice between open-source and proprietary \dataops solutions in \ITLogisticShort depends on the specific application and on the company competencies, requirements, budget constraints, and levels of control and customization needed~\cite{caulkins2013make,jaisingh2008impact}.
\color{black}
To facilitate comparison across the reviewed tools, Table~\ref{tab:dataops_tools_comparison} summarizes open-source and proprietary \dataops solutions in terms of lifecycle coverage, deployment model, end-to-end support, scalability, and main limitations.
\color{black}

The open-source solutions (\autoref{subsec:dataops:openTools}), apart from being cost-effective, are surely the most popular choice for \dataops being reliable, interoperable, strongly customizable, and supported by a large and active community of developers~\cite{boulanger2005open}.
A common limitation of these platforms is the requirement for advanced skills in integration, deployment, and maintenance, which restricts their adoption in many contexts. 
\color{black}
Specifically, open-source tools are generally designed for specific purposes, requiring practitioners to combine multiple components to build an end-to-end \dataops pipeline.
\color{black}
For example, Apache Kafka is used for ingestion, Apache Spark for processing, Apache Hadoop HDFS for storage, Jenkins for analysis and visualization, dbt for release, Docker and Kubernetes for deployment, Apache Airflow for monitoring, and Git or SVN for feedback.

Proprietary solutions, on the other hand, provide user-friendly alternatives with enterprise-level support, enabling less specialized operators to deploy \dataops pipelines.
These solutions are fully integrated and designed around ``what the user wants to achieve'' rather than ``how it should be implemented''~\cite{sen2007strategic}.
Because of this fact, proprietary solutions are less customizable by design~\cite{lin2008impact}, even if interoperability and compatibility with open-source solutions are often guaranteed.
They scale very well both in terms of software, as the open source solutions, and in terms of hardware, as they offer both horizontal and vertical scalability in the cloud.
They have the limitation of being closed sources~\cite{boulanger2005open}, pushing toward a lock-in developing mode.
Another limitation is the trade-off between costs and benefits~\cite{caulkins2013make,jaisingh2008impact}, which is not always simple to estimate, taking into account also the problems of safety and security~\cite{boulanger2005open} and service continuity~\cite{kilamo2012proprietary}.

\subsection{Case Studies}
\label{subsec:dataops:casestudy}

This section presents real case studies in \ITLogisticShort that have implemented \dataops solutions.
These case studies encompass the {\color{black}6} areas of \ITLogisticShort presented in \autoref{sec:its-areas}. 
In the next paragraphs, we present the case studies retrieved for each area in the last 10 years, highlighting complementary examples that demonstrate particularly complex data management challenges.
Notably, in most of the presented cases, the \dataops solutions implemented leverage open-source Apache technologies, demonstrating their effective integration and adaptability for real-world scenarios.

\myparagraph{Vehicle Control Systems.}
In this area, we found a total of 11 cases, distributed among 
fleet management~\cite{alexakis2022distributed,kahveci2022end,dinten2023fleet}, decision support and making~\cite{zhang2017carstream,xu2016supporting,nair2016real,sbai2020real}, vehicle incident identification~\cite{triguero2017vehicle,ademola2022improving}, and driver support~\cite{yoo2020implementation,priyanka2021real}.
The platform of Yoo et al.~\cite{yoo2020implementation} aims to provide drivers with real-time vehicle and road condition information.
It collects and processes big data from connected cars, using a messaging system to handle large-capacity sensor data (e.g., OBU, CAN, camera, LiDAR, GPS) and traffic information.
The system, built on the Hadoop ecosystem (Hadoop, Spark, Kafka), stores vehicle data in HDFS and uses Apache Spark for big data analysis. 
Apache Kafka facilitates data transfer between the vehicle and server, while Spark Streaming enables real-time data analysis.
The DBDA platform described in \cite{alexakis2022distributed} combines various data management and analytics tools, such as MongoDB, Apache Hadoop, and Apache Spark, to ensure reliable and continuous data collection, storage, processing, and analysis of data generated by vehicles. Given the large volume of data collected, the authors highlight the importance of monitoring data quality. 
To address this, the DBDA platform integrates automatic data quality control procedures, including outlier and duplicate detection, to sanitize the ingested data.
Lastly, the CARSTREAM project~\cite{zhang2017carstream} integrates multiple services to support vehicle systems while monitoring the quality of the vast amount of collected data. 
The CARSTREAM architecture collects and processes various types of driving data, such as vehicle status, driver activity, and passenger-trip information. 
CARSTREAM is connected to over $30{,}000$ vehicles distributed across $60$ different cities in China, managing terabytes of driving data and providing data analytics-based services. 
With nearly a billion data instances collected daily, CARSTREAM requires an efficient data operations pipeline capable of managing data collected during peak hours and resisting potential faults. 
To achieve this, CARSTREAM integrates tools such as Apache Spark and Kafka for data streaming operations, and Apache Hadoop YARN for fault tolerance, processor isolation, and resource management.

\myparagraph{Traffic Management.}
This area is the one that collects the highest number of case studies in the literature. 
We found a total of 19 cases, distributed between real-time urban traffic~\cite{gillani2023machine,amini2017big,sinnott2015smash,anastasiou2019admsv2}, highway traffic detection and control~\cite{guerreiro2016architecture,figueiras2018data,daaif2019efficient,khazaei2016realtime,tseng2018congestion}, traffic prediction and simulation~\cite{khalajzadeh2020end,sundareswaran2020real,yu2020dissecting,fu2019demonstrating}, IoT smart transportation~\cite{babar2019real,tarneberg2016experiences,rathore2016exploiting,castellanos2021smart}, or air traffic flow~\cite{vonitsanos2021forecasting,chen2017modeling}.
Nevertheless, according to the best knowledge of the authors the most notable ones are the following.
Gillani et al.~\cite{gillani2023machine} implemented a real-time traffic data collection system city that utilizes Apache Kafka and Apache Spark to collect, process, and store data from approximately $400,000$ unique vehicles daily on Dublin roads, Ireland. 
Authors of~\cite{guerreiro2016architecture} integrate Apache Spark and Hadoop to develop a \dataops pipeline to optimize highway traffic flows in Portugal by adjusting toll prices during peak hours, thereby encouraging drivers to use less congested routes.
Amini et al.~\cite{amini2017big} discussed the integration of Apache Spark, HDFS, and Kafka to manage real-time traffic data from Munich's highway network. 
A similar architecture has been utilized by Babar et al.~\cite{babar2019real} to analyze traffic speed data in Aarhus city, Denmark, and within the SMASH project~\cite{sinnott2015smash} to process large-scale traffic data for Adelaide and Victoria cities, Australia.
Lastly, the BiDaML tool~\cite{khalajzadeh2020end} integrates Microsoft's Azure proprietary end-to-end architecture to ingest, process, store, visualize, and simulate traffic data in Victoria, Australia.

\myparagraph{Transportation Management Systems.}
In this area, we found a total of 14 cases, distributed mostly in urban transportation analysis and planning \cite{qi2020framework,ji2017low,leung2019urban,anastasiou2019admsv2,fiore2019integrated,rathore2021cyber,antunes2019big,luckner2020iot,
alic2019bigsea,tarneberg2016experiences,oneto2018train,vonitsanos2021forecasting,vo2022practical,mulerikkal2021jp}.
Among them, the BIGSEA~\cite{alic2019bigsea} and ADMSv2~\cite{anastasiou2019admsv2} projects are the most popular ones. 
BIGSEA is a \dataops solution for performance analysis and prediction of public transportation services. 
The pipeline, which utilizes Apache Spark and HDFS, offers a scalable data management solution for European and Brazilian public transportation systems.
The BIGSEA architecture has been later utilized and integrated by Fiore et al.~\cite{fiore2019integrated} to tackle urban traffic data analysis and planning challenges of the municipality of Curitiba, a 1.8 million-inhabitants city in the south of Brazil.
Qi et al.~\cite{qi2020framework} offer a custom framework with efficient extraction and analysis of Twitter data for evaluating public opinions on transportation services.
Despite the above example implementing only some stages, the ADMSv2 project~\cite{anastasiou2019admsv2} offers an end-to-end \dataops architecture to handle a variety of data sources, including massive and high-resolution traffic sensor data, bus trajectory data, transportation network data, and traffic incidents data from the Los Angeles Metropolitan Transportation Authority. 
It integrates several open-source frameworks to ingest (Apache Kafka), store (Apache Hadoop HDFS), manage (Apache Spark), and analyze data with custom web interfaces to efficiently support urban planning, public policy, and emergency response.
T{\"a}rneberg et al.~\cite{tarneberg2016experiences} utilize cloud-based IoT infrastructure in AWS to integrate multiple sensors and source data in Charlottesville, VA roads to regulate the red-green patterns of traffic lights with the goal of increasing safety and minimizing wait/idle times.
We also found some work around transportation management of metro, trains, or flights~\cite{oneto2018train,vonitsanos2021forecasting,vo2022practical,mulerikkal2021jp} where the practical goal is to construct a \dataops pipeline for transportation delay forecasting and planning.

\myparagraph{Traveler Information and Experience Systems.}
In this area, we found a total of 9 cases, all focusing on forecasting and recommendation systems for tourists~\cite{al2021hybrid,chi2021design,asaithambi2023thematic,lapatta2022ecotourism,ambolkar2022hotel,malkawi2023real,bellandi2019designing,petrozziello2017data,suanpang2020integration}.
These applications aim to enhance travelers' experiences by providing accessible information from multiple sources. They face the challenge of integrating various data sources and delivering real-time information to users.
For instance, the platform in \cite{chi2021design} supports travel and hotel management planning by utilizing Hadoop and Spark to analyze hotel data and user comments.
Malkawi et al.~\cite{malkawi2023real} propose a web-based real-time ticket recommendation system that uses Apache Spark to retrieve information from various airline websites, allowing users to receive recommendations based on the lowest prices. 
Lastly, \cite{suanpang2020integration} integrates big data tools (i.e., Apache Spark and Apache Hadoop) in an iOS mobile app to provide users with essential information about tourist attractions and accommodations in Lampang Province, Thailand. 

\color{black}
\myparagraph{Freight and Supply Chain Logistics.}
In this area, we identified 14 cases, covering both the monitoring and planning of goods delivery~\cite{kumar2018design,Hadavi2019MonitoringUT,dimokas2021cloud,belfadel2023conceptual,selvaraj2020offline,provatas2020selis,sarabia2019seaport,silva2021advancing,ruan2021filling,lee2018design,sugathan2021espy,hong2019crowdsourcing} and the optimization of freight management policies, such as pollution-emission reduction~\cite{mhand2019scalable,de2021simulation}.
The architecture proposed by Selvaraj et al.~\cite{selvaraj2020offline} combines Apache Hadoop, Hive, and Spark into a unified data-operations solution for truck management companies, supporting logistics optimization, cost and overhead reduction, profit-margin improvement, and load balancing.
Provatas et al.~\cite{provatas2020selis} develop a data-operations framework for four distinct logistics case studies, including goods delivery by trucks for a Greek logistics company. The architecture uses Hadoop YARN for cluster and resource management, thereby supporting scalability under variable user demand, and Apache HBase to store event logs subsequently processed with Apache Spark.
Sarabia et al.~\cite{sarabia2019seaport} examine the Valencia seaport, one of the major ports on the Mediterranean coast, which handles millions of movements per year. To process large volumes of data within short time windows and support communication among seaport stakeholders, the proposed architecture integrates several open-source data-management tools, including Apache Hadoop and PostGIS for efficient storage, Apache Spark for streaming operations, and Apache Kafka for data ingestion and processing. The same architecture also supports warehousing and delivery within the seaport.

On the supply-chain and delivery side, Silva et al.~\cite{silva2021advancing} propose a framework for optimizing supply-chain management, demonstrated through a real-world case study in the logistics innovation department of an automotive factory generating large volumes of data. Their architecture integrates Apache Spark for data cleansing and transformation and Hadoop HDFS for distributed storage. Moreover, proprietary tools such as Power BI~\cite{powerBI} and Tableau~\cite{tableau} are used for data visualization, providing interactive dashboards that present analysis results to end users.
Finally, eSPY Carting Systems~\cite{sugathan2021espy} is a real-time tracking system designed to anticipate potential faults in products and services, thereby improving the reliability of goods transport and delivery. The project integrates several big-data technologies, including Apache Spark and Hadoop for data processing and storage, Apache Kafka for data ingestion, and Apache Hive for metadata management.

Although additional studies in this area exist, such as~\cite{lee2018design,hong2019crowdsourcing}, they often do not explicitly address big-data challenges or discuss the data tools employed. This indicates an opportunity for further exploration and documentation of \dataops applications across the entire supply-chain spectrum.
\color{black}

\myparagraph{Emergency Management.}
In this area, we found a total of 10 cases, distributed between traffic event detection and response (e.g., traffic, congestion, etc.)~\cite{alomari2019road,anastasiou2019admsv2,pandhare2017real,alomari2021iktishaf,seal2019real} and public health safety logistic control and planning \cite{kovalchuk2018distributed,rathore2016real,alotaibi2020sehaa,noori2020fire,huamani2020machine}.
Regarding the former family, we mention the work by Alomari et al.~\cite{alomari2019road}, which integrates Apache Spark with ML algorithms for the automatic detection of road traffic events from tweets in Saudi Arabia. 
The framework SEHAA~\cite{alotaibi2020sehaa} combines Apache Spark to collect and process huge quantities of data from Twitter with ML models to assist healthcare services in Saudi Arabia.
The ADMSv2 project~\cite{anastasiou2019admsv2}, also seen among transportation management systems case studies, further supports emergency management operations efficiently by continuously ingesting and processing vast amounts of data using Apache data operations solutions.
For public health safety logistic control and planning, we note several studies~\cite{kovalchuk2018distributed,rathore2016real,alotaibi2020sehaa,noori2020fire,huamani2020machine}.
We also highlight the case study by Noori et al.~\cite{noori2020fire}, which combines Apache Spark, Twitter data, and ML models for timely fire emergency detection.
In these applications, the use of \dataops solutions such as Apache Spark is justified by the large scale of data collected (from social media or IoT devices) and the requirement for fast processing to enable rapid decision-making.

\color{black}
\myparagraph{Remarks.} 
Overall, the reviewed case studies indicate that \dataops pipelines address recurring data management challenges across \ITLogisticShort domains, although the relative importance of these challenges differs across application areas.
Vehicle Control Systems and Traffic Management are primarily driven by the need to ingest and process high-volume, high-velocity data streams generated by vehicles, sensors, and road infrastructure.
Transportation Management Systems and Freight and Supply Chain Logistics require stronger support for data integration, as they combine geographically distributed and heterogeneous sources, including transit records, operational logs, seaport data, supply-chain events, and user-generated information.
Traveler Information and Experience Systems place greater emphasis on continuous data integration and service responsiveness, since recommendations and travel information must be continuously updated from multiple external sources.
Emergency Management applications place the strongest emphasis on low-latency processing, as timely event detection and response depend on the rapid analysis of social media, IoT, and traffic-related data.
Across these domains, Apache-based technologies such as Kafka, Spark, and Hadoop recur frequently, confirming their practical relevance for scalable ingestion, storage, stream processing, batch analytics, metadata management, and resource orchestration in real-world \dataops deployments.
\color{black}

\section{\mlops}
\label{sec:mlops}
AI, particularly in the form of ML, is a major catalyst for innovation in products and services within \ITLogisticShort applications. 
ML empowers the formulation of descriptive~\cite{roy2022complete}, diagnostic~\cite{roy2022complete}, predictive~\cite{roy2022complete}, and prescriptive analytics~\cite{roy2022complete}, enabling the monitoring, improvement, and innovation of new products and services~\cite{kibria2018big}. 
The landscape of ML can be broadly divided into two major categories: Shallow ML and Deep ML. 
Shallow ML can handle unstructured, tabular data with small to medium cardinality, leveraging effective manual feature extraction and ML model design based on domain knowledge~\cite{gorishniy2021revisiting}. 
In contrast, Deep ML is mandatory when dealing with medium to huge cardinality unstructured data, such as images, graphs, and natural language~\cite{cao2016deep,li2020survey}.
It enables the direct learning of well-suited representations from the raw data~\cite{shaheen2016impact,najafabadi2015deep}.
The building phase is commonly referred to as the training phase, while the usage of the model is termed the test phase.
The training phase can be computationally demanding, especially for deep ML, often necessitating specific hardware accelerators (i.e., GPUs). 
In contrast, the test phase is typically less computationally demanding, even with substantial data quantities.

\begin{figure}
\centering
\includegraphics[width=\columnwidth]{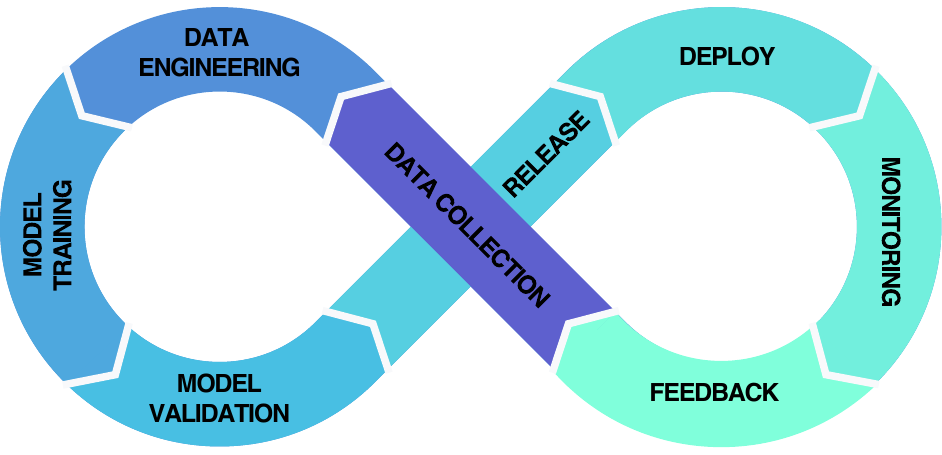}
\caption{\mlops abstract lifecycle.}
\label{fig:mlops_design}
\end{figure}

While ML can be conceptually viewed as a specific block within the \dataops pipeline, its unique characteristics necessitate a dedicated pipeline known as Machine Learning Operations (\mlops)~\cite{faubel2023mlops,symeonidis2022mlops}. 
This distinction arises from several reasons: (i) continuous update requirements, (ii) management of design complexity, (iii) management of deployment complexity, and (iv) monitoring challenges. 
\mlops, by adapting and extending DevOps principles for ML, facilitates the retrieval of training data, exploration, feature extraction, model training, testing, release, deployment, monitoring, and feedback processes with a high degree of automation (see Figure~\ref{fig:mlops_design}).
\mlops offers a promising response to these challenges in the \ITLogisticShort sector. 
It integrates practices and technologies to streamline the entire ML lifecycle, from problem data collection to model training and testing, ensuring efficient processing and continuous model availability in the deployment service~\cite{Testi2022MLOpsAT}.
\color{black}
In the end-to-end view presented in Figure~\ref{fig:dataops_mlops_tai_framework}, this section corresponds to the \mlops layer, which builds on reliable \dataops pipelines to manage the development, deployment, monitoring, and updating of ML models in operational \ITLogisticShort services.
This layer transforms reusable data assets into AI-enabled decisions and services, while providing the mechanisms required to monitor model behavior and adapt models as data distributions and operational conditions evolve.
\color{black}

In the following, we start by revisiting the underlying motivations for embracing \mlops, as detailed in~\autoref{subsec:mlops:Motivations}.
Following that, we describe the various stages encompassed within an \mlops pipeline, as outlined in~\autoref{subsec:mlops:Life}.
Subsequently, we examine the principal open-source (\autoref{subsec:mlops:openTools}) and proprietary \mlops tools (\autoref{subsec:mlops:proprietaryTools}). 
A comparative analysis of these tools is presented in~\autoref{subsec:mlops:comparison}, offering practical insights and implications.
The section then delves into the case studies of the \ITLogisticShort domain that implement \mlops technologies, presented in~\autoref{subsec:mlops:casestudy}. 
This section concludes with the discussion of a specific branch of the \mlops called \llmops, namely \mlops for large (multimodal) language models, which surely represent one of the highest impact and promising technologies of the last decade.
In particular, we will deepen the necessity of this branch of \mlops and present some related case studies inside the \ITLogisticShort.

\subsection{Motivations}
\label{subsec:mlops:Motivations}
The adoption of \mlops practices has been driven by four fundamental requirements.

\myparagraph{Continuous Update.}
Data usually dynamically evolve in the \ITLogisticShort industry, demanding a continuous model update to address this data shift and provide accurate real-time decision-making~\cite{raj2021edge,symeonidis2022mlops,pineda2023mlops}. 
Consider, for example, a scenario where a logistics company utilizes an ML model for route optimization. 
The dynamic nature of traffic conditions, road closures, and unforeseen events requires the model to continuously adapt to these changes to incorporate the latest information, ensuring that the recommended routes remain efficient and accurate~\cite{koh2020real}. 
Moreover, ML technology is also a fast-evolving field where new, more effective, or efficient models (e.g., convolutions or transformers) are developed every year, requiring, even in slow changes in the data, to advance the ML technologies continuously~\cite{menghani2023efficient}. 
Without continuous model updates, the system may rely on outdated data or ML architectures, leading to sub-optimal route recommendations and potentially impacting the timely delivery of goods~\cite{Liu2019IntelligentRP,koh2020real}.
However, model update involves a sequence of non-trivial operations, such as tuning or retraining, model versioning, functionality testing, deployment, log monitoring, etc. 
If performed manually, these operations could slow down the production process, potentially causing delays in bringing the upgraded service online~\cite{raj2021edge,faubel2023mlops}. 
\mlops facilitates the implementation of a responsiveness and automatic ML pipeline that processes and integrates novel knowledge arising from data collected over time~\cite{pineda2023mlops,Testi2022MLOpsAT} and novel ML models developed by researchers~\cite{kreuzberger2023machine}.

\myparagraph{Design Complexity.}
The deployment of ML models introduces intricacies that surpass those encountered in deploying traditional software modules. 
Unlike conventional software, ML systems combine input signals, creating entanglements that challenge the isolation of improvements~\cite{sculley2015hidden}, a fundamental principle in software testing and DevOps methodologies. 
A change in the input distribution, training algorithm, data processing, or any aspect of the ML development can lead to several changes to the whole pipeline~\cite{sculley2015hidden}. 
Additionally, ML is distinguished by numerous free parameters, known as hyperparameters, which pose a challenge as these cannot be pre-set and require meticulous tuning whenever environmental changes occur. 
Furthermore, the ML technology landscape, as outlined in the 2023 ML AI Data Landscape Report\footnote{\url{https://www.lxahub.com/stories/key-takeaways-from-the-2023-ml-ai-data-landscape-report}}, is characterized by substantial fragmentation in terms of both code quantity and quality. 
Managing this diversity presents a challenge, as a single ML project may require dozens of different libraries, each with varying levels of reliability~\cite{sculley2015hidden}. 
Some libraries may become discontinued or unmaintained, requiring replacement or patching, while others may transition from open to closed-source, demanding careful monitoring of licenses and possible library changes to maintain the project's reliance on open-source tools.
In summary, the dynamic nature of ML software modules and the landscape demands continuous monitoring and adaptation to keep pace with evolving technology.
This includes managing the interaction between traditional and ML-based software modules, tracking licenses, and being prepared to switch to alternatives. 
\mlops naively provides support for the solution of these problems.

\myparagraph{Model Monitoring.}
Accurately monitoring ML-based model operations surely leads to several advantages, but, compared to traditional software, it raises many different criticalities~\cite{sculley2015hidden,silva2020benchmarking,banerjee2020challenges}.
Monitoring ML models implies monitoring two main aspects: input data and ML model results~\cite{symeonidis2022mlops}. 
Input data may contain outliers or drifts, e.g., with respect to the training data, requiring detection (in the case of outliers) or retraining the models (in the case of drifts).
Second, tracking ML model results and behavior provides valuable insights into how well the model aligns with the current data distribution and operational context~\cite{silva2020benchmarking}. 
However, ML models are usually designed to optimize one or more metrics.
For these reasons, multiple metrics describing the average behavior of the model need to be exploited simultaneously~\cite{zeller2024toward,banerjee2020challenges}. 
These metrics can be technical (e.g., accuracy, robustness, energy, etc.) but they may be also ethical (e.g., explainability, fairness, privacy, etc.)~\cite{zeller2024toward}.
This requirement is becoming more and more vital also in view of the recent AI-related regulations (e.g., AI Act~\cite{ai-act-council,eu2021ai-act,european_ai_act_2021}).
Even if multiple metrics are continuously monitored, security issues may still arise due to the vulnerability of ML models that need to be readily patched.
Finally, ML models are often non-deterministic both in the training (the training data may produce different ML models) and in the test phase (the same test data may produce different ML model outputs).
\mlops can significantly contribute to the ML model performance evaluation, enabling proactive interventions and adjustments to ensure alignment with multiple technical and ethical metrics.
%

%
\myparagraph{Complexity of ML Deployment.} 
Two main factors influence the complexity of deploying ML models.
The first is reproducibility~\cite{sculley2015hidden}, and the second is resource allocation~\cite{sculley2015hidden}.
Reproducibility, defined as the ability to repeat experiments and obtain the same (or at least similar) results~\cite{haibe2020transparency}, is essential to increase transparency~\cite{haibe2020transparency} and build user confidence in ML applications and their components~\cite{adkins2022method}.
However, the non-deterministic nature of many ML models and the need for continuous updates undermine reproducibility~\cite{sculley2015hidden}.
Resources allocation, instead, deals with the problem of retraining ML models and deploying them automatically while ensuring the scalability of the entire system and maintaining service availability for users. 
During the training phase, the system must manage fewer jobs, potentially requiring the adoption of on-demand distributed computing solutions~\cite{silva2020benchmarking}. 
These jobs typically involve training or updating a few hundred models and demand significant computational power.
Conversely, during the testing phase, the system must handle a large volume of requests, often proportional to the number of users multiplied by the frequency of requests for the service. 
While these individual requests may have relatively modest computational requirements, the system must efficiently process and respond to many without compromising service quality or responsiveness. 
\mlops supports both reproducibility and traceability throughout the entire ML life cycle together with an instrument able to fully automatize the deployment both for training and test phases.
\subsection{\mlops Lifecycle}
\label{subsec:mlops:Life}
The \mlops lifecycle defines an iterative workflow for the continuous refinement, adaptation, monitoring, and improvement of ML processes to meet evolving application requirements and challenges~\cite{kreuzberger2023machine,Testi2022MLOpsAT,alla2021mlops}.
Hence, a comprehensive \mlops process is depicted in \autoref{fig:mlops_design} to embody DevOps principles within the ML domain.
This proposed framework distills the \mlops lifecycle into eight well-defined stages: data collection, data engineering, training, model testing, release, deployment, monitoring, and feedback.
 
\myparagraph{Data Collection.}
The \mlops pipeline starts with the data collection tailored to meet specific application requirements.
Due to the large scale of data commonly employed, particularly in training Deep ML models, integrating \dataops solutions becomes fundamental to streamlining the entire process. 
Specifically, after the data is stored within a \dataops pipeline, it can be seamlessly retrieved for ML model training and test phases.
This stage aligns with the \textit{Plan} stage in DevOps~\cite{sen2021devops}.

\myparagraph{Data Engineering.}
Data engineering includes data cleaning and feature preparation, often identified as among the most time-consuming activities in the workflow~\cite{chicco2022eleven}.
While data cleaning, i.e., improving data quality, can be part of both \dataops or \mlops, feature engineering is primarily concerned with \mlops.
In fact, feature engineering deals with the problem of transforming raw data into meaningful information that can be handled by ML algorithms.
For tabular or mildly unstructured data, the feature engineering phase is commonly time-consuming as it is manually designed by domain experts or ML practitioners~\cite{chicco2022eleven}.
The resulting features are then fed directly to shallow ML algorithms.
For unstructured data, these features cannot be manually designed and must be learned directly from the data (e.g., using convolutions or attention mechanisms).
However, while the latter allows to remove the human intervention, making the process more automatic, it requires huge amount of data to be performed.
Finally, data and the corresponding extracted features are split into multiple splits, typically clarified as training and validation, used during training (see Model Training in \mlops) and validation (see Model Validation in \mlops), respectively.
This stage aligns with the Code phase in DevOps~\cite{sen2021devops}, where developers write and integrate code for new features or improvements.

\myparagraph{Model Training.}
Model training includes both model engineering and model fitting.
The former involves defining and identifying, at higher level, relevant ML algorithms (e.g., XGBoost~\cite{chen2016xgboost} or Deep Neural Networks~\cite{najafabadi2015deep}) with the associated possible architectures (e.g., convolution~\cite{li2021survey} or attentions~\cite{vaswani2017attention}) and possible set of hyperparameters (e.g., number of weak models~\cite{chen2016xgboost} or size of the convolution kernels~\cite{li2021survey})~\cite{faubel2023mlops}.
In other words, model engineering defines the set of possible ML algorithm configurations to be trained and subsequently evaluated to choose the best options (see Model Validation in \mlops).
This phase is mainly guided by experience, data type, business and user requirements, and computational resources.
Experience, i.e., human in the loop, plays a central role because the sheer number of choices in the ML algorithms makes a thorough evaluation of each one impractical. 
Therefore, past experience becomes invaluable in guiding the decision-making process.
The data type determines whether simpler and shallower ML approaches or more complex and deeper ML techniques are appropriate.
Businesses and users provide perspectives on the potential commercial value, considering costs and benefits and users' expectations on the quality of service provided by the ML model.
Lastly, factors such as available computing power and infrastructure influence decisions, such as using pre-trained models or adopting end-to-end training approaches, narrowing down the available options~\cite{silva2020benchmarking}.

Given a specific model engineered in the previous phase, the model fitting phase focuses on optimizing its parameters using the provided training data samples.
This optimization process involves adjusting the model's parameters to minimize a predefined loss function, which measures the discrepancy between the model's predictions and the actual targets in the training dataset~\cite{shalev2014understanding}.
This process involves iterative updates to the model's parameters using gradient-based optimization algorithms~\cite{shalev2014understanding}. 
Furthermore, the choice of the model architecture and the optimization algorithm -- together with its hyperparameters like learning rate and momentum -- can significantly impact the convergence and effectiveness of the fitting process.
Shallow ML models, such as ridge or logistic regression, typically lead to convex optimization problems with efficient convergence to global minima.
In contrast, Deep ML models navigate non-convex optimization landscapes where local minima are found during training.

This stage aligns with the Build phase in DevOps~\cite{sen2021devops}, where developers refine ML models to meet the demands of the operational environment. 

\myparagraph{Model Validation.}
Model validation deals with the remaining choices left to be made during the Model Training phase of \mlops.
Specifically, during the model training phase, candidate models are engineered and trained to reduce the error on the training dataset. 
The model validation phase focuses on selecting the one that provides the best compromise in terms of multiple metrics (e.g., technical, ethical, or business) on a validation dataset to avoid data-snooping~\cite{white2000reality}.
However, the challenge lies in finding the proper set of metrics for the specific application and striking a balance among them, as they often conflict with each other.
Technical metrics aim to achieve optimal results with minimal resource utilization, while legislative frameworks like the AI-Act mandate adherence to specific ethical requirements (i.e., explainability, fairness, and robustness) to mitigate both intended and unintended harm. 
Additionally, business metrics aim to optimize user satisfaction while maximizing company gains.
Finally, training and model evaluation are closely interconnected; if model evaluation produces poor results, model training must be adjusted accordingly. 
Conversely, if the model evaluation produces excellent results (e.g., performance that is much above the one requested by the application), the model complexity may be reduced (e.g., from deep to shallow model) to decrease the computational requirements and then increase the business profit.
The validation phase of the \mlops pipeline aligns with the test phase of DevOps~\cite{sen2021devops}. Here, developers integrate \mlops tools and techniques to validate the development and correctness of the models that meet desirable technical, ethical, and business requirements.

\myparagraph{Release, Deploy, Monitoring, and Feedback.}
The final stages of the \mlops pipeline, similar to DevOps and \dataops, deal with operational and automated activities related to ML model release, deployment, monitoring, and feedback.
During the release phase, the focus is on ensuring the consistency of the model and verifying its functionality for distribution, i.e., verifying that software dependencies are aligned with the deployment environment. 
Afterward, the model is deployed in the production environment and automatically scaled to ensure system availability. 
Then, monitoring mechanisms are put in place to monitor the performance of the deployed model in real time, using a subset, the same, or a superset of the technical, ethical, and business metrics defined in the Model Validation, and collect feedback~\cite{zhang2018long}.
This allows operators to check for abrupt deviation in real-time operativity of the model behavior due to changes in the data or in how we evaluate the model~\cite{zhang2018long,symeonidis2022mlops}.
Finally, the feedback loop in \mlops involves gathering information from the monitoring phase, facilitating continuous improvement and refinement over time while ensuring the ML model effectiveness and reliability~\cite{symeonidis2022mlops}.

\subsection{Open-Source Solutions}
\label{subsec:mlops:openTools}
This section, similar to Section~\ref{subsec:dataops:Life} for \dataops, reviews the most popular open-source tools that offer free access to the implementation of the \mlops pipeline, making them cost-effective options for project development.
These tools allow developers to build and customize their ML workflows while enabling them to integrate other third-party libraries (e.g., PyTorch, Keras, etc.). 
Furthermore, given their open-source nature, they benefit from the support of the open-source community, documentation, and discussion groups. 
It is worth noting that a single tool often only implements some of the components of the \mlops pipeline.
Rather, developers usually combine several tools to obtain a complete pipeline.

\myparagraph{Metaflow.} 
Developed by Netflix, Metaflow~\cite{metaflowMetaflowFramework} is an efficient tool to implement the \mlops pipeline in data-intensive applications. 
Specifically, it facilitates feature engineering, model building, training, validation, and deployment for large-scale ML projects demanding scalability and productivity. 
With over $8$ thousand stars on GitHub as of January 2025, it offers a user-friendly interface and the adaptability to meet the requirements of both prototyping and production stages.
Metaflow excels in integrating with various on-demand cloud computing services (e.g., Amazon AWS~\cite{amazonCloudComputing}). 

\myparagraph{ZenML.} 
ZenML~\cite{zenmlZenMLMLOps} is the main MetaFlow alternative with over $4$ thousand stars on GitHub as of January 2025. 
The peculiarity of the ZenML pipelines is in their structured approach, which treats each step of the ML workflow as an individual and reusable component. 
This design philosophy promotes reproducibility and consistency across diverse environments.
Additionally, ZenML integrates advanced features for experiment tracking, data versioning, and pipeline monitoring with well-known third-party tools (e.g., Weights and Biases and MLFlow Tracking). 
These integrated tools empower developers to efficiently track experiments, version and iterate on datasets, evaluate model performance, reproduce and save models, monitor ML runtime behavior, and perform various other tasks, all from a unified dashboard interface.

\myparagraph{BentoML.}
BentoML~\cite{bentomlBentoMLUnified}, introduced in 2019 and claiming over $7$ thousand stars in GitHub as of January 2025, is an open-source tool meticulously conceived to simplify and accelerate the release and deployment of ML models, not implementing the entire \mlops pipeline.
This tool offers compatibility with several ML frameworks, such as TensorFlow and PyTorch, as well as seamless integration capabilities with \dataops systems (e.g., Apache Airflow~\cite{apacheAirflow}). 
BentoML guarantees efficient scalability and monitoring of model deployments, providing a robust infrastructure for real-time model management and monitoring. 
Furthermore, it provides a user-friendly dashboard that simplifies the management and monitoring of ML models, providing an intuitive interface for users. 
Finally, it supports various deployment methods, including Docker~\cite{dockerHome}, Kubernetes~\cite{kubernetesProduction} with Yatai and integration with various on-demand cloud services (e.g., Amazon AWS) through Bentoctl. 
BentoML is distinguished by several key features that integrate tightly with other \mlops tools (e.g., ZenML, MetaFlow), improving and accelerating large-scale deployment operations. 

\myparagraph{Kubeflow.}
Kubeflow~\cite{kubeflowKubeflow} is an open-source tool, claiming over $14$ thousand stars in GitHub as of January 2025, for integrating ML capabilities within the Kubernetes ecosystem. 
It leverages the robust Kubernetes infrastructure to manage intricate ML workflows such as training on distributed nodes.
Compared to BentoML, which offers versatility among various ML frameworks and integration with \dataops systems, Kubeflow specifically targets Kubernetes environments and deployment operations, prioritizing large-scale system reproducibility. 
By leveraging Kubeflow, users can orchestrate sophisticated ML pipelines, efficiently manage experiments, and seamlessly deploy and serve models in a Kubernetes environment. 
This approach improves scalability and resource management and promotes consistency and reproducibility of ML workflows, aligning with industry standards and best practices. 
Furthermore, Kubeflow's emphasis on integration with Kubernetes ensures compatibility and seamless deployment across different on-demand cloud services, further increasing the attractiveness for projects seeking to leverage Kubernetes for their ML operations.

\color{black}
\myparagraph{Apollo.}
Apollo~\cite{apolloMLOps}, developed by Baidu and reporting more than $24$ thousand GitHub stars as of January 2025, is an open-source autonomous-driving platform that provides an integrated operational environment for deploying vehicle control applications.
Within the \mlops lifecycle, Apollo primarily supports model development, simulation-based validation, deployment, and runtime execution for safety-critical \ITLogisticShort pipelines.
Unlike general-purpose \mlops tools, Apollo is tailored to vehicle control applications, providing \ITLogisticShort practitioners with a domain-specific operational stack for autonomous-driving systems.
\color{black}

\subsection{Proprietary Solutions}
\label{subsec:mlops:proprietaryTools}
Proprietary \mlops solutions implement all the \mlops steps depicted in Figure~\ref{fig:mlops_design}, providing an end-to-end framework for model development and deployment.
These \mlops services are commonly provided by cloud vendors~\cite{raj2021edge} who offer on-demand computational resources and data and model storage, facilitating the training and versioning of ML models.
Such a comprehensive suite of services simplifies the entire ML lifecycle, from development to deployment, providing organizations with the infrastructure and tools needed to deploy and scale ML solutions effectively~\cite{raj2021edge}.

\myparagraph{VertexAI.}
VertexAI ~\cite{googleVertexPlatform} is a proprietary platform designed by Google to facilitate the training, deployment, and customization of ML models and ML-based applications.
The platform offers various options for data preparation, model training, validation, deployment, and monitoring.
One key advantage of VertexAI is its simplicity in developing ML projects with user-friendly, accessible, and complete dashboards (it can be used as a zero-code platform).
VertexAI, for example, integrates AutoML, a service that enables ML model training on tabular, image, text, or video data without the need for coding or data preparation.
Alternatively, custom training provides complete control over the training process, including using ML frameworks and hyperparameter tuning options.
After ML model deployment, VertexAI offers end-to-end \mlops tools to automate and scale projects throughout its lifecycle.
These tools run on fully managed infrastructure that can be customized based on performance needs and budget availability.
The platform also offers services for managing model versions and monitoring model quality and trustworthiness in production environments.

\myparagraph{IBM Watson Studio.}
IBM Watson Studio ~\cite{ibmWatsonStudio} simplifies the development, training, and deployment of ML models, supporting multiple data sources. 
Akin to VertexAI, Watson Studio integrates AutoML capabilities, automating model pipeline creation, data pre-processing, model selection, and performance evaluation with new data.
Users can efficiently manage models throughout their lifecycle by utilizing advanced capabilities such as AutoML and model monitoring.
Combining open-source frameworks like PyTorch, TensorFlow, and scikit-learn with IBM tools, Watson Studio supports visual and code-based data science through Jupyter Notebook, JupyterLab, and command-line interfaces in Python, R, and Scala.
Two aspects of \mlops development distinguish Watson Studio.
First, it enables deployment on IBM cloud services (as for VertexAI) or on-premise.
The latter becomes particularly interesting for companies that retain data and model privacy.
Secondly, it provides reliable native metrics to monitor data accuracy, correctness, and drift, culminating in risk management scores to meet regulatory compliance and other risk objectives effectively.

\myparagraph{Microsoft Azure Machine Learning and Amazon Sage Maker.}
Microsoft Azure Machine Learning~\cite{microsoftAzureMachine} and Amazon SageMaker~\cite{amazonsagemaker}, developed by Microsoft and Amazon, respectively, are solid alternatives to Google VertexAI.
Both platforms facilitate the efficient and user-friendly development of large-scale \mlops pipelines by supporting popular programming languages (e.g., Python) and libraries (e.g., PyTorch).
Azure ML features seamless integration with popular third-party tools such as BigQuery and Spark for efficient data processing and analysis. 
On the other hand, Amazon SageMaker seamlessly interacts with Amazon Elastic MapReduce to efficiently manage the data used in the \mlops pipeline.
Like IBM Watson Studio, Azure ML enables on-premise usage of their services to promote data and model privacy if users or companies require it. 

\myparagraph{MATLAB and Simulink.}
Developed by MathWorks Inc., MATLAB~\cite{matlab} and Simulink~\cite{simulink} are two systems that enable users and companies to build an end-to-end pipeline of \mlops within their cloud architecture.
The AutoML service has also been integrated to automatically find the best ML model and optimal hyperparameters for a given training data set. 
The AutoML component uses MATLAB Parallel Server to accelerate this process by simultaneously training and evaluating multiple models.
In particular, the MATLAB \mlops system integrates drift detection applications that compare the values of newly observed data with those of the model's training set.
In production, it can read the observed data in near real-time from an Apache Kafka stream, detect anomalies in the new data, and automatically create a new ML model optimized for the cleaned data.

\begin{table*}[t]
\color{black}
\centering
\renewcommand{\arraystretch}{1.25}
\setlength{\tabcolsep}{0.10cm}
\caption{Comparison of representative \mlops tools. }
\label{tab:mlops_tools_comparison}
\begin{tabular}{|p{2.5cm}||p{2.5cm}|C{1cm}|C{1.25cm}|p{4.7cm}|p{4.7cm}|}
\hline
\hline
\textbf{Tool} & \textbf{\mlops phase(s)} & 
\textbf{AutoML} & \textbf{On-premise} &  
\textbf{Scalability} & \textbf{Limitation} \\
\hline
\hline

\multicolumn{6}{|l|}{\textbf{Open-source -- core ML pipeline tools}} \\
\hline

Metaflow 
& Data engineering, training, validation, deployment 
& \xmark 
& \tcheck 
& On-demand cloud integration for large-scale workloads 
& Does not cover the full \mlops pipeline; monitoring and feedback stages require additional tools \\
\hline

ZenML 
& End-to-end
& \xmark 
& \tcheck 
& Via third-party integrations with cloud and on-premise services 
& Relies on external tools, such as MLflow, Weights and Biases, and BentoML, to cover several pipeline stages, adding integration overhead \\
\hline

BentoML 
& Release, packaging, deployment, serving, monitoring support 
& \xmark 
& \tcheck 
& Docker, Kubernetes, and cloud services via Bentoctl 
& Focuses mainly on release, deployment, and serving rather than on the full training and validation lifecycle \\
\hline

Kubeflow 
& End-to-end
& \tcheck 
& \tcheck 
& Distributed execution across Kubernetes clusters 
& Tightly coupled to Kubernetes infrastructure, requiring significant DevOps expertise to deploy and maintain \\
\hline

Apollo 
& Data engineering, validation, deployment, monitoring 
& \xmark 
& \tcheck 
& Simulation-based testing environments for vehicle control applications 
& Not designed for general ML workflows domains beyond vehicle control \\
\hline

\multicolumn{6}{|l|}{\textbf{Proprietary -- end-to-end integrated platforms}} \\
\hline

VertexAI 
& End-to-end 
& \tcheck 
& \xmark 
& Fully managed auto-scaling infrastructure 
& No on-premise deployment; monitoring and governance remain platform-dependent; Google Cloud vendor lock-in \\
\hline

IBM Watson Studio 
& End-to-end 
& \tcheck 
& \tcheck 
& Cloud and on-premise hybrid 
& Native governance and risk monitoring are useful but do not exhaustively cover all trustworthiness dimensions, such as robustness and privacy~\cite{zeller2024toward} \\
\hline

Azure Machine Learning 
& End-to-end 
& \tcheck 
& \tcheck 
& Cloud and hybrid deployment with managed compute 
& Requires integration within the Microsoft Azure ecosystem; limited out-of-the-box \TAI monitoring \\
\hline

Amazon SageMaker 
& End-to-end 
& \tcheck 
& \xmark 
& Managed training, deployment, endpoints, and integration with AWS data services 
& Strong AWS ecosystem dependency; no native on-premise deployment option \\
\hline

MATLAB / Simulink 
& End-to-end 
& \tcheck 
& \tcheck 
& Parallel Server for simultaneous distributed model training and evaluation 
& Strongly tied to the MathWorks ecosystem and licensing model \\
\hline
\end{tabular}
\end{table*}
\color{black}

\subsection{Solutions Comparison}
\label{subsec:mlops:comparison}
The selection between open-source and proprietary \mlops tools necessitates careful consideration of various factors. 
\color{black}
To facilitate comparison across the reviewed tools, Table~\ref{tab:mlops_tools_comparison} summarizes open-source and proprietary \mlops solutions in terms of lifecycle coverage, AutoML support, deployment model, end-to-end support, scalability, and main limitations.
\color{black}
Proprietary tools allow developers to construct end-to-end \mlops pipelines tailored to meet specific application requirements while offering flexibility in deployment options, particularly in hardware configuration.
Leveraging features such as auto-scaling and containerization, these tools facilitate seamless scalability of ML models, vital for accommodating the fluctuating workloads inherent in \ITLogisticShort.
Furthermore, most proprietary \mlops solutions boast user-friendly (or ever zero-code) interfaces and advanced features like AutoML, simplifying development processes, particularly for non-ML experts. 
Despite offering comprehensive support and maintenance, proprietary solutions may incur higher costs and provide limited deployment options, especially for on-premise (or hybrid) scenarios. 
Moreover, they may not provide exhaustive monitoring of system performance or behavior.
For instance, IBM Watson Studio integrates trustworthy artificial intelligence (TAI) monitoring and risk management capabilities that competitors do not natively implement.
At the same time, however, the measures implemented within IBM Watson Studio are not exhaustive, leaving room for additional unmonitored ethical considerations (e.g., robustness and privacy).
In contrast, open-source \mlops tools benefit from extensive support from the open-source community and provide developers with more flexibility in developing and customizing the application, optimizing the deployment and maintenance processes, and customizing the monitoring process (e.g., with custom technical, ethical, and business metrics) while reducing ML infrastructure costs.
These tools can be integrated with one another to provide several utilities for easy deployment, including integration with proprietary cloud services or private servers, and allow the integration of external monitoring tools to ensure the correctness of ML model operations. 
However, open-source solutions may pose challenges related to maintenance and longevity, as some libraries may face discontinuation or lack of maintenance, necessitating developers to replace or patch them over time.
Additionally, transitioning from open to closed-source software could require adjustments to maintain reliance on open-source tools. 

\subsection{Case Studies}\label{subsec:mlops:casestudy}
This section presents real case studies in \ITLogisticShort that have implemented \mlops solutions on the {\color{black} 6} areas of \ITLogisticShort presented in \autoref{sec:its-areas}. 
However, in contrast to the extensive literature around the relevance of \dataops in \ITLogisticShort applications, there is a noticeable lack of practical relevance and case studies of \mlops in this field.
While AI has garnered significant interest in the \ITLogisticShort domain, many studies prioritize the predictive performance of integrated models over the computational efficiency during development and post-deployment phases~\cite{jin2021trafficbert}. 
For instance, the ADMSv2 project~\cite{anastasiou2019admsv2} employs ML models for traffic data analysis but lacks \mlops practices. 
Although it includes \dataops components to scale data operations effectively, it relies on classic training and deployment methods, lacking support for automatic scaling, thereby limiting its ability to manage increasing workloads dynamically. 
Similarly, Alomari et al.~\cite{alomari2019road} and the SEHAA framework in~\cite{alotaibi2020sehaa} integrate \dataops for data processing but do not incorporate \mlops techniques, highlighting a gap in fully automated and scalable ML model lifecycle management.
In conclusion, compared to \dataops, there are significantly fewer research papers detailing the customization of ML pipelines. 
This gap is further confirmed by a recent literature review on \mlops in data science projects~\cite{haertel2023mlops}.
In the next paragraphs, we present the few case studies retrieved for each area in the last 10 years, highlighting complementary examples that demonstrate particularly complex ML management challenges.
From their analysis, we discover that the integration of both \dataops and \mlops in \ITLogisticShort literature remains limited, with only a few noteworthy examples addressing these combined practices, including the development by Conlan et al.~\cite{conlan2023real} of a real-time traffic management system in the UK, the system by Pineda et al.~\cite{pineda2023mlops} for delay prediction in European freight rail operations, and the framework by Mishra et al.~\cite{mishra2023autonomous} for urban air vehicle logistics. 
This underlines an ongoing challenge and potential area for future research, discussed further in \autoref{sec:discussion}.

\myparagraph{Vehicle Control Systems.}
In this area, we found a total of 5 case studies that integrate \mlops practices into automotive applications~\cite{ferreira2024mlops,hartmannsgruber2024improving,steffen2023continuous,weber2024streamlining,socha2022smirk}. 
Ferreira et al.~\cite{ferreira2024mlops} explore the challenges encountered at Bosch Car Multimedia in developing \dataops and \mlops processes for autonomous driving systems, emphasizing the need for automation in data management, training, and monitoring steps to manage large daily data inflows effectively and updates on the AI components. 
In their work the authors provide concrete implementation guidelines for the integration of \dataops and \mlops tools cases for the development of an efficient end-to-end pipeline. 
Similarly, Hartmannsgruber et al.~\cite{hartmannsgruber2024improving} and Steffen et al.~\cite{steffen2023continuous} showcase the transformative potential of continuous integration and delivery coming from \mlops frameworks in autonomous driving systems. 
Complementary to these efforts, Weber et al.~\cite{weber2024streamlining} investigate \mlops maturity levels in automotive ML projects, identifying contextual factors such as data quality and tool suitability that influence adoption. 
Lastly, Socha et al.~\cite{socha2022smirk} address robustness challenges in vehicle control systems by integrating \mlops monitoring techniques to handle out-of-distribution data effectively, emphasizing the importance of resilience in real-world applications.

\myparagraph{Traffic Management.} In this area, we found a total of 3 case studies, all related to traffic forecasting and planning~\cite{angarita2020general,conlan2023real,oakley2023foresight}.
Angarita et al.~\cite{angarita2020general} explore the use of automated machine learning (AutoML)~\cite{karmaker2021automl}, a promising way to automate development, training, and validation of the ML pipeline, for continuous and automated maintenance and monitoring of ML models in traffic forecasting applications. 
However, the implementation lacks full \mlops integration, as it does not address the need for automatic and scalable deployment solutions, namely the \textit{Ops} part of \mlops, that can dynamically adapt to increasing workloads or changing traffic conditions.
On the other hand, Conlan et al.~\cite{conlan2023real} introduce Foresight, an end-to-end \dataops, and \mlops system designed for managing traffic and road networks in the West Midlands region of the United Kingdom, covering major urban areas like Birmingham and Coventry, using real-time spatiotemporal forecasting. 
Foresight utilizes Amazon services (e.g.,Amazon S3~\cite{amazonCloudObject}, Amazon Kinesis~\cite{amazonDataStream}) to set up a pipeline capable of re-training traffic models periodically, such as once per week, to continuously incorporate new traffic data. 
The same project has later been extended in a more cost-effective inference architecture, namely Foresight Plus~\cite{oakley2023foresight}, which seamlessly utilizes Amazon SageMake on-demand services to handle workloads that have idle periods between traffic spurts and can tolerate cold starts (e.g., to accommodate peak traffic).

\myparagraph{Transportation Management Systems.}
In this area, we found a total of 3 case studies, related to the railway domain~\cite{zeller2024toward} and public transportation~\cite{sengul2022application,guzhva2023evaluating}.
Zeller et al.~\cite{zeller2024toward} engineer an abstract \mlops framework to support a trustworthy development of ML models (see also \autoref{sec:trustAI} for additional details) to ensure that, when deployed on transportation systems, they adhere to principles of reliability and safety. 
However, it does not propose a specific case study or suggestion on the tools to integrate, but instead provides a detailed conceptual architecture and emphasizes the need for continuous delivery, verification, and safety assessment within a safe \mlops framework. 
Similarly, Sengul et al.~\cite{sengul2022application} present a conceptual end-to-end \dataops and \mlops-based architecture to support public transportation management in cities, but without testing it in practice on realistic data.
However, complementary to \cite{zeller2024toward}, Sengul et al.~\cite{sengul2022application} present a list of tools that can be integrated to develop their architecture (e.g., Apache Airflow for data management and Kubernetes to manage ML deployment).
Lastly, Guzhva et al.~\cite{guzhva2023evaluating} propose an \mlops pipeline to determine the precise location of a tram within an urban environment (i.e., tram positioning) to provide accurate feedback to end-users. 
Their pipeline has been validated in practice through multiple tests on two tram routes, continuously collecting and integrating images captured by onboard cameras.

\myparagraph{Traveler Information and Experience Systems.}
The only relevant case study under this domain we identified is by Urias et al.~\cite{urias2023mlops}, who explore the use of \mlops in smart city applications, focusing on smart parking and smart tourism, among other urban services.
While Urias et al.~\cite{urias2023mlops} offer valuable insights into the use of \mlops for smart city applications for travelers, their work remains largely conceptual. 
Although they provide a thorough framework, there is still an opportunity for practical exploration and real-world implementation of these design principles on actual data to demonstrate the framework's effectiveness.

\color{black}
\myparagraph{Freight and Supply Chain Logistics.}
In this area, we identified 7 case studies spanning railway freight management~\cite{pineda2023mlops}, logistics in manufacturing production~\cite{faubel2024mlops,cha2023development}, maritime logistics~\cite{morariu2024analysing}, inventory management and sales forecasting~\cite{khan2024optimizing}, e-commerce recommendation and delivery~\cite{jana2023framework}, and large-scale delivery services~\cite{michelangelo2024uber}.
Pineda et al.~\cite{pineda2023mlops} focus on \mlops for freight rail operations by developing a tool for real-time delay prediction in freight trains. Their system automates the entire \dataops and \mlops pipeline using tools such as Apache Airflow and Kubeflow to manage railway trips across multiple European countries and reduce disruptions in freight rail logistics. Although the study focuses on Luxembourg's freight rail network, the proposed methodology can be adapted to other freight rail systems, highlighting the scalability of \mlops in real-world logistics applications.
Faubel et al.~\cite{faubel2024mlops} investigate \mlops integration for logistics management in three industrial production companies, including automotive manufacturers, showing that both \dataops and \mlops are essential for automating operational decision-making and improving efficiency in logistics workflows. The tools adopted by these companies include Kubernetes, Apache Spark, MLflow, and AWS and Azure solutions.
Cha et al.~\cite{cha2023development} further emphasize the role of Kubeflow in supporting intelligent logistics for smart factories, improving manufacturing productivity through the automation of ML pipelines.
Morariu et al.~\cite{morariu2024analysing} examine the maritime industry and stress the need for real-time decision-making systems in complex maritime environments. However, their study mainly addresses real-time data-processing challenges and does not provide a detailed practical implementation of \mlops for scalable and automated model lifecycle management.

On the supply-chain and delivery side, Jana et al.~\cite{jana2023framework} propose an AWS-based \mlops pipeline for e-commerce companies aiming to improve customer experience and optimize operations, such as product recommendation, in real time. This work highlights the role of \mlops in enabling continuous model training and deployment on real data.
Khan et al.~\cite{khan2024optimizing} present a survey on the optimization of retail operations in China through \mlops for inventory management and sales forecasting. The study focuses on high-volume data and the challenges arising in this context, but does not describe the technical implementation of the \mlops pipeline in detail.
Finally, Uber's Michelangelo platform~\cite{michelangelo2024uber} exemplifies large-scale \mlops for delivery services such as Uber Eats, automating and scaling model deployment across Uber's applications and substantially reducing the operational complexity of managing ML workflows in large-scale delivery services.

\myparagraph{Emergency Management.}
In this area, we identified only 2 case studies, presented by Peramo~\cite{peramo2022intelligent} and the ML4Floods project~\cite{portales2023global}, both related to flood detection and mapping for damage prevention, public safety, and emergency response during flood events.
Specifically, Peramo~\cite{peramo2022intelligent} integrates Weights and Biases to support the development and monitoring of an ML-based flood detection system using traffic images.
ML4Floods~\cite{portales2023global}, in turn, provides an end-to-end \mlops-oriented pipeline for flood-extent segmentation from optical satellite imagery, covering data ingestion and preprocessing, model training, evaluation, inference, deployment, post-processing, and flood-map visualization.
The system operationalizes flood mapping under conditions commonly encountered during disasters, such as cloud coverage, limited post-event observations, and the need to rapidly generate flood products for damage assessment.
For example, the associated flood-mapping framework is designed to produce flood-extent maps without human intervention, can operate on both Sentinel-2 and Landsat 8/9 imagery, and was demonstrated on major flood events in Pakistan and Australia, where the generated flood maps were intersected with population, building, and cropland layers to support impact assessment~\cite{portales2023global}.

Finally, although it does not implement a specific case study, the work of Remil et al.~\cite{remil2024aiops} is worth mentioning, as it presents a comprehensive review of ML solutions for incident management, although without supporting explicit \mlops solutions.
Their work provides technical guidelines for implementing \mlops architectures to automate the detection and resolution of incidents in critical systems.
We also identified recent ML-based flood detection systems, such as~\cite{shankar2025disaster}, which combine YOLO-based detection~\cite{zhao2019object} with real-time alerting mechanisms. However, these works mainly focus on model development and application-level response rather than explicit \mlops lifecycle automation.

\color{black}
\myparagraph{Remarks.} 
Overall, the reviewed case studies indicate that \mlops adoption in \ITLogisticShort remains considerably less mature than \dataops adoption, with only a limited number of studies addressing the full ML lifecycle beyond model training and evaluation.
Across domains, these works reveal a common need for more systematic support for automated retraining, while each application area introduces additional operational requirements.

Vehicle Control Systems and Transportation Management Systems place particular emphasis on safety-aware continuous verification and monitoring, since deployed models must operate under changing road, traffic, and environmental conditions.
Traffic Management applications primarily highlight the value of periodic retraining and elastic inference infrastructure, which allow forecasting models to adapt to evolving traffic patterns and variable workloads.

Freight and Supply Chain Logistics provide some of the most complete examples of \mlops adoption, where tools such as Airflow and Kubeflow, together with cloud-based services, are used to automate training, deployment, and monitoring across geographically distributed logistics networks.

Traveler Information and Experience Systems and Emergency Management remain less explored from an \mlops perspective, with existing works often limited to conceptual frameworks or application-level ML systems rather than mature lifecycle automation.

Finally, current \mlops case studies in \ITLogisticShort highlight a clear prioritization of isolated model development and predictive performance, with limited evidence on how ML-based systems are retrained, versioned, deployed, monitored, and updated after deployment.
\color{black}

\subsection{The Recent \llmops Development}\label{subsec:mlops:llmops}
Large Language Models (LLMs) have emerged in 2019~\cite{zhao2023survey} as powerful Deep ML models, based on the transformer architecture, for language-related tasks (e.g., text classification and generation), significantly outperforming their predecessors.
LLMs have then evolved beyond text processing alone to be able to handle diverse data modalities, including images~\cite{zhou2023vision}, graphs~\cite{moghadas2024strada}, and other structured and unstructured data types~\cite{le2024multimodal}.
Today, LLMs are instrumental in extracting meaningful insights from free-form text~\cite{movahedi2024crossroads} and multimodal data~\cite{le2024multimodal}, utilizing pre-trained LLM embeddings~\cite{kenton2019bert}. 
They facilitate autonomous problem-solving through ChatBots~\cite{wang2024traffic} and assist operators in providing domain-specific answers with Retrieval Augmented Generation~\cite{yang2024transrag}. 
Consequently, organizations are increasingly incorporating LLMs into their services~\cite{wandelt2024large}. 
For example, the Melbourne Airport has developed a ChatBot based on LLMs that provides information regarding real-time flight updates, shop searches, and parking information~\cite{melbourneairportMelbourneAirport,wandelt2024large}.
However, while LLMs may appear as specific ML models within \mlops, they present distinct challenges, paving the way for Large Language Model Operations (\llmops)~\cite{diaz2024large,sinha2024llmops}. 
\color{black}
\llmops is treated as a specialized extension of \mlops for LLM-based systems, where additional operational concerns arise from prompt engineering, retrieval augmentation, fine-tuning or adaptation of pre-trained models, guardrails, hallucination monitoring, token and cost management, and the integration of human feedback.
\color{black}
For instance, in LLM-based text generation, ensuring that the output aligns with specific requirements, such as avoiding harm~\cite{weidinger2021ethical}, necessitates the implementation of mechanisms to enforce generation safety and security. 
For this purpose, the control and monitoring mechanisms, such as guardrail~\cite{rebedea2023nemo}, need to be integrated and deployed in the system. 
Furthermore, in most LLMs-based applications, it is pretty uncommon to train an LLM from scratch.
This process is mainly accessible to the big techs (e.g., OpenAI, Google, and Mistral) who have both the huge data sources and computational power needed for this purpose~\cite{narayanan2021efficient,naveed2023comprehensive}.
What is commonly employed, instead, are the pre-trained LLMs~\cite{hu2021lora,min2023recent}, often referred to as foundation models~\cite{zhou2024comprehensive,yang2023foundation}.
\color{black}
Note indeed that the term Foundation Model Operations (\fmops)~\cite{tantithamthavorn2024mlops,bhattacharjee2025overview} can be understood as a broader generalization of \llmops to foundation models beyond text, including multimodal and generative models such as text-to-image, text-to-video, and speech-to-text systems~\cite{sinha2024llmops}.
However, since \fmops and \llmops share many operational challenges, and since \llmops is the most consolidated term in the literature, in this survey we use \llmops as the main reference while noting that several considerations also extend to broader foundation-model deployments.

In the following, we describe how to align the \mlops lifecycle (i.e., from the data collection until the feedback steps) of Figure~\ref{fig:mlops_design} toward \llmops.
\color{black}

\myparagraph{Data Collection.} Due to the use of pre-trained foundation models, \llmops requires data diversity and representatives more than volume~\cite{shi2024maximizing}.
Traditional \mlops often generate ML models from scratch, and consequently, large volumes of labeled data are required~\cite{john2021towards}. 
\llmops, focusing on adjusting the foundation model to the specific task through model adaptation techniques, require small vertical and high-quality datasets~\cite{shi2024maximizing,shan2024enterprise}.
For this reason, these data must be carefully curated to ensure that they are representative of the problem at hand.
\color{black}
Concretely, for \ITLogisticShort services, this shifts the emphasis from collecting more data to selecting reliable domain evidence, such as updated traffic conditions, operational constraints, regulations, and service information, that can safely support LLM-based decisions and recommendations.
\color{black}

\myparagraph{Feature engineering.} 
Similarly to \mlops, when dealing with Deep ML, in \llmops the feature engineering phase is performed by the trainable or pretrained foundational models (e.g., BERT~\cite{kenton2019bert}), which generate effective representations directly from raw data avoiding or strongly reducing the need for human intervention~\cite{luo2024autom3l}.
This can be done at the cost of the high computational requirement using these large models~\cite{zhao2023survey}.
While foundation models often demonstrate considerable performance on their own, some applications require greater accuracy (e.g., legal~\cite{cheong2024not} or medical~\cite{goyal2024healai} decisions) or better alignment with domain-specific metrics, such as ethical or cultural~\cite{cheong2024not} considerations. 
In these cases, two feature engineering approaches can be integrated to further improve their effectiveness.
The first one, prompt engineering~\cite{chen2023unleashing}, modifies the input (prompt) to enable the adaptation of models to specific tasks, further enhancing their performance and versatility across diverse applications leveraging the fact that a better input usually generates a better response in LLMs~\cite{chen2023unleashing,white2023prompt}. 
The second one involves enriching the original input (prompt) by integrating external knowledge sources~\cite{dong2022survey}.
Techniques such as Retrieval Augmented Generation (RAG) pipelines~\cite{fan2024survey} or Knowledge Graphs~\cite{pan2023large} enrich the original input with additional context-specific information augmenting the model's understanding and reasoning capabilities~\cite{dong2022survey}.
\color{black}
In \llmops, feature engineering is therefore not limited to transforming raw data into model features; it also includes enriching the context provided to the model at inference time.
In RAG-based systems, for example, relevant documents or data sources are retrieved when a user query is received and incorporated into the prompt to ground the model response.
\color{black}

\myparagraph{Model Training.} 
The training process of LLMs is challenging and might be conceptualized into two (possibly three) modules.
The first one encompasses one or multiple large Deep ML models designed to embed the raw data (e.g., text, images, graphs, and audio) into a representation vector (or embedding)~\cite{shalev2014understanding}.
Subsequently, this representation is utilized by a second module, called the subtask module, for classification~\cite{ruan2024twitter}, clustering~\cite{bhandari2024urban}, or generation~\cite{melbourneairportMelbourneAirport} purposes.
When the subtask generation module is utilized, an additional control module is introduced for refining~\cite{min2023recent} or safeguarding~\cite{dong2024building} the generated response.

\color{black}
Analogously to \mlops, model training in \llmops involves both model engineering and model fitting.
However, unlike classical \mlops, \llmops in downstream deployments rarely focuses on full pre-training~\cite{sinha2024llmops,diaz2024large}.
Instead, model engineering typically concerns the selection of a suitable pre-trained foundation model, the definition of instruction-tuning or parameter-efficient fine-tuning strategies, and the configuration of safety alignment mechanisms and guardrails.
When a subtask module is used, model engineering also includes selecting the corresponding model type and hyperparameters.
Finally, the model fitting phase focuses on adapting the selected components using the available task-specific data. 
\color{black}

\myparagraph{Model Validation.}
The model validation of \llmops is conceptually very similar to the one of \mlops with one exception: a strong focus on the validation metrics for the generative tasks.
When LLMs are used for classification or clustering subtasks, they can be viewed as an incremental evolution of classical ML techniques, similar to enhanced pre-trained models. 
However, when employed for generative purposes, LLMs have the potential to produce harmful or unethical content, whether intentionally or unintentionally.

For this reason, relying solely on metrics commonly used in classical ML is insufficient. 
To comprehensively evaluate the performance of the final LLM-based tool, a combination of quantitative and qualitative metrics is necessary, striking a balance between automation and human intervention.
Quantitative metrics (e.g., ROUGE~\cite{lin2003automatic}, BERTScore~\cite{zhang2019bertscore}, BLEU~\cite{papineni2002bleu}, and many others~\cite{chang2023survey}) are task-specific benchmarks for delivering automated quantitative insights into LLM performance.
Human evaluators provide further insights into the quality, coherence, and relevance of generated content by identifying subtleties that automated metrics may overlook, such as appropriateness in context, grammatical correctness, and adherence to stylistic conventions.

\myparagraph{Release, Deploy, Monitoring, and Feedback.} 
Deployment in \llmops requires specialized considerations due to the nature of LLMs, which demand significantly more computational resources than previous ML models integrated in \mlops pipelines. 
While both \mlops and \llmops involve managing computational resources, \llmops require them throughout the training and deployment phases. 
In \mlops, computational resources are primarily focused on training tasks, where large-scale datasets and complex neural network architectures demand substantial computing power for model training. 
However, once the model is trained, deployment often involves less intensive computational requirements, shifting the focus to serving predictions efficiently and reliably. 
Conversely, \llmops faces distinct challenges related to training and deployment phases. 
LLMs involve performing more calculations and exhibit large memory footprints during training, fine-tuning, and inference, necessitating specialized infrastructure and optimization techniques to handle their computational demands effectively. 
Specifically, once trained, LLM inference requires processing input sequences through multiple neural network layers, affecting inference times. 

Regarding model validation, monitoring in \llmops is a multifaceted process that tracks various quantitative metrics and human evaluations. 
During the monitoring phase, the focus is on detecting potential biases, ethical issues, or unintended consequences in the outputs generated by LLMs to ensure that LLMs operate ethically and effectively in diverse contexts~\cite{chen2023hallucination,liu2024efficient}.
Finally, the feedback stage plays a more relevant role in refining and optimizing the performance of LLMs. 
The data collected from this stage can be further utilized to refine LLMs through interactions with human feedback~\cite{ouyang2022training} (e.g., explicit labels, ratings, or human corrections in response to the model's outputs), leading to continuous improvement and adaptation to real-world interactions.

\color{black}
\myparagraph{Tools.}
Beyond the general-purpose \mlops platforms discussed in Sections~\ref{subsec:mlops:openTools} and~\ref{subsec:mlops:proprietaryTools}, the \llmops landscape includes a growing set of dedicated tools designed to address the specific operational requirements of foundation-model deployments.
Hugging Face~\cite{HuggingFace}, with more than $160$ thousand GitHub stars as of June 2026, has become a widely adopted hub of the open-source LLM ecosystem, offering model and dataset repositories, the Transformers library for model loading and fine-tuning, and parameter-efficient adaptation methods such as PEFT.
These resources support several stages of the \llmops lifecycle, including data collection, feature engineering, model adaptation, and deployment.
For inference serving, vLLM~\cite{kwon2023efficient}, with more than $60$ thousand GitHub stars as of June 2026, provides a high-throughput serving engine with continuous batching, improving GPU memory utilization and inference efficiency.
At the application layer, LangChain~\cite{LangChain}, with more than $130$ thousand GitHub stars as of June 2026, supports the development of LLM-powered applications and their interaction with external tools, data sources, and memory.
Finally, tools such as Deepchecks~\cite{Deepchecks} and benchmarks such as HarmBench~\cite{mazeika2024harmbench} support continuous validation, monitoring, and safety evaluation, complementing the \TAI principles of robustness and trustworthiness discussed in \autoref{sec:trustAI}.
Collectively, these tools reflect a rapidly expanding and globally distributed \llmops ecosystem that complements the general-purpose \mlops platforms surveyed in this work.
\color{black}

\color{black}
\subsection{Emerging \agentops for Agentic AI Systems}
\label{subsec:mlops:agentops}

The discussion above focused on LLM-based applications in which the model is invoked to generate, classify, retrieve, or summarize information.
However, recent AI deployments are increasingly moving toward agentic systems~\cite{acharya2025agentic,pati2025agentic}, namely systems in which an AI component is not limited to producing a single response, but can pursue a goal through multiple intermediate steps.
In this context, an agent can be understood as an AI-based system that observes an input or environment, decides which action to take next, and may call external tools, retrieve information, store or use memory, interact with users, or coordinate with other agents to complete a task~\cite{sapkota2025ai}.
For example, a conventional LLM-based application may answer a travel question, summarize a traffic report, or retrieve information about a transport service.
By contrast, an agentic system may autonomously retrieve live traffic data, compare alternative routes, call booking or fleet-management tools, coordinate with other agents, and propose operational actions.

This shift motivates an additional operational layer, known as Agent Operations (\agentops)~\cite{wang2025agentops,dong2024agentops,xia2025agentops}.
Compared with \llmops, which focuses on managing LLM-based applications, \agentops focuses on the runtime behavior of agentic workflows.
While \mlops addresses the lifecycle of ML models and \llmops specializes this lifecycle for LLM-based applications, \agentops is more system-oriented.
It monitors and governs agentic workflows composed of models, prompts, tools, memory, external data sources, and environment interactions.
Therefore, \agentops does not replace \mlops or \llmops; rather, it extends them in terms of functionalities, responsibilities, and risks.

From a functional perspective, \agentops shifts the monitoring target from the model output alone to the full execution trajectory of the agent, including prompts, intermediate reasoning steps, retrieved information, tool calls, memory updates, interactions with external systems, and final actions.
From a responsibility perspective, this means that developers and operators must govern not only the underlying model, but also the surrounding system that decides when to call tools, which information to use, and which actions to execute.
This broader operational scope also increases the potential impact of failures, since an incorrect action may propagate across tools, services, users, or other agents.
As a result, agentic systems introduce additional risks, including incorrect tool calls, unstable planning, memory failures, prompt injection, cascading errors across agents, and emergent behavior in multi-agent coordination~\cite{su2026survey}.

For this reason, \agentops introduces operational mechanisms beyond \llmops to observe, control, diagnose, and recover the behavior of agentic systems during execution.
These mechanisms aim to make agent workflows traceable, verify that actions remain consistent with operational and safety policies, identify the origin of failures, and support corrective interventions when the system behaves unexpectedly.
They remain aligned with the broader \mlops lifecycle, but operate at a higher level of abstraction, where the monitored object is not only the underlying model but the complete agentic system built around it.

We argue that \agentops will be highly relevant for future \ITLogisticShort deployments.
At the same time, \agentops is still an emerging and rapidly evolving area.
Unlike \dataops, \mlops, and, to a lesser extent, \llmops, it has not yet reached the same level of methodological maturity, standardization, or systematic adoption in the scientific literature.
Accordingly, this survey does not aim to provide a complete treatment of \agentops.
Rather, we introduce it as a recent development likely to influence future trustworthy \ITLogisticShort systems, and refer the reader to dedicated surveys and technical reports for a broader discussion of agentic AI operations~\cite{wang2025agentops,dong2024agentops,xia2025agentops}.
\color{black}


%
\section{\TAI}
\label{sec:trustAI}
In various decision-making domains, including \ITLogisticShort~\cite{guo2023sustainability,bevsinovic2021artificial,Gaur2022ITSXAI}, healthcare~\cite{rajpurkar2022ai}, and policing, the widespread integration of AI has raised significant technical and ethical challenges and concerns~\cite{du2021paradoxes}. 
This is due to the increasing complexity of AI-based systems, which, while achieving higher performance in specific metrics (e.g., accuracy and computational requirements), pose many technical challenges (i.e., the so-called technical debt~\cite{sculley2015hidden}) and raise ethical concerns from society at large~\cite{du2021paradoxes}.

In simpler words, when designing technical instruments, optimizing single metrics may lead to problems with other hidden or implicit metrics.
For example, focusing solely on predictive accuracy (i.e., correlation) without considering model adherence to physical principles (i.e., causality) risks shortcut learning and spurious correlations. 
This lack of consideration for underlying principles can result in models that perform well on specific tasks but fail in real-world scenarios where causal relationships are crucial.
Using historical data for predictive accuracy in social phenomena (e.g., hiring) can perpetuate societal biases (e.g., on gender and race). 
Overemphasis on removing biases might result in hallucinations or misrepresentations. 
Prioritizing predictive accuracy over robustness can expose models to malicious users, potentially leading to undeserved benefits or social issues.

These concerns were highlighted by the record number of AI-related incidents reported in 2023, marking a significant reckoning for the field~\footnote{\url{https://incidentdatabase.ai}} and driven partly by the introduction of the AI-Act in Europe~\cite{ai-act-council,eu2021ai-act,european_ai_act_2021}. 
Gartner estimates that by 2025, approximately $30\%$ of all AI-based products will require a trusted AI framework~\cite{kaur2022trustworthy}. 
Recent studies also indicate that $86\%$ of users are likely to trust and remain loyal to companies supporting ethical AI principles~\cite{Edelman2024,kaur2022trustworthy}.
For this reason, researchers have approached this problem from different perspectives, identifying two main research directions~\cite{winfield2019machine}. 
The first one studies how to design systems that can follow or learn technical principles and moral notions and autonomously behave accordingly. 
Works like the Comparative Moral Turing Test~\cite{allen2000prolegomena} or Ethical Turing Test~\cite{anderson2018geneth} assess the morality of automated systems' choices. 
Due to the strong connection between philosophical, ethical, and technical problems, this research branch remains relatively unexplored. 
The second direction focuses on how humans should design and develop AI-based systems to minimize possible harm from poor design, inappropriate application, or misuse~\cite{kaur2022trustworthy,eu2021ai-act}. 
Our work falls within this category. 
In this context, the best solution that researchers have found is designing an increasing number of metrics that properly characterize the behavior of AI-based systems, considering all aspects needed for the particular application~\cite{oneto2024towards}. 
This approach aligns well with the \dataops and \mlops (including \llmops) frameworks discussed in the previous sections. 
Both \dataops and \mlops include steps like ``Analysis \& Visualization'' and ``Monitor'' for \dataops (see \autoref{fig:dataops_design}) and ``Model Validation'' and ``Monitor'' for \mlops (see \autoref{fig:mlops_design}), where these metrics can be integrated during both design (``Analysis \& Visualization'' for \dataops and ``Model Validation'' for \mlops) and operations (``Monitor'' for both \dataops and \mlops).
\color{black}
In the end-to-end view presented in Figure~\ref{fig:dataops_mlops_tai_framework}, this section corresponds to the \TAI layer, which cuts across both the \dataops and \mlops lifecycles.
Rather than serving as a final assessment step, \TAI provides the metrics and mitigation strategies needed to evaluate data quality, model behavior, and deployed services in terms of sustainability, robustness, explainability, fairness, and privacy.
\color{black}

In the following, we address the importance of developing new technical and ethical metrics to monitor AI-based systems and, in case of misbehavior, techniques to mitigate problems towards more \TAI. 
These issues are especially relevant in the \ITLogisticShort domain, where the combination of technical, economic, legal, and human-related factors and requirements makes trustworthy AI crucial for the widespread adoption of AI.
\color{black}
In the following sections, we focus on the main metrics that characterize the development of more \TAI-based systems for \ITLogisticShort: sustainability (\autoref{subsec:tai:sustainability}), robustness (\autoref{subsec:tai:robustness}), explainability (\autoref{subsec:tai:explainability}), fairness (\autoref{subsec:tai:fairness}), and privacy (\autoref{subsec:tai:privacy}).
Note that some metrics can be both technical and ethical.
For instance, sustainability is a technical metric because it concerns the efficiency of ML models, but it is also an ethical requirement.
Similarly, robustness prevents AI models from being manipulated, serving as both a technical and ethical requirement.
Note also that metrics can be combined, such as fairness and robustness as in~\cite{OnetoC128}, generating secondary yet important metrics.
Finally, the metrics discussed in this section are those that, from the authors' perspective, are among the most important, widespread, and relevant for \ITLogisticShort, although other important metrics can also be considered, such as the ability of a model to be formally verified~\cite{huang2017safety}.
For additional metrics, please refer to the following surveys~\cite{kaur2022trustworthy,Oneto22TrustGraph,Cin2022MachineLS}.
\color{black}
The section then concludes with case studies in \ITLogisticShort (\autoref{subsec:trustAI:casestudies}).
\color{black}

\subsection{Sustainability}\label{subsec:tai:sustainability}
The two main phases of ML training and inference present significant challenges regarding their economic costs and associated environmental impacts~\cite{Strubell20Energy,Garcia19Energy}.

Training, in particular, requires collecting large amounts of data and optimizing (or fine-tuning) a substantial number of parameters, which increases the computational demands and energy consumption of both the training and inference phases~\cite{Garcia19Energy}.
To mitigate these challenges, one approach involves optimizing data usage by leveraging techniques such as data selection~\cite{kaushal2019learning,wei2006feature}, or transfer learning from pre-trained models~\cite{zhou2022leveraging,liu2019exploring}. 
Another strategy is to modify the model's architecture to reduce the number of parameters~\cite{dong2017learning,cheng2024survey} or their precision~\cite{hubara2018quantized,rokh2023comprehensive}. 
Additionally, selecting appropriate training optimizers can speed up the training process, and using custom hardware (e.g., GPUs, TPUs) can reduce training time or enhance efficiency using the same amount of energy~\cite{Jouppi17GoogleAsicTPU,Sze17SurveyAsic}.
However, high-end hardware often remains necessary, which may be cost-prohibitive for smaller enterprises or beyond the technical skills of larger ones in \ITLogisticShort~\cite{ribeiro2015mlaas}. 
Consequently, organizations typically rely on Machine Learning as a Service (MLaaS) providers to outsource the training process, thereby reducing complexity and costs~\cite{ribeiro2015mlaas,yao2017complexity}. 
Nonetheless, adopting these strategies necessitates trust in third-party entities responsible for training the models~\cite{cina2023wild}.

The inference phase poses fewer computational challenges than the training phase, as it is generally computationally efficient. 
Even complex models (e.g., LLMs) may take only a few seconds to make a prediction. 
However, in the \ITLogisticShort domain, the sheer volume of required predictions (e.g., for all delivery vehicles) and limitations in available hardware (e.g., computational requirements or battery life) can pose significant challenges. 
As a result, even small savings in the computational cost of individual predictions can substantially impact sustainability, prediction time, and costs~\cite{Yin2021DeepLO}. 
In other cases, the use of hardware accelerators may be impractical~\cite{Jouppi17GoogleAsicTPU,Sze17SurveyAsic}.

\myparagraph{Metrics and Tools.}
Given the growing call for the development and incorporation of new evaluative metrics that capture a comprehensive range of factors to ensure sustainable AI, various metrics have been developed to monitor high-level factors (e.g., costs, carbon emissions, power consumption) influencing the environmental footprint of training or operating AI models.
Costs indicate the financial resources required for training and operating AI models, including hardware, energy consumption, and personnel expenses.
The carbon emissions metric measures the number of greenhouse gases emitted during the training and operation of AI models.
To measure and manage these metrics effectively, several tools have been developed. 
CodeCarbon~\cite{benoit_courty_2024_11171501} and CarbonTracker~\cite{anthony2020carbontracker} are popular tools for measuring carbon emissions and footprint. 
Eco2AI~\cite{eco2AI} accumulates statistics about power consumption and CO$_2$ emissions during code execution, while Zeus~\cite{zeus-nsdi23} focuses on deep learning energy measurement and optimization. 
For monitoring power consumption, Scaphandre~\cite{hubbloorgIntroductionScaphandre} and PowerJoular~\cite{noureddine-ie-2022} are widely used; they provide metrics on electrical power consumption across multiple platforms and processes.
Complementary to these high-level metrics, low-level factors (e.g., number of epochs to reach convergence, representation precision, number of weights to be trained) can be tuned to further decrease the computational complexity and power consumption of AI. 
All these metrics and tools can be integrated together within \mlops pipelines into the ``Monitoring'' phase to balance the economic and environmental footprint of training or operating AI models with other performance metrics, such as accuracy.

\subsection{Robustness}
\label{subsec:tai:robustness}
As ML-based systems are increasingly deployed in critical applications (e.g., autonomous driving, traffic management, predictive maintenance), their robustness and resilience against unforeseen errors, misuse, and malicious users become of paramount importance.
In autonomous driving, for example, vehicles need to be extremely reliable in the face of both intentional and unintentional changes in the surrounding environment. 
However, ML models have demonstrated inherent vulnerabilities to carefully crafted attacks, including data poisoning~\cite{cina2023wild,Cin2022MachineLS}, adversarial examples~\cite{biggio18wild}, model stealing~\cite{orekondy2019knockoff}, membership inference~\cite{shokri2017membership}, and energy-latency attacks~\cite{cina2022energy}.
These attacks can influence both training (e.g., poisoning) and inference (e.g., adversarial examples) phases of an ML pipeline.
Moreover, attacks can target different aspects of the model's functionality and security, including availability~\cite{barreno2010security}, making the ML models unavailable for use at critical times or making them inaccessible to legitimate users, integrity~\cite{barreno2010security}, leading to incorrect or misleading predictions, and privacy~\cite{barreno2010security}, resulting in data breaches or leakage of confidential data (see \cite{biggio18wild,cina2023wild} for details). 
Therefore, exhaustive data monitoring, cleaning, and anomaly removal, together with preventive security testing of AI systems, become fundamental for safeguarding the business and ensuring compliance with relevant regulatory standards and guidelines.
Implementing these measures helps ensure that AI models maintain their reliability, accuracy, and security throughout their lifecycle. 
However, the complexity of testing and measuring the reliability of ML-based systems increases with the complexity of the model itself and its intrinsic explainability (see Section~\ref{subsec:tai:explainability}). 
Explainable models are, by definition, much easier to test, while deep ML models are significantly harder to evaluate. 
Moreover, there is often an asymmetry between the ease of testing and attacking; in some cases, testing can be hard while attacking is simple (and vice versa), highlighting the inherent advantage attackers may have over defenders~\cite{marro2023computational}.

\myparagraph{Metrics and Tools.}
Multiple robust testing procedures have been developed to proactively assess the prerequisites for robustness and compliance before deploying ML systems in critical applications. 
To check the model's robustness, widely adopted measures include the Attack Success Rate (ASR), security (or robustness) evaluation curves, and accuracy under distribution shift.
The ASR quantifies the percentage of adversarial examples or poisoning attacks that successfully deceive the model into making incorrect predictions.
Security (or robustness) evaluation curves plot the model's performance (e.g., accuracy) against different attack strengths or perturbation levels, offering insights into its vulnerability across various scenarios.
The security evaluation curves are also used to compute the \emph{optimality} score~\cite{cina2024attackbench}, which measures the effectiveness of the attack used to evaluate the adversarial robustness of the model under inspection. 
The higher the optimality score, the better the attack algorithm is at assessing the model's robustness.
Accuracy under distribution shift evaluates how well a model performs when the distribution of the input data shifts from the training distribution~\cite{Sun2022SHIFTAS}.
In other words, it measures the model's ability to generalize to new, unseen data distributions, which is crucial for real-world deployment where data may vary over time or across different environments.
Lastly, the energy consumption ratio~\cite{Shumailov21Sponge,cina2022energy} is the measure monitored when mounting attacks targeting the energy-latency performance of a model.
These metrics together evaluate whether ML systems exhibit unpredictable behavior, deviating from established safety protocols and potentially leading to accidents.
Among the tools and resources integrating them, we find FoolBox~\cite{rauber2017foolboxnative}, SecML~\cite{melis2019secml}, CleverHans~\cite{papernot2018cleverhans}, and Art~\cite{art2018}, which serve as benchmarks for assessing the robustness of ML models in the presence of malicious users. 
The algorithm proposed in \cite{cina2022energy,Shumailov21Sponge} can be utilized to assess the resilience of AI systems against recent attacks designed to increase power consumption. 
Additionally, the SHIFT dataset~\cite{Sun2022SHIFTAS} has been introduced to investigate performance degradation in perception systems and assess the generality of ML models.
However, it is worth mentioning that most of these tools and datasets target the image vision domain, leaving the need for standardized benchmarks for other input domains (e.g., graphs and tabular data) open. 
Furthermore, to date, robustness is mostly evaluated empirically, with gradient-based~\cite{chakraborty2018adversarial} or gradient-free attacks~\cite{zhu2024review}. 
Nevertheless, empirical evaluations have been proven to fail in some cases due to the inherent complexity of the attack optimization process, giving a false sense of security~\cite{athalye2018obfuscated}.
To address these common challenges, recent work has proposed standardized evaluation frameworks~\cite{cina2024attackbench,chao2024jailbreakbench,croce2021robustbench}, practical recommendation guidelines~\cite{cina2024attackbench,li2021towards,tramer2020adaptive}, and debugging tools~\cite{pintor2022indicators}. 
These efforts aim to enhance the reliability and robustness of ML systems across various domains, ensuring their safe and effective deployment in critical applications.
Alternatively, formal verification tools have been developed to theoretically guarantee the robustness of ML models~\cite{kantchelian2016evasion,guidotti2020verification,jia2020efficient,ranzato2020abstract}. 
However, the robustness verification problem is NP-complete~\cite{kantchelian2016evasion,wang2020lp}, and these tools often suffer from scalability issues~\cite{calzavara2023verifiable}, making them mainly applicable to smaller models.

\subsection{Explainability}
\label{subsec:tai:explainability}
Explainable AI (XAI) involves elucidating or revealing the decision-making mechanisms of ML models, allowing users to understand how inputs and outputs have been related, thereby understanding why AI models make specific decisions~\cite{ali2023explainable}.
In other words, explainability aims at providing model/outcome explanations for black-box ML models (e.g., deep networks), encouraging trust in their users~\cite{oneto2024towards,Repetto2025Evaluating}.
Trust is essential in \ITLogisticShort as the decisions may have an impact on individuals or customers.
For example, explainable route planning or automatic
vehicle location~\cite{gaur2022explainable} may improve dispatcher comfort and service level for users.
However, defining explainability within a single, universally accepted notion is challenging, as many proposed definitions offer partial clarity.
Many researchers argue that the ongoing research on XAI lacks a formal framework~\cite{Paez19pragmatic,lipton2018mythos,yao2021explanatory}.
A consequence of this lack is the high fragmentation of the field~\cite{ali2023explainable} even if recently a lot of efforts have been spent to rationalize and systematize the field~\cite{ali2023explainable,dwivedi2023explainable,Guidotti2019Survey,mattioli2024understanding}.

In this work, we follow the taxonomy developed in~\cite{arrieta2020explainable}, which distinguishes XAI between models that are interpretable by design (\emph{transparent}) and those explained by external XAI techniques, called \emph{post-hoc} methods, which focus on explaining the entire model (\emph{global explanations}) or single (or a group of) predictions (\emph{local explanations}).
\emph{Transparent} models are inherently explainable algorithms, allowing users to comprehend the model's processes to produce input data outputs~\cite{tibshirani1996regression,kuo2019interpretable,kawaguchi2016deep}.
The level of transparency can vary significantly. 
Fully transparent models, such as linear or rule-based models, provide clear insights into their decision-making processes, making them easy for both humans and mathematical tools to inspect and query. 
On the other hand, less transparent models, like Random Forests~\cite{cutler2012random} or Kolmogorov-Arnold Networks~\cite{liu2024kan}, are more complex and less interpretable. 
When a model is not sufficiently transparent for the application or its inspection is too complex for humans, \emph{post-hoc} methods need to be applied. 
These methods help to approximate and explain the non-transparent models, balancing the need for accuracy and interoperability.

Post-hoc techniques are further divided into global and local explanations.
Global explanations provide a global overview of the model's decision-making process~\cite{Oneto22TrustGraph}. 
They aim to explain the general patterns, trends, and feature importance that the model has learned.
For example, they may approximate the entire model to offer a general understanding of what has been learned from the data and how the model performs predictions~\cite{soares2020explaining}, or use perturbation techniques (e.g., permutation test) to evaluate the feature importance for the model~\cite{robnik2018perturbation}.
Local explanations provide insights into why a model made a specific prediction for a single (or a group of) instance(s). 
An example is LIME~\cite{ribeiro2016should}, which offers insights into how a black-box model arrived at a particular prediction by generating a simpler, more interpretable model near that prediction. 
Lastly, techniques such as SHAP~\cite{lundberg2017unified} can support both local and global explanations, offering flexible tools to interpret model decisions across different contexts.

\myparagraph{Metrics and Tools.}
In XAI, various metrics can be utilized to assess the effectiveness and reliability of model explanations~\cite{carvalho2019machine,hosain2024explainable}. 
Among them, we find quantitative metrics to evaluate the quality, efficacy, and reliability of explanations in XAI. 
\emph{Accuracy} is a fundamental metric that compares explanations against a reference to ascertain their alignment with the model's behavior.
\emph{Fidelity} (or Faithfulness) is associated with how well the explanation approximates the prediction of the black box model. 
It is essential in post-hoc methods using surrogate models like LIME~\cite{ribeiro2016should} to assess how well the surrogate model approximates the black-box model's behavior locally. 
\emph{Fidelity} complements \emph{Correctness}, which evaluates the alignment between the surrogate model's predictions and the original model's predictions for the same input data.
\emph{Consistency} evaluates the stability of explanations across instances or datasets, where statistical measures like variance or standard deviation are utilized to ensure reliability under varying conditions. 
It evaluates whether functionally equivalent models produce similar explanations.
\emph{Stability} (or Robustness) assesses the resilience of explanations to input variations, scrutinizing changes induced by small (natural or adversarial) perturbations to the input data. 
\emph{Certainty} (or Uncertainty) quantifies the confidence level conveyed by explanations, employing metrics like entropy or confidence intervals to gauge reliability.
Lastly, \emph{Importance} elucidates the significance of features in explanations, using techniques such as feature importance scores or permutation importance to identify key elements driving model decisions.
Complementary, qualitative metrics use language, narratives, and contextual details to explain concepts, phenomena, or processes, providing insight into a model's behavior. 
The goal is to make the model's decisions understandable to humans by describing the rationale behind predictions in a comprehensible way~\cite{hosain2024explainable}.

Dedicated tools and frameworks have emerged to evaluate and improve model explainability and compliance with transparency requirements. 
For instance, ktrain simplifies the building and training of explainable models through an interface to Keras, supporting both text and image classification tasks. 
PyTorch leverages Captum for model interpretability, offering techniques such as smoothGrad and integrated gradients to understand the model behavior. 
AI Explainability 360 is an open-source library that provides explainable AI techniques for tabular, text-based, image, and time series data.
Arize AI provides an ML observability framework for model monitoring and assessment, aiding in diagnosing model outputs and detecting validation checks.
InterpretML, developed by Microsoft, facilitates the training of interpretable models and explaining black-box systems, offering a suite of methods for model explanation and comparison. 

\subsection{Fairness}
\label{subsec:tai:fairness}
Algorithmic Fairness is an emerging field that aims to ensure that the results of models are not unfairly influenced by sensitive attributes, thus avoiding perpetuating historical discrimination (e.g., based on ethnicity or gender)~\cite{mehrabi2021survey,caton2024fairness}.
Discrimination occurs when models treat differently similar individuals or groups of individuals just based on a sensitive, and not influent for the task, feature~\cite{oneto2024towards}.
Discrimination can be explicit (i.e., the model explicitly leverages the sensitive attribute to make decisions) or implicit (i.e., even if the sensitive attribute is unknown, the model possibly leverages a correlation to that in the data)~\cite{Oneto22TrustGraph}. 
An example of an unfair model might be one that unintentionally prioritizes hiring people of a specific gender for a specific job~\cite{amazonbias}.
Similarly, regarding \ITLogisticShort, an algorithm might favor delivery in specific, more lucrative areas or clients~\cite{loginextsolutions}. 
Fairness notions in AI can be broadly divided into two families: individual and group fairness definition~\cite{castelnovo2022clarification}. 
Individual notions of fairness ensure that similar individuals receive similar results, such as ensuring that two people with the same skills are equally likely to be hired, regardless of race or gender~\cite{caton2024fairness}. 
Group fairness aims to treat different demographic groups statistically equally, for example, by ensuring that the acceptance rates of financial loans are similar for white and black-skinned individuals~\cite{caton2024fairness}. 

Algorithmic discrimination is mainly due to two causes: data and model biases.
In fact, ML models are trained on historical data that are affected by societal biases (e.g., jobs usually performed by a certain gender or ethnicity).
Moreover, the concept of fairness varies over time (what is considered fair today could not be fair in a few years) and space (what is considered fair in a country could not be fair in another).
As a result, data may be affected by historical, representation, measurement, evaluation, temporal, etc. biases~\cite{mehrabi2021survey}.
Moreover, AI models are currently mostly based on ML and then statistical machines that tend to exacerbate biases, disregard minorities, and find spurious correlation~\cite{ye2024spurious}. 
Then, the combination of data and model biases can lead learning algorithms to either exaggerate existing biases or introduce new ones, resulting in discrimination~\cite{elzayn2019fair}. 

Three main approaches exist to address fairness in AI: pre-processing, in-processing, and post-processing techniques. 
Pre-processing techniques~\cite{caton2024fairness} involve cleaning the data to remove biases before training the model. 
This approach is helpful in making out-of-the-box ML algorithms fairer without altering the algorithms themselves. 
In-processing techniques~\cite{caton2024fairness} directly incorporate fairness constraints into the learning phase, ensuring fairness is embedded within the model's structure. 
Post-processing techniques~\cite{caton2024fairness} adjust the outputs of already trained models to ensure fairness.
This approach is advantageous when retraining or fine-tuning complex models is impractical. 
Each approach has advantages and applications, and often, a combination of these methods is employed to ensure comprehensive fairness in AI systems~\cite{caton2024fairness}.

\myparagraph{Metrics and Tools.}
In a recent work~\cite{islam2022through}, more than 20 different notions and measurements of fairness have been reviewed, categorizing them into group and individual fairness notions. 
Regarding the notion of group fairness, Demographic Parity, Equal Opportunity, and Equal Odds are surely the most used ones~\cite{barocas2023fairness}.
Demographic Parity checks that the positive outcome rate should be similar across different groups.
Equality of Opportunity checks that the subset of individuals in each subgroup who should receive a positive outcome actually have an equal chance of being predicted to receive that outcome. 
Equal Odds measure the prediction quality (e.g., accuracy) difference between the different subgroups. 
Individual fairness, in contrast, delves into the concept of treating similar individuals similarly, advocating for decisions that are independent of sensitive attributes. 
In this direction, metrics like Fairness Through Unawareness~\cite{barocas2023fairness}, Fairness Through Awareness~\cite{dwork2012fairness}, Counterfactual Fairness~\cite{kusner2017counterfactual}, and Path-specific Counterfactual Fairness~\cite{chiappa2019path} are the most known ones and search to measure why and how the decision would have changed by changing the sensitive attribute of the individual.

Because of the importance of the subject, various development environments and libraries for algorithmic fairness have been developed.
These tools include techniques for training, monitoring, and adjusting the fairness of such models.
Among them, Fairlearn~\cite{weerts2023fairlearn}, an open-source initiative, focuses on reducing biases in ML models to promote fairness and transparency. 
Developed by Microsoft Research, Fairlearn provides a suite of algorithms and visualization tools designed to help users understand and mitigate biases in their AI models. 
The toolkit includes techniques for fairness assessment, mitigation, and visualization, enabling data scientists and developers to analyze and address fairness issues in their models. 
AIF360~\cite{aif360-oct-2018}, another valuable tool, offers a Python toolkit dedicated to detecting and mitigating bias in AI models. 
Developed by the AI Fairness 360 team at IBM Research, AIF360 provides a comprehensive set of algorithms and metrics for assessing and mitigating bias across various stages of the machine learning pipeline. 
Lastly, Fairness Indicators~\cite{tensorflowFairnessIndicators} is a tool designed to help TensorFlow developers assess model performance across different data slices, identify biases, and understand the impact of model decisions on various demographic groups.

\subsection{Privacy}
\label{subsec:tai:privacy}
Privacy focuses on securing and protecting the data of individuals or groups from being disclosed or violated~\cite{Oneto22TrustGraph}. 
Privacy considerations within AI encompass safeguarding both training and test data (e.g., patient data in healthcare)~\cite{al2019privacy,liu2021machine} and the proprietary information inherent in the models themselves (e.g., model architecture and weights)~\cite{al2019privacy,liu2021machine}. 
There is an inherent tension between providing valuable services and exposing sensitive information; for example, location data or travel patterns can improve travel time or comfort but may also lead to potential breaches in user privacy~\cite{sucasas2016autonomous}.
Such breaches violate individual privacy rights and expose users to potential exploitation by malicious entities seeking to misuse or manipulate this data for nefarious purposes.
On the other hand, protecting the privacy of the model becomes essential for companies striving to maintain a competitive edge in the market. 
Malicious users can stage multiple attacks to leak information about both data and models.
The major ones include: (i) reconstruction~\cite{al2016reconstruction}, which aims to reconstruct raw data from pre-processed ones (e.g., after the feature engineering phase or anonymization); (ii) model inversion~\cite{fredrikson2015model}, which reconstructs the inputs of ML models that have generated a particular output; (iii) membership inference~\cite{shokri2017membership}, used to determine if a data record was used to train an ML model; and (iv) model discovery~\cite{orekondy2019knockoff}, which reconstructs the ML model architecture and weights.

Four major families of privacy-preserving ML exists: anonymization, perturbation, cryptographic techniques (CTs), and distributed protocols~\cite{liu2021machine,al2019privacy}.
Each method must consider three aspects: its ability to preserve privacy (i.e., the protection of individual or group information), its utility (i.e., the ability to extract meaningful information), and whether it requires a trusted curator to supervise and enforce the privacy-preserving ML method.
Anonymization methods~\cite{machanavajjhala2007diversity, sweeney2002k} aim to protect the privacy of data subjects by obscuring identifying information within a dataset while preserving data utility.
However, these methods often compromise utility to ensure confidentiality and are not suitable for preserving model privacy~\cite{oneto2024towards}.
Perturbation methods~\cite{abadi2016deep,arachchige2019local} exploit noise to corrupt the data, the ML algorithm, or the learned model to obscure sensitive information while allowing meaningful analysis.
Even in this case, there is also a trade-off between privacy and utility: increasing resilience against attacks or privacy breaches requires injecting more noise, which subsequently reduces utility. 
Furthermore, both anonymization and perturbation techniques assume the existence of a trusted data curator. 
When such a trusted entity is unavailable, cryptographic techniques and distributed protocols can be employed~\cite{oneto2024towards}.
CT approaches~\cite{aono2017privacy,fang2021privacy} consist of performing computations on encrypted data to protect both data and model privacy (i.e., privacy is preserved with no utility loss) at the cost of significant computational overhead. 
Distributed protocols~\cite{sattler2020clustered,hu2020personalized} involve multiple participants, each with private data, computing aggregated results while keeping their inputs secret.
Similar to CTs, these protocols preserve privacy and utility but at the cost of high computational and information exchange overhead.
To mitigate this, distributed protocols are often relaxed to federated methods, where only aggregated information is exchanged~\cite{liu2022distributed}.
Nevertheless, privacy can still be compromised when local parameters, such as updating gradients, are shared~\cite{lyu2022privacy}.
Thus, these protocols are often combined with perturbation and cryptographic methods to enhance privacy~\cite{oneto2024towards}.\\

\myparagraph{Metrics and Tools.}
The metrics employed in the context of privacy-preserving ML are mainly three: privacy-leakage metrics, data utility metrics, and computation metrics~\cite{supritha2018privacy,liu2021machine}.
Privacy-leakage metrics~\cite{agrawal2000privacy,agrawal2001design} measure the security of data assessing how much sensitive information can be inferred from transformed data or model after applying privacy-preserving methods.
While some privacy-leakage metrics can be theoretically computed, they are generally evaluated empirically by executing known attacks and calculating the resulting privacy leakage~\cite{shokri2017membership}. 
Data utility metrics~\cite{fletcher2015measuring,alvim2012differential} quantify the loss of information or utility due to privacy-preserving techniques. 
These metrics are generally assessed by comparing a model's performance (e.g., accuracy) on the original data versus the privacy-preserved data.
The aim is to determine how much the privacy-preserving transformation degrades the model's effectiveness.
Computation metrics~\cite{mohassel2017secureml,tan2021cryptgpu} focus on the computational overhead (e.g., pre-processing, training, and inference execution time, or memory requirements) introduced by the privacy-preserving techniques. \\
\indent Several popular tools and frameworks have been developed, each addressing different aspects of privacy-preserving ML techniques. 
ARX~\cite{prasser2020flexible} is an open-source library supporting data anonymization techniques (e.g., k-anonymity~\cite{sweeney2002k} and l-anonomity~\cite{machanavajjhala2007diversity}). 
Differential privacy tools, such as Google's Differential Privacy and TensorFlow Privacy, implement $\epsilon$-differentially private algorithms~\cite{wilson2019differentially, tensorflowTensorFlowPrivacy} and provide optimizers for training models with differential privacy, ensuring that individual data points remain protected during training. 
Homomorphic encryption~\cite{li2020privacy} is another key approach, with tools like Concrete-ML~\cite{ConcreteML} and Microsoft SEAL~\cite{sealcrypto} simplifying fully homomorphic encryption (FHE) for secure computations on encrypted data. 
Federated learning~\cite{zhang2021survey} frameworks such as FATE (Federated AI Technology Enabler)~\cite{fateHomeFATE}, FedML~\cite{githubGitHubFedMLAIFedML}, and Flower~\cite{beutel2020flower} enable distributed learning processes, facilitating collaborative data use while maintaining privacy. 
Confidential computing solutions like BastionLab enhance secure data science collaboration using confidential computing, access controls, and differential privacy techniques. 
Multi-Party Computation (MPC) frameworks, including PySyft~\cite{openminedQuickInstall} and Rosetta~\cite{Rosetta}, integrate secure, private deep learning into popular ML libraries like PyTorch and TensorFlow, separating private data from model training. 
Combined privacy solutions like OpenFL~\cite{openfl_citation} and Substra~\cite{substraSubstraDocumentation} offer comprehensive frameworks for secure, traceable, and collaborative machine learning. 
These tools collectively provide robust mechanisms for implementing privacy-preserving machine learning, ensuring data and model security throughout the AI lifecycle.



\subsection{Case Studies}\label{subsec:trustAI:casestudies}
This section presents real case studies in \ITLogisticShort that have implemented \TAI solutions to align their applications with regulatory and ethical principles~\cite{eu2021ai-act,european_ai_act_2021}. 
At first glance, our review reveals that most papers address only one or a few aspects of trustworthiness, focusing on its relevance within specific application domains and proposing algorithmic or system solutions to meet these needs. 
However, no existing works take a holistic approach to \TAI principles (as also noted in other research outside \ITLogisticShort~\cite{oneto2024towards}). 
Moreover, very few studies provide detailed information on the verification tools used, often relying on custom implementations. 
The use of standardized testing and verification tools provided by libraries-where quality control is typically stringent or supported by a large community-remains largely unexplored. 
Lastly, only a few articles address the relevance and integration of trustworthiness principles within \mlops pipelines.
This highlights a gap not only in the literature on the integration of \dataops and \mlops (as noted in \autoref{sec:mlops} and further elaborated in \autoref{sec:discussion}) but also in the scientific exploration of their alignment with tools designed to develop and monitor systems under trustworthiness guidelines.
In the following paragraphs, we present the case studies retrieved for each \ITLogisticShort area over the past 10 years, covering trustworthiness principles and highlighting complementary examples that particularly demonstrate the need for integrating them in real applications.

\myparagraph{Vehicle Control Systems.} 
This area includes 21 case studies focusing on privacy, XAI, robustness and sustainability for intelligent driving assistance systems~\cite{djenouri2023secure,sucasas2016autonomous,jiang2022trust,bylykbashi2019effect,vemoori2020towards,vemoori2022towards,mankodiya2022od,mankodiya2021xai,madhav2022explainable,berghoff2021robustness,unal2023towards,deng2020analysis,he2024trustworthy,pei2024dual,ahmad2019design,lee2022intelligent,ke2023lightweight,mahaur2023improved,chen2020denselightnet,khan2022lldnet,liang2022alodad,liu2022lightweight}. 
Djenouri et al.~\cite{djenouri2023secure} develop a blockchain-based framework, SISIV, to ensure data transmission privacy among vehicles and enable accurate traffic forecasting using distributed data. 
They validate their approach on four standard benchmark datasets for connected vehicles. 
Sucasas et al.~\cite{sucasas2016autonomous} design an authentication scheme for vehicle communication systems that preserves data privacy while requiring a trusted authority only once. 
Similarly, Jiang et al.~\cite{jiang2022trust} create secure data-sharing schemas to protect privacy in-vehicle networks, focusing on sustainability and privacy-preserving mechanisms to enhance trust and security. 
Bylykbashi et al.~\cite{bylykbashi2019effect} optimize vehicle communication for addressing traffic congestion, achieving fast and cost-efficient data transfer to improve road safety and passenger comfort. 
Vemoori et al.~\cite{vemoori2020towards} explore XAI's role in enhancing decision-making robustness for autonomous vehicles, testing computational intelligence frameworks. 
While they identify vulnerabilities to adversarial examples, they do not assess real-world performance. 
Extending this work, further studies by Vemoori et al.~\cite{mankodiya2022od,mankodiya2021xai,madhav2022explainable} investigate XAI applications in vehicle control systems. 
These studies emphasize interpretability in tasks such as obstacle detection, route planning, and system transparency, aiming to build trust and reliability in autonomous driving. 
On the robustness front, Berghoff et al.\cite{berghoff2021robustness} propose a systematic method for testing the reliability of AI-powered traffic sign recognition systems in autonomous vehicles under varying ambient conditions. 
Tested on the on the German Traffic Sign Recognition Benchmark.
Unal et al.\cite{unal2023towards} assess model robustness by generating challenging inputs that intentionally maximize prediction uncertainty. 
Deng et al.~\cite{deng2020analysis} present a systematic approach to evaluate the robustness of four autonomous driving models, testing them against five adversarial attack methods and four defense techniques using the Udacity dataset of real-world road images.
Xu et al.\cite{xu2022adversarial} introduce a graph-based approach to select more resilient models in automated machine learning pipelines, focusing on improving reliability of models during their development.
He et al.~\cite{he2024trustworthy} propose a defense-aware reinforcement learning technique to enhance the safety and robustness of autonomous vehicle policies, ensuring they can withstand adversarial attacks.
Lastly, there is a growing focus on improving the sustainability and efficiency of AI components in autonomous driving systems, as highlighted by several case studies~\cite{pei2024dual,ahmad2019design,lee2022intelligent,ke2023lightweight,mahaur2023improved,chen2020denselightnet,khan2022lldnet,liang2022alodad,liu2022lightweight}.
These works explore multiple techniques to reduce computational demands while preserving model performance.
For instance, Ke et al.~\cite{ke2023lightweight}, Mahaur et al.~\cite{mahaur2023improved}, and Pei et al.~\cite{pei2024dual} apply pruning methods to remove redundant model parameters, improving efficiency without compromising accuracy. 
Liu et al.~\cite{liu2022lightweight} further reduces model complexity using multi-modal tensor decomposition to compress the model lowering both parameter counts and computational costs.
Alternatively, works such as~\cite{ahmad2019design,lee2022intelligent,chen2020denselightnet,khan2022lldnet,liang2022alodad}, focus on designing smaller, lightweight network architectures that maintain strong performance through optimized structural design.

\myparagraph{Traffic Management.} 
In this area, we identified a total of 10 case studies focusing on traffic forecasting, route planning, and traffic control, accentuating interpretability, privacy, robustness, and sustainability principles~\cite{ou2024building,Liu2019IntelligentRP,xie2021explanation,wollenstein2020explainability,raja2024intuitive,ying2022privacysignal,tan2020robust,carvalho2020lightweight,chen2020edge,dai2022making}. 
Ou et al.~\cite{ou2024building} propose an interpretable traffic flow forecasting framework utilizing ensemble learning and tree-based algorithms. The framework is validated on traffic datasets collected from arterial roads, expressways, and freeways in China and the USA, demonstrating its flexibility across diverse scenarios. 
Liu et al.~\cite{Liu2019IntelligentRP} address route planning in large-scale road networks with a privacy-preserving, graph-based solution. Their approach ensures that cloud servers cannot deduce real user trajectories, safeguarding data privacy while providing efficient route recommendations. 
Wollenstein et al.~\cite{wollenstein2020explainability} present an interpretability tool for traffic light control systems that use deep neural networks. By analyzing historical state-action pairs, the tool offers actionable insights into system behavior, helping to reason about and improve the controller's decision-making. 
Raja et al.~\cite{raja2024intuitive} develop an intuitive and privacy-preserving traffic light control system for autonomous vehicles. Their system minimizes privacy risks associated with data collection, ensuring secure and efficient traffic flow management. 
Ying et al.~\cite{ying2022privacysignal} design PrivacySignal, a privacy-preserving traffic signal control system for intelligent transportation. The system enables vehicles to share data with roadside units for signal optimization without exposing sensitive information, maintaining user privacy while enhancing traffic efficiency. 
On the robustness frontier, Tan et al.~\cite{tan2020robust} propose a robust training framework for reinforcement learning (RL) agents in noisy traffic environments. 
Evaluating their model with the SUMO traffic simulator~\cite{krajzewicz2002sumo}, the authors demonstrate improved policy robustness under varying levels of data noise, highlighting the framework's potential for real-world urban mobility scenarios. 
Lastly, we observe multiple case studies related to the sustainability and need for efficient ML models in production, especially in traffic control applications. 
Carvalho et al.~\cite{carvalho2020lightweight} propose a lightweight design strategy to decrease the execution time of the vehicle identification model they develed and tested on Brazilian vehicle roads. 
Chen et al.~\cite{chen2020edge} and Dai et al.~\cite{dai2022making}, in a similar manner, develop lightweight models to accelerate the detection of traffic flow under the premise of high precision on real-word traffic scenes in the UA-DETRAC dataset~\cite{wen2020ua}.

\myparagraph{Transportation Management Systems.} 
In this area, we identified a total of 6 case studies addressing privacy-preserving frameworks, explainable predictions, and integrative approaches to manage public and safety-critical transportation systems~\cite{zeller2023safety,zeller2024toward,alic2019bigsea,monje2022deep,sobrie2024real,sariyer2024leveraging}. 
Zeller et al.~\cite{zeller2023safety} emphasize safe \mlops practices for transportation systems, focusing on aligning workflows with safety regulations to ensure secure ML deployment. While the study offers detailed safety guidelines, it does not extend to other trustworthiness principles or propose an integrated \mlops framework. 
Expanding on this, Zeller et al.~\cite{zeller2024toward} introduce a conceptual \mlops framework for trustworthy ML in transportation, addressing principles like reliability, safety, and continuous verification. 
However, the study lacks practical implementations or specific tools to demonstrate its applicability. 
The BIGSEA project~\cite{alic2019bigsea} introduces a privacy-preserving \dataops framework for public transportation systems by leveraging k-anonymity techniques~\cite{sweeney2002k} to protect user data during processing. 
The approach integrates privacy-conscious practices into \dataops workflows, enhancing data trustworthiness in transportation management.
Monje et al.~\cite{monje2022deep} develop a deep learning-based explainable AI model for bus passenger forecasting, using extensive historical data from Madrid's municipal transport system. Their model combines prediction accuracy with linguistic interpretability, offering actionable insights for decision-makers. 
Sobrie et al.~\cite{sobrie2024real} design a support system to provide explainable predictions and personalized recommendations for railway control room operators.
The final system has been successfully tested for its effectiveness in real-time operations at Infrabel's railway control rooms in Belgium.
Finally, Sariyer et al.~\cite{sariyer2024leveraging} present a XAI-driven model to predict and interpret public transportation usage rates, incorporating data from buses, ferries, and railways. 
Tested on an open dataset from Izmir City, Turkey, the model identifies key features like passenger types and special calendar days, demonstrating its versatility in analyzing and forecasting multimodal transportation usage trends. 

\myparagraph{Traveler Information and Experience Systems.}
In this area, we identified a total of 4 case studies, primarily focusing on addressing user privacy and sustainability concerns when leveraging data for tourism support~\cite{shao2024spatial,munawar2024collaborative,cai2023personalized,zhuang2022edge,li2021feature}. 
Indeed, the tourism industry faces complex regulatory and privacy challenges, which can limit the application of ML to customer data~\cite{nunez2024machine}. 
Shao et al.~\cite{shao2024spatial} utilize explainable AI to help stakeholders understand why regions like Shanghai achieve the most effective integration between cultural heritage resources and the tourism industry, providing actionable insights to address disparities in other regions. 
Munawar et al.~\cite{munawar2024collaborative} propose a federated learning framework for autonomous taxi demand forecasting, enabling collaborative model training across regions while preserving passenger privacy. 
Using a real-world dataset from users in Bangkok, the authors demonstrate that the framework can provide a scalable and privacy-preserving solution for smart city transportation systems.
Similarly, Cai et al.~\cite{cai2023personalized} employ federated learning to develop a personalized travel recommendation system for cultural tourism attractions in Fujian-Taiwan.
Regarding sustainability, Zhuang et al.~\cite{zhuang2022edge} address the resource constraints of edge devices in parking surveillance systems. 
They propose a quantized model deployed on edge devices to reduce computational complexity while maintaining accuracy, minimizing data transmission latency and central server load. 

\color{black}
\myparagraph{Freight and Supply Chain Logistics.}
In this area, we identified 12 case studies addressing different dimensions of trustworthiness across the supply chain, including security and privacy in fleet and freight management~\cite{akram2017security,ekatpure2023artificial,belletti2017privacy,yao2024sensitive,di2023machine}, fairness in delivery operations~\cite{lu2023value,chen2023same,gupta2022fairfoody}, and sustainability through lightweight model design~\cite{ye2022stable,hussain2023custom,zhang2024lightweight,li2021feature}.

On the privacy and security frontier, Akram et al.~\cite{akram2017security} evaluate security, privacy, and safety issues for both dynamic and static fleets of logistics drones, highlighting the challenges of ensuring robust protections in drone-based operations, a rapidly expanding area of logistics management.
Ekatpure et al.~\cite{ekatpure2023artificial} examine data-security and privacy challenges in AI-powered fleet management systems, emphasizing the need for robust privacy-preserving mechanisms as fleets increasingly rely on AI for decision-making and operational efficiency.
Belletti et al.~\cite{belletti2017privacy} propose privacy-preserving mechanisms for fleet management in Mobility-as-a-Service systems, while Yao et al.~\cite{yao2024sensitive} apply $\varepsilon$-differential privacy techniques to protect sensitive data in logistics operations, including cargo activities. Their work illustrates how differential privacy can safeguard carrier data while allowing intelligent logistics systems to operate efficiently and comply with data-protection regulations.
Di et al.~\cite{di2023machine} extend privacy considerations to warehousing logistics by exploring privacy-preserving federated learning, enabling multiple stakeholders to collaboratively train ML models while keeping their data decentralized and private.

On the fairness frontier, Lu et al.~\cite{lu2023value} introduce fairness-aware delivery systems that optimize delivery processes while ensuring that delivery times and costs are distributed equitably between service providers and customers.
Chen et al.~\cite{chen2023same} adopt a similar perspective in same-day delivery systems, using reinforcement learning to balance fairness and efficiency while reducing inequalities in service experience.
Gupta et al.~\cite{gupta2022fairfoody} address fairness in food-delivery platforms, with a particular focus on the income distribution of gig workers.
By analyzing data from real-world food-delivery platforms in India, they propose FairFoody, a system designed to ensure fairer income distribution among delivery agents while preserving timely deliveries, achieving up to a tenfold improvement in equitable pay over baseline strategies with limited impact on customer satisfaction.

On the sustainability frontier, Ye et al.~\cite{ye2022stable} propose a lightweight model to reduce prediction time for timely obstacle detection on railway lines, addressing a major safety threat to train operations.
Similarly, Hussain et al.~\cite{hussain2023custom} and Zhang et al.~\cite{zhang2024lightweight} propose lightweight ML models to reduce the computational footprint of edge-deployed systems for fast automated object detection in smart warehouses, while Li et al.~\cite{li2021feature} introduce a lightweight model for predicting liner handling volumes at container terminals, balancing model complexity and predictive accuracy.
These works emphasize the practical need for lightweight and sustainable models, which are often deployed on edge devices with limited computational resources across the supply chain.
\color{black}

\myparagraph{Emergency Management.}
In this area, we found 2 case studies focusing on fairness in resource allocation and emergency response~\cite{elzayn2019fair,aringhieri2022fairness}.
Elzayn et al.~\cite{elzayn2019fair} explore fairness in allocation problems, applying their allocation and learning algorithms to the publicly available Philadelphia Crime Incidents dataset. 
Their approach demonstrates how fair algorithms can effectively address societal challenges by ensuring equitable resource distribution in emergency scenarios. 
While the study provides a practical application of fairness principles, it does not consider other aspects of trustworthiness, such as robustness or privacy.
Building on this, Aringhieri et al.~\cite{aringhieri2022fairness} extend fairness considerations to ambulance routing in post-disaster management. 
They incorporate fairness constraints into their routing algorithms to ensure equitable access to emergency services during crises. 
The proposed methodology is tested on synthetically generated data, highlighting its potential for practical deployment in real-world disaster scenarios. 

\color{black}
\myparagraph{Remarks.} 
Overall, the reviewed case studies indicate that \TAI research in \ITLogisticShort remains more fragmented than holistic, with most contributions addressing one or a few trustworthiness principles rather than their joint integration into operational pipelines.
Across domains, the studies reveal a shared need for more systematic support to verify, monitor, and document trustworthiness properties, while also showing that each application area introduces additional domain-specific requirements.

Vehicle Control Systems constitute the most active area of \TAI research, with substantial attention devoted to privacy, robustness, explainability, and sustainability.
In this domain, the main challenge is to integrate these principles into safety-critical ML lifecycles that must operate under changing road conditions, distribution shifts, and adversarial perturbations.
Traffic Management applications primarily highlight the need to balance privacy, robustness, and sustainability, since algorithmic decisions directly influence road efficiency, congestion, environmental impact, and users' mobility experience.
Transportation Management Systems emphasize explainability, privacy preservation, and safety compliance, especially in public-facing and regulated settings where data-driven decisions must remain transparent, auditable, and aligned with operational constraints.
Traveler Information and Experience Systems mainly focus on privacy-preserving and sustainable ML solutions, reflecting both the sensitivity of user mobility and tourism data and the need for efficient deployment on resource-constrained devices.
Freight and Supply Chain Logistics provide one of the broadest views of \TAI requirements, spanning privacy and security in fleet operations, fairness in delivery services, and sustainability in edge-deployed models.
Finally, Emergency Management remains comparatively underrepresented in the \TAI literature; however, the available studies clearly show the importance of fairness in resource allocation and emergency response, where algorithmic decisions can directly affect vulnerable populations and high-stakes public-safety outcomes.
\color{black}

\section{Open Problems and Future Perspectives}
\label{sec:discussion}
\begin{table*}[t]
\centering
\renewcommand{\arraystretch}{1.2}
\setlength{\tabcolsep}{.1cm}
\caption{Case studies summary for \dataops, \mlops, and \TAI. }
\begin{tabular}{|c||c||c||c||c||c|c|c|c|c|}
\hline
\hline
\multirow{2}{*}{\textbf{Areas}} & \multirow{2}{*}{\textbf{Ref(s)}} &  \multirow{2}{*}{\textbf{\# Ref(s)}} & \multirow{2}{*}{\textbf{\dataops}} & \multirow{2}{*}{\textbf{\mlops}} & \multicolumn{5}{c|}{\textbf{\TAI}} \\
&&&&& \textbf{Sustain.} & \textbf{Robust.} & \textbf{XAI} & \textbf{Fairness} & \textbf{Privacy} \\
\hline
\hline

\multirow{9}{*}{Vehicle Control} & \cite{alexakis2022distributed,zhang2017carstream,kahveci2022end,dinten2023fleet,zhang2017carstream,xu2016supporting,nair2016real,sbai2020real,triguero2017vehicle,ademola2022improving,yoo2020implementation,priyanka2021real} & 12 &  \tcheck & & & & & & \\
\cline{2-10}
& \cite{hartmannsgruber2024improving,steffen2023continuous,weber2024streamlining} & 3  &  & \tcheck & & & & & \\
\cline{2-10}
&\cite{pei2024dual,ahmad2019design,lee2022intelligent,ke2023lightweight,mahaur2023improved,chen2020denselightnet,khan2022lldnet,liang2022alodad,liu2022lightweight}& 9  &  & & \tcheck &  & & & \\
\cline{2-10}
& \cite{berghoff2021robustness,unal2023towards,deng2020analysis,xu2022adversarial,he2024trustworthy} & 5  &  & & & \tcheck & & & \\
\cline{2-10}
& \cite{vemoori2020towards,mankodiya2022od,mankodiya2021xai,madhav2022explainable} & 4  &  & & & & \tcheck & & \\
\cline{2-10}
&\cite{djenouri2023secure,sucasas2016autonomous,jiang2022trust,bylykbashi2019effect} & 4  &  &  & & & & & \tcheck\\
\cline{2-10}
& \cite{ferreira2024mlops}& 1  & \tcheck & \tcheck & &  & & & \\
\cline{2-10}
& \cite{socha2022smirk}& 1  & \tcheck & & & \tcheck & & & \\
\cline{2-10}
& \emph{Conceptual}:~\cite{mishra2023autonomous}& 1  & \tcheck & \tcheck & & \tcheck & & & \tcheck\\
\hline\hline

\multirow{7}{*}{Traffic Management} & \cite{gillani2023machine,amini2017big,sinnott2015smash,anastasiou2019admsv2,guerreiro2016architecture,figueiras2018data,daaif2019efficient,khazaei2016realtime,tseng2018congestion,khalajzadeh2020end,sundareswaran2020real,yu2020dissecting,fu2019demonstrating,babar2019real,tarneberg2016experiences,rathore2016exploiting,castellanos2021smart,vonitsanos2021forecasting,chen2017modeling} & 19 & \tcheck & & & & & & \\
\cline{2-10}
& \cite{angarita2020general,oakley2023foresight} & 2 &  & \tcheck & & & & & \\
\cline{2-10}
& \cite{carvalho2020lightweight,chen2020edge,dai2022making} & 3 & & & \tcheck & & & & \\
\cline{2-10}
& \cite{ou2024building,wollenstein2020explainability}& 2 & & & & & \tcheck & & \\
\cline{2-10}
& \cite{tan2020robust}& 1 & & & & \tcheck & &  & \\
\cline{2-10}
& \cite{Liu2019IntelligentRP,raja2024intuitive,ying2022privacysignal}& 3 & & & & & & & \tcheck\\
\cline{2-10}
&\cite{conlan2023real}& 1 & \tcheck & \tcheck & & & & & \\
\hline\hline

\multirow{5}{*}{Transportation Management} & \cite{qi2020framework,ji2017low,leung2019urban,anastasiou2019admsv2,fiore2019integrated,rathore2021cyber,antunes2019big,luckner2020iot,tarneberg2016experiences,oneto2018train,vonitsanos2021forecasting,vo2022practical,mulerikkal2021jp}& 13 &  \tcheck & & & & & & \\
\cline{2-10}
&\cite{zeller2024toward,sengul2022application,guzhva2023evaluating}& 3  &  & \tcheck & & & & & \\
\cline{2-10}
&\cite{monje2022deep,sobrie2024real,sariyer2024leveraging}& 3 & & & & & \tcheck & & \\
\cline{2-10}
&\cite{zeller2023safety,zeller2024toward}& 2& &\tcheck & & \tcheck & & & \\
\cline{2-10}
&\cite{alic2019bigsea}& 1& \tcheck & & & & & &\tcheck \\
\hline\hline

\multirow{4}{*}{\begin{tabular}[c]{@{}l@{}}Traveler Information and\\ Experience Systems \end{tabular}} &  \cite{al2021hybrid,chi2021design,asaithambi2023thematic,lapatta2022ecotourism,ambolkar2022hotel,malkawi2023real,bellandi2019designing,petrozziello2017data,suanpang2020integration} & 9 &\tcheck & & & & & & \\
\cline{2-10}
& \emph{Conceptual}:~\cite{urias2023mlops}  & 1 &  & \tcheck & & & & & \\
\cline{2-10}
&\cite{zhuang2022edge} & 1 & & & \tcheck & & & & \\
\cline{2-10}
&\cite{shao2024spatial,munawar2024collaborative,cai2023personalized} & 3 & & & & & & & \tcheck\\
\hline\hline


\color{black}
\multirow{7}{*}{\begin{tabular}[c]{@{}c@{}}Freight and\\ Supply Chain\\ 
Logistics\end{tabular}} & 
\cite{kumar2018design,Hadavi2019MonitoringUT,dimokas2021cloud,
belfadel2023conceptual,selvaraj2020offline,provatas2020selis,
mhand2019scalable,de2021simulation,sarabia2019seaport,
silva2021advancing,ruan2021filling,lee2018design,sugathan2021espy,hong2019crowdsourcing} 
& 14 & \tcheck & & & & & & \\
\cline{2-10}
& \cite{faubel2024mlops,cha2023development,morariu2024analysing,
khan2024optimizing,jana2023framework,michelangelo2024uber} & 6 
& & \tcheck & & & & & \\
\cline{2-10}
& \cite{pineda2023mlops} & 1 & \tcheck & \tcheck & & & & & \\
\cline{2-10}
& \cite{ye2022stable,hussain2023custom,zhang2024lightweight,
li2021feature} & 4 & & & \tcheck & & & & \\
\cline{2-10}
& \cite{akram2017security,ekatpure2023artificial} & 2 & & & & \tcheck 
& & & \tcheck \\
\cline{2-10}
& \cite{belletti2017privacy,yao2024sensitive,di2023machine} & 3 
& & & & & & & \tcheck \\
\cline{2-10}
& \cite{lu2023value,chen2023same,gupta2022fairfoody} & 3 & & & & & 
& \tcheck & \\
\hline\hline
\color{black}

\multirow{3}{*}{Emergency Management} & \cite{alomari2019road,anastasiou2019admsv2,pandhare2017real,alomari2021iktishaf,seal2019real,kovalchuk2018distributed,rathore2016real,alotaibi2020sehaa,noori2020fire,huamani2020machine}& 10 &  \tcheck & & & & & & \\
\cline{2-10}
& \cite{peramo2022intelligent}& 1  &  & \tcheck & & & & & \\
\cline{2-10}
& \cite{elzayn2019fair,aringhieri2022fairness} & 2 &  &  & & & & \tcheck& \\

\hline\hline

\multicolumn{1}{c||}{} & \multicolumn{1}{r||}{\textbf{\# Ref(s)}} & 153 & 82 & 19 & 17 & 12 & 9 & 5 & 17\\

\hhline{~=========}
\end{tabular}
\label{tab:summary:casestudy:MLDataTAI}
\end{table*}

\color{black}
In the previous sections, we examined the integration of \dataops, \mlops, and \TAI principles in the context of \ITLogisticShort.
We showed that, together, these perspectives support the development and operation of trustworthy AI-enabled services by addressing complementary aspects of the data, model, and governance lifecycles.
Building on this integrated perspective, we now synthesize the main insights, identify open challenges, and outline future research directions in this domain.
\color{black}
To assist readers, \autoref{tab:summary:casestudy:MLDataTAI} summarizes the case studies presented in Sections~\ref{subsec:dataops:casestudy}, \ref{subsec:mlops:casestudy}, and~\ref{subsec:trustAI:casestudies}.
Notably, this table offers a visual overview of the current state of \mlops, \dataops, and \TAI integration in \ITLogisticShort, thereby underscoring important unresolved challenges and suggesting avenues for further investigation.

\myparagraph{Lack of a Unified Framework for \dataops, \mlops, and \TAI Integration.}
A major challenge in the AI lifecycle for \ITLogisticShort is the lack of a single framework that effectively integrates \dataops, \mlops, and \TAI principles.
As shown in \autoref{tab:summary:casestudy:MLDataTAI}, no existing approach comprehensively addresses all three domains within a unified solution; most efforts focus on these practices either separately or in pairs.
For instance, several \dataops frameworks for large-scale applications, including those from Amini et al.~\cite{amini2017big} and Gillani et al.~\cite{gillani2023machine}, primarily address data automation but overlook facets such as data quality, security, defenses against data poisoning attacks~\cite{cina2023wild}, and fairness considerations~\cite{clifford2018data}.
Other projects, such as the privacy-preserving \dataops BIGSEA~\cite{alic2019bigsea} initiative and the end-to-end pipelines by Pineda et al.~\cite{pineda2023mlops} and Conlan et al.~\cite{conlan2023real}, incorporate at least two of these domains and have demonstrated promising results in real-world large-scale environments. 
However, such integrative approaches remain the exception rather than the norm.
A truly comprehensive framework that unites \dataops, \mlops, and \TAI principles is still missing, which presents a valuable opportunity for advancing the efficiency, reliability, and trustworthiness of Data and AI systems.
Although Mishra et al.~\cite{mishra2023autonomous} propose a conceptual model combining \dataops, \mlops, and safety practices, their work lacks concrete implementation guidelines, further underscoring the need for robust, unified solutions in this space.

\myparagraph{Limited Adoption to \mlops and Emerging Needs for \llmops in \ITLogisticShort.}
Compared to \dataops, which is now widely adopted for large-scale data pipelines (as evidenced by the higher number of case studies), the use of \mlops in \ITLogisticShort remains comparatively limited.
The absence of standardized practices often leads to custom ML solutions that fail to scale effectively or ensure long-term reliability~\cite{haertel2023mlops}.
For example, Alomari et al.~\cite{alomari2019road} and Alotaibi et al.~\cite{alotaibi2020sehaa} combine \dataops solutions with ML algorithms but do not rely on \mlops principles.

This gap is exacerbated by the growing adoption of LLMs, which deliver strong performance but introduce greater complexity in lifecycle governance, making \llmops practices increasingly important.
Nevertheless, there is limited evidence of publicly documented \llmops implementations within \ITLogisticShort.
Consequently, the underutilization of \mlops and \llmops offers a compelling avenue for future research, where systematic integration of these practices could enhance real-time data processing, model management, and scalability in \ITLogisticShort contexts.

\myparagraph{Unequal and Non-Holistic Attention to \TAI Principles.}
The adoption of \TAI principles in \ITLogisticShort applications is often imbalanced, with some metrics receiving far more attention than others.
Privacy considerations, driven by regulations such as GDPR~\cite{azam2022data} and the EU AI Act~\cite{ai-act-council,eu2021ai-act,european_ai_act_2021}, frequently take precedence, prompting the development of privacy-preserving technologies (e.g., data anonymization and consent management)~\cite{djenouri2023secure,sucasas2016autonomous}. 
Sustainability has also gained substantial traction in recent years, particularly in response to green-transition policies~\cite{fetting2020european}, fueling efforts to reduce the computational overhead of increasingly large models for deployment on resource-constrained edge systems~\cite{pei2024dual,ahmad2019design,ke2023lightweight}.
Robustness, while supported by numerous case studies, tends to be explored primarily in the context of natural data corruption (e.g., sensor failures), with limited focus on adversarial attacks that deliberately exploit model vulnerabilities~\cite{cina2023wild}.
Meanwhile, fairness and explainability dimensions have received comparatively less scrutiny, despite their growing relevance under evolving regulatory frameworks such as the AI Act~\cite{eu2021ai-act}. 
Moreover, \TAI principles are often tackled individually, with research typically optimizing one metric in isolation without assessing its broader impact on the others.
However, regulatory requirements call for a comprehensive approach that addresses \TAI dimensions holistically. 
Consequently, the deployment of ML systems must concurrently ensure compliance with all relevant \TAI principles, underscoring the need for novel methodologies and tools that seamlessly integrate these considerations throughout the AI lifecycle.
Currently, fairness and privacy emerge as recurring gaps across multiple sectors, underscoring the need for more balanced and comprehensive research efforts.

\myparagraph{Unbalanced Coverage of Case Studies Across Areas.}
The analysis of case studies in \ITLogisticShort from \autoref{tab:summary:casestudy:MLDataTAI} highlights an uneven distribution of research focus and exploration.
While most work concentrates on traffic management and vehicle control, significant gaps exist in domains such as emergency management and warehousing and delivery.
This imbalance suggests that several high-impact areas remain underexplored, despite clear opportunities for \dataops, \mlops, and \TAI principles to orchestrate, deploy, and maintain AI-driven solutions.
In fields like emergency management, strict confidentiality constraints often prevent data from being shared openly.
As a result, large-scale empirical validations and detailed methodological publications become difficult to pursue~\cite{larsson2023data}.
Organizations may be unwilling to release sensitive datasets out of concern for confidentiality breaches and regulatory noncompliance.
Nevertheless, these very domains, particularly critical infrastructures and public services, stand to benefit from \dataops, \mlops pipelines, and \TAI principles, which could enhance robustness, reproducibility, and trustworthiness in current systems.
In conclusion, the current distribution of case studies in \ITLogisticShort exhibits a clear imbalance in research emphasis.
While high-profile sectors receive greater scrutiny and benefit from more advanced AI implementations, other equally or even more impactful domains garner insufficient attention.
Bridging these gaps requires collaborative efforts among academia, industry, and policymakers to foster transparent data-sharing initiatives, robust methodological frameworks, and \TAI standards that can be responsibly applied across every dimension of intelligent transportation and logistics.

\color{black}
\section{Conclusions}
\label{sec:Conclusions}
In this paper, we examine the role of \dataops, \mlops, and \TAI in modern \ITLogisticShort, highlighting how these complementary paradigms support the development of reliable, scalable, and trustworthy AI-enabled services.
Through the analysis of tools, methodologies, and real-world case studies, we show that \dataops provides the foundation for managing large-scale, heterogeneous data pipelines, while \mlops enables the systematic development, deployment, monitoring, and maintenance of ML models in production environments.
Together, these capabilities are fundamental for addressing critical operational challenges in \ITLogisticShort applications.

Beyond these technical foundations, we emphasize the growing need to build \TAI-oriented solutions that ensure such systems remain robust, explainable, fair, sustainable, and privacy-preserving throughout their lifecycle.
Our review underscores that trustworthiness extends beyond model performance, requiring additional metrics and dimensions to be continuously considered.
In this context, adopting \dataops and \mlops is not merely a matter of operational convenience, but a strategic requirement for ensuring trustworthiness in AI-driven \ITLogisticShort.
However, our analysis reveals that, while \dataops solutions are reaching a relatively mature stage of adoption, the systematic adoption of \mlops remains more limited, and the integration of \dataops, \mlops, and \TAI within a unified framework is still uncommon in real-world deployments.

Looking ahead, the rapid evolution of AI creates new opportunities for \ITLogisticShort across its diverse application domains.
In particular, the growing adoption of LLMs, generative AI, and multimodal systems is extending traditional \mlops practices toward \llmops, while recent advances in autonomous and agentic systems are motivating the development of \agentops.
These developments substantially increase the capabilities of AI-enabled services, facilitating more advanced interaction, reasoning, planning, and autonomous decision support.
However, they also increase the technical debt associated with AI systems by introducing additional layers of models, tools, data sources, and software components that must be developed, monitored, maintained, and governed throughout their lifecycle.
As a result, ensuring trustworthiness becomes increasingly challenging as AI systems grow in complexity.
In this context, \dataops, \mlops, \llmops, and future \agentops frameworks are expected to play a fundamental role in automating operational processes and enabling organizations to manage increasingly sophisticated AI systems at scale.
More broadly, the evolution of AI in \ITLogisticShort is likely to shift the focus from developing isolated models toward governing complete data- and AI-driven ecosystems, with future services expected to integrate increasingly complex combinations of data pipelines, AI models, external services, autonomous agents, and human decision-makers across the entire operational lifecycle.

Future progress will therefore depend not only on improving model capabilities, but also on developing methodologies, tools, and standards that ensure these systems remain reliable, transparent, sustainable, and aligned with trustworthiness requirements throughout their lifecycle.
From this perspective, tighter integration of \dataops, \mlops, and \TAI principles represents a promising direction for supporting the next generation of intelligent transportation and logistics services.
We hope this work provides a useful reference for researchers, practitioners, and decision-makers, helping them navigate the evolving landscape of data and AI in \ITLogisticShort while offering a foundation for future research and trustworthy AI adoption.
\color{black}

\section*{Acknowledgments}
\noindent 
This work is partially supported by (i) EU - NGEU National Sustainable Mobility Center (CN00000023) Italian Ministry of University and Research Decree n. 1033-17/06/2022 (Spoke 10), (ii) project SERICS (PE00000014) under the NRRP MUR program funded by the EU - NGEU, (iii) project FAIR (PE00000013) under the NRRP MUR program funded by the EU - NGEU, and (iv) project RAISE (ECS00000035) under the NRRP MUR program funded by the EU - NGEU.

\bibliographystyle{IEEEtran}
\bibliography{main}

@misc{european_ai_act_2021,
	howpublished = {\url{https://eur-lex.europa.eu/legal-content/EN/TXT/?uri=celex%3A52021PC0206}},
	note = {[Accessed 15-12-2023]},
	title = {Proposal for a Regulation laying down harmonised rules on Artificial Intelligence and amending certain union legislative acts}}

@misc{michelangelo2024uber,
	howpublished = {\url{https://www.uber.com/en-IT/blog/michelangelo-machine-learning-platform/}},
	note = {[Accessed 13-10-2024]},
	title = {{M}eet {M}ichelangelo: {U}ber's {M}achine {L}earning {P}latform}}

@misc{Edelman2024,
	howpublished = {\url{https://www.edelman.com/sites/g/files/aatuss191/files/2019-02/2019_Edelman_Trust_Barometer_Global_Report.pdf}},
	note = {[Accessed 15-02-2025]},
	title = {Trust Barometer}}

@misc{LangChain,
	howpublished = {\url{https://www.langchain.com/}},
	note = {[Accessed 16-06-2026]},
	title = {{L}ang{C}hain}}

@misc{Deepchecks,
	howpublished = {\url{https://github.com/deepchecks/deepchecks}},
	note = {[Accessed 16-06-2026]},
	title = {{D}eepchecks}}

@misc{benoit_courty_2024_11171501,
	howpublished = {\url{https://github.com/mlco2/codecarbon}},
	note = {[Accessed 15-02-2025]},
	title = {{CodeCarbon}}}

@misc{loginextsolutions,
	howpublished = {\url{https://www.loginextsolutions.com/blog/logistics-route-optimization-software-bias-fix-solution/}},
	note = {[Accessed 15-02-2025]},
	title = {LogiNext}}

@misc{amazonbias,
	howpublished = {\url{https://www.reuters.com/article/idUSKCN1MK0AG/}},
	note = {[Accessed 15-02-2025]},
	title = {Amazon scraps secret AI recruiting tool that showed bias against women}}

@misc{art2018,
	howpublished = {\url{https://github.com/Trusted-AI/adversarial-robustness-toolbox}},
	note = {[Accessed 15-02-2025]},
	title = {Adversarial Robustness Toolbox}}

@article{larsson2023data,
	author = {Larsson, A. and Gro{\ss}e, C.},
	journal = {Journal of Risk Research},
	number = {5},
	pages = {524-546},
	title = {Data use and data needs in critical infrastructure risk analysis},
	volume = {26},
	year = {2023}}

@article{Li2021TrustworthyAF,
	author = {Li, B. and Qi, P. and Liu, B. and Di, S. and Liu, J. and Pei, J. and Yi, J. and Zhou, B.},
	journal = {ACM Computing Surveys},
	number = {9},
	pages = {1-46},
	title = {Trustworthy AI: From Principles to Practices},
	volume = {55},
	year = {2021}}

@book{garg2013apache,
	author = {Garg, N.},
	publisher = {Packt Publishing Birmingham, UK},
	title = {Apache kafka},
	year = {2013}}

@article{winfield2019machine,
	author = {Winfield, A. F. and Michael, K. and Pitt, J. and Evers, V.},
	journal = {Proceedings of the IEEE},
	number = {3},
	pages = {509-517},
	title = {Machine ethics: The design and governance of ethical AI and autonomous systems [scanning the issue]},
	volume = {107},
	year = {2019}}

@article{allen2000prolegomena,
	author = {Allen, C. and Varner, G. and Zinser, J.},
	journal = {Journal of Experimental \& Theoretical Artificial Intelligence},
	number = {3},
	pages = {251-261},
	title = {Prolegomena to any future artificial moral agent},
	volume = {12},
	year = {2000}}

@article{anderson2018geneth,
	author = {Anderson, M. and Anderson, S. L.},
	journal = {Paladyn, Journal of Behavioral Robotics},
	number = {3},
	pages = {251-261},
	title = {GenEth: A general ethical dilemma analyzer},
	volume = {12},
	year = {2018}}

@inproceedings{OnetoC128,
	author = {Franco, D. and Oneto, L. and Anguita, D.},
	booktitle = {European Symposium on Artificial Neural Networks, Computational Intelligence and Machine Learning},
	title = {Mitigating Robustness Bias: Theoretical Results and Empirical Evidences},
	year = {2023}}

@article{oneto2024towards,
	author = {Oneto, L. and Ridella, S. and Anguita, D.},
	journal = {Neurocomputing},
	title = {Towards algorithms and models that we can trust: A theoretical perspective},
	volume = {In Press},
	year = {2024}}

@inproceedings{huang2017safety,
	author = {Huang, X. and Kwiatkowska, M. and Wang, S. and Wu, M.},
	booktitle = {International Conference on Computer Aided Verification},
	title = {Safety verification of deep neural networks},
    pages="3--29",
	year = {2017}}

@article{Cin2022MachineLS,
	author = {Cin{\`a}, A. E. and Grosse, K. and Demontis, A. and Biggio, B. and Roli, F. and Pelillo, M.},
	journal = {Computer},
	number = {3},
	pages = {26-34},
	title = {Machine learning security against data poisoning: Are we there yet?},
	volume = {57},
	year = {2024}}

@article{novakovic2017evaluation,
	author = {Novakovi{\'c}, J. D. and Veljovi{\'c}, A. and Ili{\'c}, S. S. and Papi{\'c}, {\v{Z}}. and Tomovi{\'c}, M.},
	journal = {Theory and Applications of Mathematics \& Computer Science},
	number = {1},
	pages = {39},
	title = {Evaluation of classification models in machine learning},
	volume = {7},
	year = {2017}}

@article{chang2023survey,
	author = {Chang, Y. and Wang, X. and Wang, J. and Wu, Y. and Others},
	journal = {ACM Transactions on Intelligent Systems and Technology},
	number = {3},
	pages = {1-45},
	title = {A survey on evaluation of large language models},
	volume = {15},
	year = {2024}}

@inproceedings{dautov2017quantifying,
	author = {Dautov, R. and Distefano, S.},
	booktitle = {IEEE International Conference on Big Data},
	title = {Quantifying volume, velocity, and variety to support (Big) data-intensive application development},
	year = {2017}}

@article{bharadiya2023machine,
	author = {Bharadiya, J. P.},
	journal = {International Journal of Computer},
	number = {1},
	pages = {123-134},
	title = {Machine learning and AI in business intelligence: Trends and opportunities},
	volume = {48},
	year = {2023}}

@article{bhimani2014digitisation,
	author = {Bhimani, A. and Willcocks, L.},
	journal = {Accounting and business research},
	number = {4},
	pages = {469-490},
	title = {Digitisation,`Big Data'and the transformation of accounting information},
	volume = {44},
	year = {2014}}

@article{Vlahogianni2015ComputationalIA,
	author = {Vlahogianni, E. I.},
	journal = {Engineering and Applied Sciences Optimization: Dedicated to the Memory of Professor MG Karlaftis},
	pages = {107-128},
	title = {Computational intelligence and optimization for transportation big data: challenges and opportunities},
    publisher="Springer International Publishing",
    address="Cham",
	year = {2015}}

@inproceedings{yan2021positive,
	author = {Yan, S. and Xiong, Y. and Kundu, K. and Yang, S. and Deng, S. and Wang, M. and Xia, W. and Soatto, S.},
	booktitle = {IEEE/CVF Conference on Computer Vision and Pattern Recognition},
	title = {Positive-congruent training: Towards regression-free model updates},
	year = {2021}}

@article{Testi2022MLOpsAT,
	author = {Testi, M. and Ballabio, M. and Frontoni, E. and Iannello, G. and Moccia, S. and Soda, P. and Vessio, G.},
	journal = {IEEE Access},
	pages = {63606-63618},
	title = {MLOps: A Taxonomy and a Methodology},
	volume = {10},
	year = {2022}}

@article{Das2023WhatDM,
	author = {Das, S. D. and Bala, P. K.},
	journal = {Journal of Decision Systems},
	number = {3},
	pages = {376-412},
	title = {What drives MLOps adoption? An analysis using the TOE framework},
	volume = {33},
	year = {2024}}

@article{chowdhury2020trustworthiness,
	author = {Chowdhury, A. and Karmakar, G. and Kamruzzaman, J. and Islam, S.},
	journal = {IEEE Transactions on Industrial Informatics},
	number = {2},
	pages = {961-970},
	title = {Trustworthiness of self-driving vehicles for intelligent transportation systems in industry applications},
	volume = {17},
	year = {2020}}

@article{Hadavi2019MonitoringUT,
	author = {Hadavi, S. and Verlinde, S. and Verbeke, W. and Macharis, C. and Guns, T.},
	journal = {IEEE Transactions on Intelligent Transportation Systems},
	pages = {3747-3758},
	title = {Monitoring Urban-Freight Transport Based on GPS Trajectories of Heavy-Goods Vehicles},
	volume = {20},
	year = {2019}}

@article{tao2019trustdata,
	author = {Tao, H. and Bhuiyan, M. Z. A. and Rahman, M. A. and Wang, T. and Wu, J. and Salih, S. Q. and Li, Y. and Hayajneh, T.},
	journal = {IEEE Transactions on Industrial Informatics},
	number = {5},
	pages = {3311-3321},
	title = {TrustData: Trustworthy and secured data collection for event detection in industrial cyber-physical system},
	volume = {16},
	year = {2019}}

@article{Laa2018RoadTF,
	author = {La\~{n}a, I. and Del Ser, J. and V\'{e}lez, M. and Vlahogianni, E. I.},
	journal = {IEEE Intelligent Transportation Systems Magazine},
	pages = {93-109},
	title = {Road Traffic Forecasting: Recent Advances and New Challenges},
	volume = {10},
	year = {2018}}

@article{Yin2021DeepLO,
	author = {Yin, X. and Wu, G. and Wei, J. and Shen, Y. and Qi, H. and Yin, B.},
	journal = {IEEE Transactions on Intelligent Transportation Systems},
	pages = {4927-4943},
	title = {Deep Learning on Traffic Prediction: Methods, Analysis, and Future Directions},
	volume = {23},
	year = {2021}}

@inproceedings{harby2022data,
	author = {Harby, A. A. and Zulkernine, F.},
	booktitle = {IEEE International Conference on Big Data},
	title = {From Data Warehouse to Lakehouse: A Comparative Review},
	year = {2022}}

@inproceedings{armbrust2021lakehouse,
	author = {Armbrust, M. and Ghodsi, A. and Xin, R. and Zaharia, M.},
	booktitle = {Innovative Data Systems Research},
	title = {Lakehouse: a new generation of open platforms that unify data warehousing and advanced analytics},
    pages={1-8},
	year = {2021}}

@article{caird2004vehicle,
	author = {Caird, J.},
	journal = {Transportation in an Aging Society},
	pages = {1-15},
	title = {In-vehicle intelligent transportation systems},
	volume = {236},
    month = {October},
	year = {2004}}

@article{barenji2019intelligent,
	author = {Barenji, A. V. and Wang, W. M. and Li, Z. and Guerra-Zubiaga, D. A.},
	journal = {Transportation Research Part E: Logistics and Transportation Review},
	pages = {15-31},
	title = {Intelligent E-commerce logistics platform using hybrid agent based approach},
	volume = {126},
	year = {2019}}

@inproceedings{dessalk2020scalable,
	author = {Dessalk, Y. D. and Nikolov, N. and Matskin, M. and Soylu, A. and Roman, D.},
	booktitle = {International Conference on Management of Digital EcoSystems},
	title = {Scalable execution of big data workflows using software containers},
	year = {2020}}

@article{sahal2021blockchain,
	author = {Sahal, R. and Alsamhi, S. H. and Brown, K. N. and O'shea, D. and McCarthy, C. and Guizani, M.},
	journal = {Machines},
	number = {9},
	pages = {193},
	title = {Blockchain-empowered digital twins collaboration: Smart transportation use case},
	volume = {9},
	year = {2021}}

@article{iyer2021ai,
	author = {Iyer, L. S.},
	journal = {Transportation Engineering},
	pages = {100083},
	title = {AI enabled applications towards intelligent transportation},
	volume = {5},
	year = {2021}}

@article{fui2023generative,
	author = {Fui-Hoon Nah, F. and Zheng, R. and Cai, J. and Siau, K. and Chen, L.},
	journal = {Journal of Information Technology Case and Application Research},
	number = {3},
	pages = {277-304},
	title = {Generative AI and ChatGPT: Applications, challenges, and AI-human collaboration},
	volume = {25},
	year = {2023}}

@article{Ahmad2023DeepLM,
	author = {Ahmad, S. and Shakeel, I. and Mehfuz, S. and Ahmad, J.},
	journal = {Computer Science Review},
	pages = {100568},
	title = {Deep learning models for cloud, edge, fog, and IoT computing paradigms: Survey, recent advances, and future directions},
	volume = {49},
	year = {2023}}

@article{chen2012research,
	author = {Chen, X. Y. and Jin, Z. G.},
	journal = {Physics Procedia},
	pages = {561-566},
	title = {Research on key technology and applications for internet of things},
	volume = {33},
	year = {2012},
    month = {June}
    }

@article{Sathyanarayana2023DeepLO,
	author = {Sathyanarayana, M. and Akshay Reddy, K.},
	journal = {International Journal of Advanced Research in Science, Communication and Technology},
	number = {6},
	pages = {4927-4943},
	title = {Deep Learning on Traffic Prediction Methods Analysis and Future Directions},
	volume = {23},
	year = {2021}}

@article{Zhou2022ACS,
	author = {Zhou, Z. and Yang, Z.-J. and Zhang, Y. and Huang, Y. and Chen, H. and Yu, Z.},
	journal = {iScience},
	pages = {1-29},
	title = {A comprehensive study of speed prediction in transportation system: From vehicle to traffic},
	volume = {25},
	year = {2022}}

@inproceedings{Munappy2020FromAD,
	author = {Munappy, A. R. and Mattos, D. I. and Bosch, J. and Holmstr{\"o}m Olsson, H. and Dakkak, A.},
	booktitle = {IEEE/ACM International Conference on Software and System Processes},
	title = {From Ad-Hoc Data Analytics to DataOps},
	year = {2020}}

@book{Atwal2019PracticalDD,
	author = {Atwal, H.},
	publisher = {Springer},
	title = {Practical DataOps: Delivering Agile Data Science at Scale},
	year = {2019}}

@article{Rodrguez2020GoodPF,
	author = {Rodr{\'\i}guez, M. and Pires de Ara{\'u}jo, L. J. and Mazzara, M.},
	journal = {Journal of Physics: Conference Series},
	number = {1},
	pages = {012032},
	title = {Good practices for the adoption of DataOps in the software industry},
	volume = {1694},
	year = {2020}}

@article{Larman2003IterativeAI,
	author = {Larman, C. and Basili, V. R.},
	journal = {Computer},
	pages = {47-56},
	title = {Iterative and incremental developments. a brief history},
	volume = {36},
	year = {2003}}

@inproceedings{Barrachina2012CAOVAAC,
	author = {Barrachina, J. and Garrido, P. and Fogu{\'e}, M. and Martinez, F. J. and Cano, J.-C. and Tavares Calafate, C. M. and Manzoni, P.},
	booktitle = {IEEE Wireless Communications and Networking Conference},
	title = {CAOVA: A Car Accident Ontology for VANETs},
	year = {2012}}

@book{sladkowski2016intelligent,
	author = {S{\l}adkowski, A. and Pamu{\l}a, W.},
	publisher = {Springer},
	title = {Intelligent transportation systems-problems and perspectives},
	year = {2016}}

@article{Liu2019IntelligentRP,
	author = {Liu, Q. and Hou, P. and Wang, G. and Peng, T. and Zhang, S.},
	journal = {Journal of Parallel Distributed Computing},
	pages = {93-106},
	title = {Intelligent route planning on large road networks with efficiency and privacy},
	volume = {133},
	year = {2019}}

@article{hireche2020machine,
	author = {Hireche, S. and Dennai, A.},
	journal = {Smart Energy Empowerment in Smart and Resilient Cities: Renewable Energy for Smart and Sustainable Cities},
	pages = {60-69},
	title = {Machine learning techniques for road traffic automatic incident detection systems: A review},
    publisher="Springer International Publishing", address="Cham",
	year = {2020}}

@article{elallid2022comprehensive,
	author = {Elallid, B. B. and Benamar, N. and Hafid, A. S. and Rachidi, T. and Mrani, N.},
	journal = {Journal of King Saud University-Computer and Information Sciences},
	number = {9},
	pages = {7366-7390},
	title = {A comprehensive survey on the application of deep and reinforcement learning approaches in autonomous driving},
	volume = {34},
	year = {2022}}

@article{haghighat2020applications,
	author = {Haghighat, A. K. and Ravichandra-Mouli, V. and Chakraborty, P. and Esfandiari, Y. and Arabi, S. and Sharma, A.},
	journal = {Journal of Big Data Analytics in Transportation},
	pages = {115-145},
	title = {Applications of deep learning in intelligent transportation systems},
	volume = {2},
	year = {2020},
    month= {August}}

@inproceedings{sen2021devops,
	author = {Sen, A.},
	booktitle = {Evolving Technologies for Computing, Communication and Smart World},
	title = {DevOps, DevSecOps, AIOPS-paradigms to IT operations},
	year = {2021}}

@article{borthakur2007hadoop,
	author = {Borthakur, D.},
	journal = {Hadoop Project Website},
	number = {2007},
	pages = {21},
	title = {The hadoop distributed file system: Architecture and design},
	volume = {11},
	year = {2007}, month={March}}

@article{zaharia2016apache,
	author = {Zaharia, M. and Xin, R. S. and Wendell, P. and Das, T. and Others},
	journal = {Communications of the ACM},
	number = {11},
	pages = {56-65},
	title = {Apache spark: a unified engine for big data processing},
	volume = {59},
	year = {2016}}

@article{carbone2015apache,
	author = {Carbone, P. and Katsifodimos, A. and Ewen, S. and Markl, V. and Haridi, S. and Tzoumas, K.},
	journal = {The Bulletin of the Technical Committee on Data Engineering},
	number = {4},
	pages = {28-38},
	title = {Apache flink: Stream and batch processing in a single engine},
	volume = {38},
	year = {2015}}

@misc{GoogleDataFlow,
	howpublished = {\url{https://cloud.google.com/dataflow?hl=en}},
	note = {[Accessed 15-12-2023]},
	title = {Google Cloud Dataflow}}

@misc{amazonDataPlatform,
	howpublished = {\url{https://aws.amazon.com/emr/}},
	note = {[Accessed 15-12-2023]},
	title = {{A}mazon {E}{M}{R}}}

@misc{microsoftAzureData,
	howpublished = {\url{https://azure.microsoft.com/en-us/products/data-factory}},
	note = {[Accessed 15-12-2023]},
	title = {{A}zure {D}ata {F}actory}}

@misc{powerBI,
	howpublished = {\url{https://www.microsoft.com/en-us/power-platform/products/power-bi}},
	note = {[Accessed 12-11-2024]},
	title = {Microsoft Power BI}}

@misc{tableau,
	howpublished = {\url{https://www.tableau.com/}},
	note = {[Accessed 12-11-2024]},
	title = {Tableau}}

@misc{amazonCloudObject,
	howpublished = {\url{https://aws.amazon.com/s3/}},
	note = {[Accessed 15-12-2023]},
	title = {{A}mazon {S}3}}

@misc{amazonsagemaker,
	howpublished = {\url{https://aws.amazon.com/it/sagemaker/}},
	note = {[Accessed 23-10-2024]},
	title = {AWS SageMaker}}

@misc{amazonDataStream,
	howpublished = {\url{https://aws.amazon.com/kinesis/}},
	note = {[Accessed 15-12-2024]},
	title = {{A}mazon {K}inesis}}

@misc{apacheApacheSparkx2122,
	howpublished = {\url{https://spark.apache.org/}},
	note = {[Accessed 15-12-2023]},
	title = {{A}pache {S}park}}

@misc{apacheApacheHadoop,
	howpublished = {\url{https://hadoop.apache.org/}},
	note = {[Accessed 15-12-2023]},
	title = {{A}pache {H}adoop}}

@misc{apacheApacheKafka,
	howpublished = {\url{https://kafka.apache.org/}},
	note = {[Accessed 15-12-2023]},
	title = {{A}pache {K}afka}}

@misc{apacheApacheFlink,
	howpublished = {\url{https://flink.apache.org/}},
	note = {[Accessed 15-12-2023]},
	title = {{A}pache {F}link{\textregistered}}}

@article{gillani2023machine,
	author = {Gillani, M. and Niaz, H. A.},
	journal = {Complex \& Intelligent Systems},
	number = {2},
	pages = {1879-1897},
	title = {Machine learning based data collection protocol for intelligent transport systems: a real-time implementation on Dublin M50, Ireland},
	volume = {10},
	year = {2024}}

@inproceedings{d2017real,
	author = {D'Silva, G. M. and Khan, A. and Bari, S. and Others},
	booktitle = {IEEE International conference on recent trends in electronics, information \& communication technology},
	title = {Real-time processing of IoT events with historic data using Apache Kafka and Apache Spark with dashing framework},
	year = {2017}}

@article{zhu2018big,
	author = {Zhu, L. and Yu, F. R. and Wang, Y. and Ning, B. and Tang, T.},
	journal = {IEEE Transactions on Intelligent Transportation Systems},
	number = {1},
	pages = {383-398},
	title = {Big data analytics in intelligent transportation systems: A survey},
	volume = {20},
	year = {2018}}

@inproceedings{amini2017big,
	author = {Amini, S. and Gerostathopoulos, I. and Prehofer, C.},
	booktitle = {IEEE international conference on models and technologies for intelligent transportation systems},
	title = {Big data analytics architecture for real-time traffic control},
	year = {2017}}

@article{alic2019bigsea,
	author = {Alic, A. S. and Almeida, J. and Aloisio, G. and Andrade, N. and Antunes, N. and Others},
	journal = {Future generation computer systems},
	pages = {243-269},
	title = {BIGSEA: A Big Data analytics platform for public transportation information},
	volume = {96},
	year = {2019}}

@inproceedings{Khalajzadeh2018DataOps,
	author = {Khalajzadeh, H. and Abdelrazek, M. and Grundy, J. and Hosking, J. and He, Q.},
	booktitle = {IEEE International Congress on Big Data},
	title = {A Survey of Current End-User Data Analytics Tool Support},
	year = {2018}}

@article{zeller2024toward,
	author = {Zeller, M. and Waschulzik, T. and Schmid, R. and Bahlmann, C.},
	journal = {AI and Ethics},
	number = {1},
	pages = {123-130},
	title = {Toward a safe MLOps process for the continuous development and safety assurance of ML-based systems in the railway domain},
	volume = {4},
	year = {2024}}

@inproceedings{makinen2021needs,
	author = {M{\"a}kinen, S. and Skogstr{\"o}m, H. and Laaksonen, E. and Mikkonen, T.},
	booktitle = {IEEE/ACM Workshop on AI Engineering-Software Engineering for AI},
	title = {Who needs MLOps: What data scientists seek to accomplish and how can MLOps help?},
	year = {2021}}

@book{treveil2020introducing,
	author = {Treveil, M. and Omont, N. and Stenac, C. and Lefevre, K. and Phan, D. and Zentici, J. and Lavoillotte, A. and Miyazaki, M. and Heidmann, L.},
	publisher = {O'Reilly Media},
    address={Sebastopol, Calif},
	title = {Introducing MLOps},
	year = {2020}}

@inproceedings{tamburri2020sustainable,
	author = {Tamburri, D. A.},
	booktitle = {International Symposium on Symbolic and Numeric Algorithms for Scientific Computing},
	title = {Sustainable mlops: Trends and challenges},
	year = {2020}}

@inproceedings{alla2021mlops,
	author = {Alla, S. and Adari, S. K.},
	booktitle = {Beginning MLOps with MLFlow: Deploy Models in AWS SageMaker, Google Cloud, and Microsoft Azure},
	title = {What is mlops?},publisher="Springer International Publishing", address="Cham",
	year = {2021}}

@article{sha2023languagempc,
	author = {Sha, H. and Mu, Y. and Jiang, Y. and Chen, L. and Xu, C. and Luo, P. and Li, S. E. and Tomizuka, M. and Zhan, W. and Ding, M.},
	journal = {arXiv preprint arXiv:2310.03026},
	title = {Languagempc: Large language models as decision makers for autonomous driving},
	year = {2023}}

@inproceedings{cui2024survey,
	author = {Cui, C. and Ma, Y. and Cao, X. and Ye, W. and Zhou, Y. and Liang, K. and Chen, J. and Lu, J. and Yang, Z. and Liao, K.-D. and others},
	booktitle = {IEEE/CVF Winter Conference on Applications of Computer Vision},
	title = {A survey on multimodal large language models for autonomous driving},
	year = {2024},
    month = {June}    }

@inproceedings{spring2017scalable,
	author = {Spring, R. and Shrivastava, A.},
	booktitle = {ACM SIGKDD International Conference on Knowledge Discovery and Data Mining},
	title = {Scalable and sustainable deep learning via randomized hashing},
	year = {2017}}

@inproceedings{sperrle2021survey,
	author = {Sperrle, F. and El-Assady, M. and Guo, G. and Borgo, R. and Chau, D. H. and Endert, A. and Keim, D.},
	booktitle = {Computer Graphics Forum},
	title = {A Survey of Human-Centered Evaluations in Human-Centered Machine Learning},
	year = {2021}}

@article{cina2023wild,
	author = {Cin{\`a}, A. E. and Grosse, K. and Demontis, A. and Vascon, S. and Zellinger, W. and Moser, B. A. and Oprea, A. and Biggio, B. and Pelillo, M. and Roli, F.},
	journal = {ACM Computing Surveys},
	number = {13s},
	pages = {1-39},
	title = {Wild patterns reloaded: A survey of machine learning security against training data poisoning},
	volume = {55},
	year = {2023}}

@article{biggio18wild,
	author = {Biggio, B. and Roli, F.},
	journal = {Pattern Recognition},
	pages = {317-331},
	title = {Wild patterns: Ten years after the rise of adversarial machine learning},
	volume = {84},
	year = {2018}}

@article{liu2021machine,
	author = {Liu, B. and Ding, M. and Shaham, S. and Rahayu, W. and Farokhi, F. and Lin, Z.},
	journal = {ACM Computing Surveys},
	number = {2},
	pages = {1-36},
	title = {When machine learning meets privacy: A survey and outlook},
	volume = {54},
	year = {2021}}

@article{Oneto22TrustGraph,
	author = {Oneto, L. and Navarin, N. and Biggio, B. and Errica, F. and Others},
	journal = {Neurocomputing},
	pages = {217-243},
	title = {Towards learning trustworthily, automatically, and with guarantees on graphs: An overview},
	volume = {493},
	year = {2022}}

@inproceedings{guerreiro2016architecture,
	author = {Guerreiro, G. and Figueiras, P. and Silva, R. and Costa, R. and Jardim-Goncalves, R.},
	booktitle = {IEEE International Conference on Intelligent Systems},
	title = {An architecture for big data processing on intelligent transportation systems. An application scenario on highway traffic flows},
    publisher="Springer International Publishing", address="Cham",
	year = {2016}}

@inproceedings{leung2019urban,
	author = {Leung, C. K. and Braun, P. and Hoi, C. S. H. and Souza, J. and Cuzzocrea, A.},
	booktitle = {Big Data Analytics and Knowledge Discovery},
	title = {Urban analytics of big transportation data for supporting smart cities},
      pages        = {24--33},
	year = {2019}}

@article{ji2017low,
	author = {Ji, S. and Sun, Q.},
	journal = {Sustainability},
	number = {11},
	pages = {2052},
	title = {Low-carbon planning and design in B\&R logistics service: a case study of an e-commerce big data platform in China},
	volume = {9},
	year = {2017}}

@inproceedings{raj2021edge,
	author = {Raj, E. and Buffoni, D. and Westerlund, M. and Ahola, K.},
	booktitle = {IEEE International Conference on Cloud Engineering},
	title = {Edge mlops: An automation framework for aiot applications},
	year = {2021}}

@article{pineda2023mlops,
	author = {Pineda-Jaramillo, J. and Viti, F.},
	journal = {Engineering Applications of Artificial Intelligence},
	pages = {106222},
	title = {MLOps in freight rail operations},
	volume = {123},
	year = {2023}}

@article{faubel2023mlops,
	author = {Faubel, L. and Schmid, K. and Eichelberger, H.},
	journal = {SN Computer Science},
	number = {6},
	pages = {828},
	title = {MLOps Challenges in Industry 4.0},
	volume = {4},
	year = {2023}}

@inproceedings{symeonidis2022mlops,
	author = {Symeonidis, G. and Nerantzis, E. and Kazakis, A. and Papakostas, G. A.},
	booktitle = {IEEE Annual Computing and Communication Workshop and Conference},
	title = {Mlops-definitions, tools and challenges},
	year = {2022}}

@article{koh2020real,
	author = {Koh, S. and Zhou, B. and Fang, H. and Yang, P. and Yang, Z. and Yang, Q. and Guan, L. and Ji, Z.},
	journal = {Applied Soft Computing},
	pages = {106694},
	title = {Real-time deep reinforcement learning based vehicle navigation},
	volume = {96},
	year = {2020}}

@article{jin2021trafficbert,
	author = {Jin, K. and Wi, J. A. and Lee, E. J. and Kang, S. J. and Kim, S. K. and Kim, Y. B.},
	journal = {Expert Systems with Applications},
	pages = {115738},
	title = {TrafficBERT: Pre-trained model with large-scale data for long-range traffic flow forecasting},
	volume = {186},
	year = {2021}}

@inproceedings{zhang2018long,
	author = {Zhang, C. and Patras, P.},
	booktitle = {ACM International Symposium on Mobile Ad Hoc Networking and Computing},
	title = {Long-term mobile traffic forecasting using deep spatio-temporal neural networks},
	year = {2018}}

@inproceedings{banerjee2020challenges,
	author = {Banerjee, A. and Chen, C.-C. and Hung, C.-C. and Huang, X. and Wang, Y. and Chevesaran, R.},
	booktitle = {USENIX Conference on Operational Machine Learning},
	title = {Challenges and Experiences with $\{$MLOps$\}$ for Performance Diagnostics in $\{$Hybrid-Cloud$\}$ Enterprise Software Deployments},
    pages={1-2},
	year = {2020}}

@article{haibe2020transparency,
	author = {Haibe-Kains, B. and Adam, G. A. and Hosny, A. and Khodakarami, F. and Others},
	journal = {Nature},
	number = {7829},
	pages = {E14-E16},
	title = {Transparency and reproducibility in artificial intelligence},
	volume = {586},
	year = {2020}}

@inproceedings{adkins2022method,
	author = {Adkins, D. and Alsallakh, B. and Cheema, A. and Kokhlikyan, N. and Others},
	booktitle = {International Conference on AI Engineering: Software Engineering for AI},
	title = {Method cards for prescriptive machine-learning transparency},
	year = {2022}}

@inproceedings{sculley2015hidden,
	author = {Sculley, D. and Holt, G. and Golovin, D. and Davydov, E. and Phillips, T. and Ebner, D. and Chaudhary, V. and Young, M. and Crespo, J.-F. and Dennison, D.},
	booktitle = {Neural Information Processing Systems},
	title = {Hidden technical debt in machine learning systems},
      pages        = {2503--2511},
	year = {2015}}

@article{gorishniy2021revisiting,
	author = {Gorishniy, Y. and Rubachev, I. and Khrulkov, V. and Babenko, A.},
	journal = {Neural Information Processing Systems},
	pages = {18932-18943},
	title = {Revisiting deep learning models for tabular data},
	volume = {34},
	year = {2021}}

@inproceedings{cao2016deep,
	author = {Cao, S. and Lu, W. and Xu, Q.},
	booktitle = {AAAI conference on artificial intelligence},
	title = {Deep neural networks for learning graph representations},
      pages        = {1145--1152},
	year = {2016}}

@article{li2020survey,
	author = {Li, Q. and Peng, H. and Li, J. and Xia, C. and Yang, R. and Sun, L. and Yu, P. S. and He, L.},
	journal = {arXiv preprint arXiv:2008.00364},
	title = {A survey on text classification: From shallow to deep learning},
	year = {2020}}

@article{kibria2018big,
	author = {Kibria, M. G. and Nguyen, K. and Villardi, G. P. and Zhao, O. and Ishizu, K. and Kojima, F.},
	journal = {IEEE access},
	pages = {32328-32338},
	title = {Big data analytics, machine learning, and artificial intelligence in next-generation wireless networks},
	volume = {6},
	year = {2018}}

@inproceedings{roy2022complete,
	author = {Roy, D. and Srivastava, R. and Jat, M. and Karaca, M. S.},
	booktitle = {Decision intelligence analytics and the implementation of strategic business management},
	title = {A complete overview of analytics techniques: descriptive, predictive, and prescriptive},publisher="Springer International Publishing", address="Cham",
	year = {2022}}

@inproceedings{silva2020benchmarking,
	author = {Silva, L. C. and Zagatti, F. R. and Sette, B. S. and dos Santos Silva, L. N. and Lucr\'edio, D. and Silva, D. F. and de Medeiros Caseli, H.},
	booktitle = {IEEE International Conference on Machine Learning and Applications},
	title = {Benchmarking machine learning solutions in production},
	year = {2020}}

@article{chicco2022eleven,
	author = {Chicco, D. and Oneto, L. and Tavazzi, E.},
	journal = {PLOS Computational Biology},
	number = {12},
	pages = {e1010718},
	title = {Eleven quick tips for data cleaning and feature engineering},
	volume = {18},
	year = {2022}}

@article{white2000reality,
	author = {White, H.},
	journal = {Econometrica},
	number = {5},
	pages = {1097-1126},
	title = {A reality check for data snooping},
	volume = {68},
	year = {2000}}

@article{kaur2022trustworthy,
	author = {Kaur, D. and Uslu, S. and Rittichier, K. J. and Durresi, A.},
	journal = {ACM computing surveys},
	number = {2},
	pages = {1-38},
	title = {Trustworthy artificial intelligence: a review},
	volume = {55},
	year = {2022}}

@inproceedings{Sun2022SHIFTAS,
	author = {Sun, T. and Segu, M. and Postels, J. and Wang, Y. and Van Gool, L. and Schiele, B. and Tombari, F. and Yu, F.},
	booktitle = {IEEE/CVF Conference on Computer Vision and Pattern Recognition},
	title = {SHIFT: A Synthetic Driving Dataset for Continuous Multi-Task Domain Adaptation},
	year = {2022}}

@article{cina2022energy,
	author = {Cin{\`a}, Antonio Emanuele and Demontis, Ambra and Biggio, Battista and Roli, Fabio and Pelillo, Marcello},
	journal = {Information Sciences},
	pages = {121905},
	title = {Energy-latency attacks via sponge poisoning},
	volume = {702},
	year = {2025}}

@inproceedings{ribeiro2015mlaas,
	author = {Ribeiro, M. and Grolinger, K. and Capretz, M. A. M.},
	booktitle = {IEEE international conference on machine learning and applications},
	title = {Mlaas: Machine learning as a service},
	year = {2015}}

@inproceedings{yao2017complexity,
	author = {Yao, Y. and Xiao, Z. and Wang, B. and Viswanath, B. and Zheng, H. and Zhao, B. Y.},
	booktitle = {Internet Measurement Conference},
	title = {Complexity vs. performance: empirical analysis of machine learning as a service},
	year = {2017}}

@article{Sze17SurveyAsic,
	author = {Sze, V. and Chen, Y.-H. and Yang, T.-J. and Emer, J. S.},
	journal = {Proceedings of the IEEE},
	number = {12},
	pages = {2295-2329},
	title = {Efficient Processing of Deep Neural Networks: {A} Tutorial and Survey},
	volume = {105},
	year = {2017}}

@inproceedings{Jouppi17GoogleAsicTPU,
	author = {Jouppi, N. P. et al.},
	booktitle = {International Symposium on Computer Architecture},
	title = {In-Datacenter Performance Analysis of a Tensor Processing Unit},
	year = {2017}}

@inproceedings{Shumailov21Sponge,
	author = {Shumailov, I. and Zhao, Y. and Bates, D. and Papernot, N. and Mullins, R. D. and Anderson, R.},
	booktitle = {IEEE European Symposium on Security and Privacy},
	title = {Sponge Examples: Energy-Latency Attacks on Neural Networks},
	year = {2021}}

@inproceedings{marro2023computational,
	author = {Marro, S. and Lombardi, M.},
	booktitle = {International Conference on Machine Learning},
	title = {Computational asymmetries in robust classification},
      pages        = {24082--24138},
	year = {2023}}

@article{supritha2018privacy,
	author = {Supritha, S. and Sushmitha, S. and Asghar, D. and Mohan, K. and Basavaraju, S.},
	journal = {Perspectives in Communication, Embedded-systems and Signal-processing-PiCES},
	number = {9},
	pages = {218-222},
	title = {Privacy-Preserving Data Mining: Methods, Metrics and Applications},
	volume = {2},
	year = {2018}}

@inproceedings{shokri2017membership,
	author = {Shokri, R. and Stronati, M. and Song, C. and Shmatikov, V.},
	booktitle = {IEEE symposium on security and privacy},
	title = {Membership inference attacks against machine learning models},
	year = {2017}}

@inproceedings{orekondy2019knockoff,
	author = {Orekondy, T. and Schiele, B. and Fritz, M.},
	booktitle = {IEEE/CVF conference on computer vision and pattern recognition},
	title = {Knockoff nets: Stealing functionality of black-box models},
	year = {2019}}

@inproceedings{agrawal2000privacy,
	author = {Agrawal, R. and Srikant, R.},
	booktitle = {ACM SIGMOD international conference on Management of data},
	title = {Privacy-preserving data mining},
	year = {2000}}

@inproceedings{agrawal2001design,
	author = {Agrawal, D. and Aggarwal, C. C.},
	booktitle = {ACM SIGMOD-SIGACT-SIGART symposium on Principles of database systems},
	title = {On the design and quantification of privacy preserving data mining algorithms},
	year = {2001}}

@inproceedings{cina2024attackbench,
	author = {Cin{\`a}, A. E. and Rony, J. and Pintor, M. and Demetrio, L. and Demontis, A. and Biggio, B. and Ayed, I. B. and Roli, F.},
	booktitle = {AAAI Conference on Artificial Intelligence},
    pages={2600–2608},
	title = {AttackBench: Evaluating Gradient-based Attacks for Adversarial Examples},
	year = {2025}}

@inproceedings{kantchelian2016evasion,
	author = {Kantchelian, A. and Tygar, J. D. and Joseph, A.},
	booktitle = {International conference on machine learning},
	title = {Evasion and hardening of tree ensemble classifiers},
    pages   = {2387--2396},
	year = {2016}}

@inproceedings{wang2020lp,
	author = {Wang, Y. and Zhang, H. and Chen, H. and Boning, D. and Hsieh, C.-J.},
	booktitle = {International Conference on Machine Learning},
	title = {On lp-norm robustness of ensemble decision stumps and trees},
     pages        = {10104--10114},
	year = {2020}}

@inproceedings{calzavara2023verifiable,
	author = {Calzavara, S. and Cazzaro, L. and Pibiri, G. E. and Prezza, N.},
	booktitle = {ACM SIGSAC Conference on Computer and Communications Security},
	title = {Verifiable learning for robust tree ensembles},
	year = {2023}}

@article{guidotti2020verification,
	author = {Guidotti, D. and Leofante, F. and Pulina, L. and Tacchella, A.},
	journal = {arXiv preprint arXiv:2003.07636},
	title = {Verification of neural networks: Enhancing scalability through pruning},
	year = {2020}}

@inproceedings{jia2020efficient,
	author = {Jia, K. and Rinard, M.},
	booktitle = {Neural Information Processing Systems},
	title = {Efficient exact verification of binarized neural networks},
	year = {2020}}

@inproceedings{ranzato2020abstract,
	author = {Ranzato, F. and Zanella, M.},
	booktitle = {AAAI Conference on Artificial Intelligence},
      pages        = {5478--5486},
	title = {Abstract interpretation of decision tree ensemble classifiers},
	year = {2020}}

@inproceedings{fredrikson2015model,
	author = {Fredrikson, M. and Jha, S. and Ristenpart, T.},
	booktitle = {ACM SIGSAC conference on computer and communications security},
	title = {Model inversion attacks that exploit confidence information and basic countermeasures},
	year = {2015}}

@article{al2016reconstruction,
	author = {Al-Rubaie, M. and Chang, J. M.},
	journal = {IEEE Transactions on Information Forensics and Security},
	number = {12},
	pages = {2648-2663},
	title = {Reconstruction attacks against mobile-based continuous authentication systems in the cloud},
	volume = {11},
	year = {2016}}

@article{al2019privacy,
	author = {Al-Rubaie, M. and Chang, J. M.},
	journal = {IEEE Security \& Privacy},
	number = {2},
	pages = {49-58},
	title = {Privacy-preserving machine learning: Threats and solutions},
	volume = {17},
	year = {2019}}

@article{machanavajjhala2007diversity,
	author = {Machanavajjhala, A. and Kifer, D. and Gehrke, J. and Venkitasubramaniam, M.},
	journal = {Acm transactions on knowledge discovery from data (tkdd)},
	number = {1},
	pages = {3-es},
	title = {l-diversity: Privacy beyond k-anonymity},
	volume = {1},
	year = {2007}}

@article{sweeney2002k,
	author = {Sweeney, L.},
	journal = {International journal of uncertainty, fuzziness and knowledge-based systems},
	number = {05},
	pages = {557-570},
	title = {k-anonymity: A model for protecting privacy},
	volume = {10},
	year = {2002}}

@inproceedings{abadi2016deep,
	author = {Abadi, M. and Chu, A. and Goodfellow, I. and McMahan, H. B. and Mironov, I. and Talwar, K. and Zhang, L.},
	booktitle = {ACM SIGSAC conference on computer and communications security},
    pages = {308–318}, 
	title = {Deep learning with differential privacy},
	year = {2016}}

@article{arachchige2019local,
	author = {Arachchige, P. C. M. and Bertok, P. and Khalil, I. and Liu, D. and Camtepe, S. and Atiquzzaman, M.},
	journal = {IEEE Internet of Things Journal},
	number = {7},
	pages = {5827-5842},
	title = {Local differential privacy for deep learning},
	volume = {7},
	year = {2019}}

@article{fang2021privacy,
	author = {Fang, H. and Qian, Q.},
	journal = {Future Internet},
	number = {4},
	pages = {94},
	title = {Privacy preserving machine learning with homomorphic encryption and federated learning},
	volume = {13},
	year = {2021}}

@article{aono2017privacy,
	author = {Aono, Y. and Hayashi, T. and Wang, L. and Moriai, S. and others},
	journal = {IEEE transactions on information forensics and security},
	number = {5},
	pages = {1333-1345},
	title = {Privacy-preserving deep learning via additively homomorphic encryption},
	volume = {13},
	year = {2017}}

@article{sattler2020clustered,
	author = {Sattler, F. and M{\"u}ller, K.-R. and Samek, W.},
	journal = {IEEE transactions on neural networks and learning systems},
	number = {8},
	pages = {3710-3722},
	title = {Clustered federated learning: Model-agnostic distributed multitask optimization under privacy constraints},
	volume = {32},
	year = {2020}}

@article{hu2020personalized,
	author = {Hu, R. and Guo, Y. and Li, H. and Pei, Q. and Gong, Y.},
	journal = {IEEE Internet of Things Journal},
	number = {10},
	pages = {9530-9539},
	title = {Personalized federated learning with differential privacy},
	volume = {7},
	year = {2020}}

@article{lyu2022privacy,
	author = {Lyu, L. and Yu, H. and Ma, X. and Chen, C. and Sun, L. and Zhao, J. and Yang, Q. and Philip, S. Y.},
	journal = {IEEE transactions on neural networks and learning systems},
	number = {7},
	pages = {8726 - 8746},
	title = {Privacy and robustness in federated learning: Attacks and defenses},
	volume = {35},
	year = {2024}}

@article{sucasas2016autonomous,
	author = {Sucasas, V. and Mantas, G. and Saghezchi, F. B. and Radwan, A. and Rodriguez, J.},
	journal = {Computers \& Security},
	pages = {193-205},
	title = {An autonomous privacy-preserving authentication scheme for intelligent transportation systems},
	volume = {60},
	year = {2016}}

@article{jiang2022trust,
	author = {Jiang, B. and Huang, G. and Wang, T. and Gui, J. and Zhu, X.},
	journal = {Transactions on Emerging Telecommunications Technologies},
	number = {6},
	pages = {e3942},
	title = {Trust based energy efficient data collection with unmanned aerial vehicle in edge network},
	volume = {33},
	year = {2022}}

@article{fletcher2015measuring,
	author = {Fletcher, S. and Islam, M. Z.},
	journal = {International Journal of Computer Theory and Engineering},
	number = {1},
	pages = {21-28},
	title = {Measuring information quality for privacy preserving data mining},
	volume = {7},
	year = {2015}}

@inproceedings{alvim2012differential,
	author = {Alvim, M. S. and Andr{\'e}s, M. E. and Chatzikokolakis, K. and Degano, P. and Palamidessi, C.},
	booktitle = {Formal Aspects of Security and Trust: 8th International Workshop},
	title = {Differential privacy: on the trade-off between utility and information leakage},
      pages        = {39--54},
	year = {2012}}

@inproceedings{mohassel2017secureml,
	author = {Mohassel, P. and Zhang, Y.},
	booktitle = {IEEE symposium on security and privacy},
	title = {Secureml: A system for scalable privacy-preserving machine learning},
	year = {2017}}

@inproceedings{tan2021cryptgpu,
	author = {Tan, S. and Knott, B. and Tian, Y. and Wu, D. J.},
	booktitle = {IEEE Symposium on Security and Privacy},
	title = {CryptGPU: Fast privacy-preserving machine learning on the GPU},
	year = {2021}}

@article{ali2023explainable,
	author = {Ali, S. and Abuhmed, T. and El-Sappagh, S. and Muhammad, K. and Others},
	journal = {Information Fusion},
	pages = {101805},
	title = {Explainable {Artificial Intelligence (XAI)}: {W}hat We Know and What Is Left To Attain Trustworthy {A}rtificial {I}ntelligence},
	volume = {99},
	year = {2023}}

@article{carvalho2019machine,
	author = {Carvalho, D. V. and Pereira, E. M. and Cardoso, J. S.},
	journal = {Electronics},
	number = {8},
	pages = {832},
	title = {Machine learning interpretability: A survey on methods and metrics},
	volume = {8},
	year = {2019}}

@article{arrieta2020explainable,
	author = {Arrieta, A. B. and D{\'\i}az-Rodr{\'\i}guez, N. and Del Ser, J. and Bennetot, A. and Others},
	journal = {Information fusion},
	pages = {82-115},
	title = {Explainable Artificial Intelligence (XAI): Concepts, taxonomies, opportunities and challenges toward responsible AI},
	volume = {58},
	year = {2020}}

@article{Paez19pragmatic,
	author = {P{\'a}ez, A.},
	issue_date = {Sep 2019},
	journal = {Minds and Machines},
	pages = {441-459},
	title = {The Pragmatic Turn in Explainable {Artificial} {Intelligence} ({XAI})},
	volume = {29},
	year = {2019}}

@article{lipton2018mythos,
	author = {Lipton, Z. C.},
	journal = {Queue},
	pages = {31-57},
	title = {The Mythos of Model Interpretability: {I}n {Machine Learning}, the Concept of Interpretability is Both Important and Slippery},
	volume = {16},
	year = {2018}}

@inproceedings{yao2021explanatory,
	author = {Yao, Y.},
	booktitle = {Machine Learning and Knowledge Extraction},
	title = {Explanatory Pluralism in Explainable {AI}},
    pages="275--292",
	year = {2021}}

@article{dwivedi2023explainable,
	author = {Dwivedi, R. and Dave, D. and Naik, H. and Singhal, S. and Others},
	journal = {ACM Computing Surveys},
	number = {9},
	pages = {1-33},
	title = {Explainable {AI} ({XAI}): {C}ore Ideas, Techniques, and Solutions},
	volume = {55},
	year = {2023}}

@article{Guidotti2019Survey,
	author = {Guidotti, R. and Monreale, A. and Ruggieri, S. and Turini, F. and Giannotti, F. and Pedreschi, D.},
	journal = {ACM Computer Survey},
	number = {5},
	numpages = {42},
	title = {A Survey of Methods for Explaining Black Box Models},
	volume = {51},
	year = {2018}}

@article{tibshirani1996regression,
	author = {Tibshirani, R.},
	journal = {Journal of the Royal Statistical Society Series B: Statistical Methodology},
	number = {1},
	pages = {267-288},
	title = {Regression shrinkage and selection via the lasso},
	volume = {58},
	year = {1996}}

@article{kuo2019interpretable,
	author = {Kuo, C.-C. J. and Zhang, M. and Li, S. and Duan, J. and Chen, Y.},
	journal = {Journal of Visual Communication and Image Representation},
	pages = {346-359},
	title = {Interpretable convolutional neural networks via feedforward design},
	volume = {60},
	year = {2019}}

@inproceedings{kawaguchi2016deep,
	author = {Kawaguchi, K.},
	booktitle = {Neural Information Processing Systems},
	title = {Deep learning without poor local minima},
	year = {2016}}

@inproceedings{ribeiro2016should,
	author = {Ribeiro, M. T. and Singh, S. and Guestrin, C.},
	booktitle = {ACM SIGKDD international conference on knowledge discovery and data mining},
      pages        = {1135--1144},
	title = {" Why should i trust you?" Explaining the predictions of any classifier},
	year = {2016}}

@inproceedings{lundberg2017unified,
	author = {Lundberg, S. M. and Lee, S.-I.},
	booktitle = {Neural Information Processing Systems},
      pages        = {4765--4774},
	title = {A unified approach to interpreting model predictions},
	year = {2017}}

@book{gaur2022explainable,
	author = {Gaur, L. and Sahoo, B. M.},
	publisher = {Springer},
	title = {Explainable Artificial Intelligence for Intelligent Transportation Systems: Ethics and Applications},
	year = {2022}}

@article{castelnovo2022clarification,
	author = {Castelnovo, A. and Crupi, R. and Greco, G. and Regoli, D. and Penco, I. G. and Cosentini, A. C.},
	journal = {Scientific Reports},
	number = {1},
	pages = {4209},
	title = {A clarification of the nuances in the fairness metrics landscape},
	volume = {12},
	year = {2022}}

@book{barocas2023fairness,
	author = {Barocas, S. and Hardt, M. and Narayanan, A.},
	publisher = {MIT press},
	title = {Fairness and machine learning: Limitations and opportunities},
	year = {2023}}

@inproceedings{kusner2017counterfactual,
	author = {Kusner, M. J. and Loftus, J. and Russell, C. and Silva, R.},
	booktitle = {Neural Information Processing Systems},
	title = {Counterfactual fairness},
	year = {2017}}

@inproceedings{chiappa2019path,
	author = {Chiappa, S.},
	booktitle = {AAAI conference on artificial intelligence},
    pages        = {7801--7808},
	title = {Path-specific counterfactual fairness},
	year = {2019}}

@inproceedings{dwork2012fairness,
	author = {Dwork, C. and Hardt, M. and Pitassi, T. and Reingold, O. and Zemel, R.},
	booktitle = {Innovations in theoretical computer science conference},
	title = {Fairness through awareness},
	year = {2012}}

@article{liu2022distributed,
	author = {Liu, J. and Huang, J. and Zhou, Y. and Li, X. and Ji, S. and Xiong, H. and Dou, D.},
	journal = {Knowledge and Information Systems},
	number = {4},
	pages = {885-917},
	title = {From distributed machine learning to federated learning: A survey},
	volume = {64},
	year = {2022}}

@article{prasser2020flexible,
	author = {Prasser, F. and Eicher, J. and Spengler, H. and Bild, R. and Kuhn, K. A.},
	journal = {Software: Practice and Experience},
	number = {7},
	pages = {1277-1304},
	title = {Flexible data anonymization using ARX-Current status and challenges ahead},
	volume = {50},
	year = {2020}}

@article{hosain2024explainable,
	author = {Hosain, M. T. and Jim, J. R. and Mridha, M. F. and Kabir, M. M.},
	journal = {Computers and Electrical Engineering},
	pages = {109246},
	title = {Explainable AI approaches in deep learning: Advancements, applications and challenges},
	volume = {117},
	year = {2024}}

@article{soares2020explaining,
	author = {Soares, E. and Angelov, P. P. and Costa, B. and Castro, M. P. G. and Nageshrao, S. and Filev, D.},
	journal = {IEEE Transactions on Fuzzy Systems},
	number = {8},
	pages = {2399-2407},
	title = {Explaining deep learning models through rule-based approximation and visualization},
	volume = {29},
	year = {2020}}

@inproceedings{robnik2018perturbation,
	author = {Robnik-{\v S}ikonja, M. and Bohanec, M.},
	booktitle = {Human and Machine Learning: Visible, Explainable, Trustworthy and Transparent},
	title = {Perturbation-based explanations of prediction models},publisher="Springer International Publishing", address="Cham",
	year = {2018}}

@inproceedings{islam2022through,
	author = {Islam, M. T. and Fariha, A. and Meliou, A. and Salimi, B.},
	booktitle = {International Conference on Management of Data},
	title = {Through the data management lens: Experimental analysis and evaluation of fair classification},
	year = {2022}}

@inproceedings{elzayn2019fair,
	author = {Elzayn, H. and Jabbari, S. and Jung, C. and Kearns, M. and Neel, S. and Roth, A. and Schutzman, Z.},
	booktitle = {Conference on Fairness, Accountability, and Transparency},
	title = {Fair algorithms for learning in allocation problems},
	year = {2019}}

@inproceedings{anastasiou2019admsv2,
	author = {Anastasiou, C. and Lin, J. and He, C. and Chiang, Y. Y. and Shahabi, C.},
	booktitle = {ACM SIGSPATIAL International Workshop on Advances on Resilient and Intelligent Cities},
	title = {Admsv2: A modern architecture for transportation data management and analysis},
	year = {2019}}

@article{babar2019real,
	author = {Babar, M. and Arif, F.},
	journal = {Journal of Ambient Intelligence and Humanized Computing},
	pages = {4167-4177},
	title = {Real-time data processing scheme using big data analytics in internet of things based smart transportation environment},
	volume = {10},
	year = {2019}}

@inproceedings{alomari2019road,
	author = {Alomari, E. and Mehmood, R. and Katib, I.},
	booktitle = {IEEE SmartWorld, Ubiquitous Intelligence \& Computing, Advanced \& Trusted Computing, Scalable Computing \& Communications, Internet of People and Smart City Innovation},
	title = {Road traffic event detection using twitter data, machine learning, and apache spark},
      pages={1888-1895},
	year = {2019}}

@inproceedings{sinnott2015smash,
	author = {Sinnott, R. O. and Morandini, L. and Wu, S.},
	booktitle = {IEEE International Conference on Data Science and Data Intensive Systems},
	title = {SMASH: A cloud-based architecture for big data processing and visualization of traffic data},
	year = {2015}}

@inproceedings{wang2023technical,
	author = {Wang, X. and Schuster, H. and Borrison, R. and Kl{\"o}pper, B.},
	booktitle = {International Conference on Industrial Informatics},
	title = {Technical Debt Management in Industrial ML-State of Practice and Management Model Proposal},
	year = {2023}}

@inproceedings{haertel2023mlops,
	author = {Haertel, C. and Staegemann, D. and Daase, C. and Pohl, M. and Nahhas, A. and Turowski, K.},
	booktitle = {IEEE International Conference on Big Data (BigData)},
	title = {MLOps in Data Science Projects: A Review},
	year = {2023}}

@article{khalajzadeh2020end,
	author = {Khalajzadeh, H. and Simmons, A. J. and Abdelrazek, M. and Grundy, J. and Hosking, J. and He, Q.},
	journal = {Journal of Computer Languages},
	pages = {100964},
	title = {An end-to-end model-based approach to support big data analytics development},
	volume = {58},
	year = {2020}}

@article{qureshi2013survey,
	author = {Qureshi, K. N. and Abdullah, A. H.},
	journal = {Middle-East Journal of Scientific Research},
	number = {5},
	pages = {629-642},
	title = {A survey on intelligent transportation systems},
	volume = {15},
	year = {2013}}

@inproceedings{alam2016introduction,
	author = {Alam, M. and Ferreira, J. and Fonseca, J.},
	booktitle = {Intelligent transportation systems: Dependable vehicular communications for improved road safety},
	title = {Introduction to intelligent transportation systems},
	year = {2016},    publisher="Springer International Publishing", address="Cham"}

@article{zhang2011data,
	author = {Zhang, J. and Wang, F.-Y. and Wang, K. and Lin, W.-H. and Xu, X. and Chen, C.},
	journal = {IEEE Transactions on Intelligent Transportation Systems},
	number = {4},
	pages = {1624-1639},
	title = {Data-driven intelligent transportation systems: A survey},
	volume = {12},
	year = {2011}}

@article{tsolaki2023utilizing,
	author = {Tsolaki, K. and Vafeiadis, T. and Nizamis, A. and Ioannidis, D. and Tzovaras, D.},
	journal = {ICT Express},
	number = {3},
	pages = {284-295},
	title = {Utilizing machine learning on freight transportation and logistics applications: A review},
	volume = {9},
	year = {2023}}

@article{ouallane2022fusion,
	author = {Ouallane, A. A. and Bakali, A. and Bahnasse, A. and Broumi, S. and Talea, M.},
	journal = {Information Fusion},
	pages = {218-248},
	title = {Fusion of engineering insights and emerging trends: Intelligent urban traffic management system},
	volume = {88},
	year = {2022}}

@article{degas2022survey,
	author = {Degas, A. and Islam, M. R. and Hurter, C. and Barua, S. and Others},
	journal = {Applied Sciences},
	number = {3},
	pages = {1295},
	title = {A survey on artificial intelligence (ai) and explainable ai in air traffic management: Current trends and development with future research trajectory},
	volume = {12},
	year = {2022}}

@article{nti2022environmental,
	author = {Nti, E. K. and Cobbina, S. J. and Attafuah, E. E. and Opoku, E. and Gyan, M. A.},
	journal = {Sustainable Futures},
	pages = {100068},
	title = {Environmental sustainability technologies in biodiversity, energy, transportation and water management using artificial intelligence: A systematic review},
	volume = {4},
	year = {2022},
    month={March}}

@article{tang2022literature,
	author = {Tang, R. and De Donato, L. and Besinovi{\'c}, N. and Flammini, F. and Goverde, R. M. P. and Lin, Z. and Liu, R. and Tang, T. and Vittorini, V. and Wang, Z.},
	journal = {Transportation Research Part C: Emerging Technologies},
	pages = {103679},
	title = {A literature review of Artificial Intelligence applications in railway systems},
	volume = {140},
	year = {2022}}

@inproceedings{zhilenkov2018problems,
	author = {Zhilenkov, A. A. and Epifantsev, I. R.},
	booktitle = {IEEE Conference of Russian Young Researchers in Electrical and Electronic Engineering},
	title = {Problems of a trajectory planning in autonomous navigation systems based on technical vision and AI},
	year = {2018}}

@article{musa2023sustainable,
	author = {Musa, A. A. and Malami, S. I. and Alanazi, F. and Ounaies, W. and Alshammari, M. and Haruna, S. I.},
	journal = {Sustainability},
	number = {13},
	pages = {9859},
	title = {Sustainable Traffic Management for Smart Cities Using Internet-of-Things-Oriented Intelligent Transportation Systems (ITS): Challenges and Recommendations},
	volume = {15},
	year = {2023}}

@article{de2021simulation,
	author = {de la Torre, R. and Corlu, C. G. and Faulin, J. and Onggo, B. S. and Juan, A. A.},
	journal = {Sustainability},
	number = {3},
	pages = {1551},
	title = {Simulation, optimization, and machine learning in sustainable transportation systems: models and applications},
	volume = {13},
	year = {2021}}

@article{fiore2019integrated,
	author = {Fiore, S. and Elia, D. and Pires, C. E. and Mestre, D. G. and Others},
	journal = {IEEE access},
	pages = {117652-117677},
	title = {An integrated big and fast data analytics platform for smart urban transportation management},
	volume = {7},
	year = {2019}}

@article{ushakov2022artificial,
	author = {Ushakov, D. and Dudukalov, E. and Shmatko, L. and Shatila, K.},
	journal = {Transportation Research Procedia},
	pages = {2401-2408},
	title = {Artificial Intelligence as a factor of public transportations system development},
	volume = {63},
	year = {2022}, month={June}}

@inproceedings{martinez2015driving,
	author = {Mart{\'\i}nez, M. V. and Del Campo, I. and Echanobe, J. and Basterretxea, K.},
	booktitle = {IEEE International Conference on Intelligent Transportation Systems},
	title = {Driving behavior signals and machine learning: A personalized driver assistance system},
	year = {2015}}

@article{tong2019artificial,
	author = {Tong, W. and Hussain, A. and Wang, X. B. and Maharjan, S.},
	journal = {IEEE Access},
	pages = {10823-10843},
	title = {Artificial intelligence for vehicle-to-everything: A survey},
	volume = {7},
	year = {2019}}

@article{grigorescu2020survey,
	author = {Grigorescu, S. and Trasnea, B. and Cocias, T. and Macesanu, G.},
	journal = {Journal of field robotics},
	number = {3},
	pages = {362-386},
	title = {A survey of deep learning techniques for autonomous driving},
	volume = {37},
	year = {2020}}

@article{ferguson2007effectiveness,
	author = {Ferguson, S. A.},
	journal = {Traffic injury prevention},
	number = {4},
	pages = {329-338},
	title = {The effectiveness of electronic stability control in reducing real-world crashes: a literature review},
	volume = {8},
	year = {2007}}

@article{singh2021highway,
	author = {Singh, R. and Sharma, R. and Akram, S. V. and Gehlot, A. and Buddhi, D. and Malik, P. K. and Arya, R.},
	journal = {Safety science},
	pages = {105407},
	title = {Highway 4.0: Digitalization of highways for vulnerable road safety development with intelligent IoT sensors and machine learning},
	volume = {143},
	year = {2021}}

@article{huang2021systematic,
	author = {Huang, D. and Wang, S. and Liu, Z.},
	journal = {International Journal of Disaster Risk Reduction},
	pages = {102412},
	title = {A systematic review of prediction methods for emergency management},
	volume = {62},
	year = {2021}}

@article{saleem2022smart,
	author = {Saleem, M. and Abbas, S. and Ghazal, T. M. and Khan, M. A. and Sahawneh, N. and Ahmad, M.},
	journal = {Egyptian Informatics Journal},
	number = {3},
	pages = {417-426},
	title = {Smart cities: Fusion-based intelligent traffic congestion control system for vehicular networks using machine learning techniques},
	volume = {23},
	year = {2022}}

@article{thibaud2018internet,
	author = {Thibaud, M. and Chi, H. and Zhou, W. and Piramuthu, S.},
	journal = {Decision Support Systems},
	pages = {79-95},
	title = {Internet of Things (IoT) in high-risk Environment, Health and Safety (EHS) industries: A comprehensive review},
	volume = {108},
	year = {2018}}

@inproceedings{arinta2019natural,
	author = {Arinta, R. R. and WR, E. A.},
	booktitle = {International Conference on Information Technology, Information Systems and Electrical Engineering},
	title = {Natural disaster application on big data and machine learning: A review},
	year = {2019}}

@article{huamani2020machine,
	author = {Huaman{\'\i}, E. L. and Alva, M. A. and Roman-Gonzalez, A.},
	journal = {International Journal of Advanced Computer Science and Applications},
	number = {4},
	title = {Machine learning techniques to visualize and predict terrorist attacks worldwide using the global terrorism database},
	volume = {11},
    pages = {562-570},
	year = {2020}}

@article{garriga2022machine,
	author = {Garriga, R. and Mas, J. and Abraha, S. and Nolan, J. and Harrison, O. and Tadros, G. and Matic, A.},
	journal = {Nature medicine},
	number = {6},
	pages = {1240-1248},
	title = {Machine learning model to predict mental health crises from electronic health records},
	volume = {28},
	year = {2022}}

@article{levinson2003value,
	author = {Levinson, D.},
	journal = {Transportation Research Part C: Emerging Technologies},
	number = {1},
	pages = {75-87},
	title = {The value of advanced traveler information systems for route choice},
	volume = {11},
	year = {2003}}

@article{al2021hybrid,
	author = {Al Fararni, K. and Nafis, F. and Aghoutane, B. and Yahyaouy, A. and Riffi, J. and Sabri, A.},
	journal = {Big Data Mining and Analytics},
	number = {1},
	pages = {47-55},
	title = {Hybrid recommender system for tourism based on big data and AI: A conceptual framework},
	volume = {4},
	year = {2021}}

@article{cepeda2022deep,
	author = {Cepeda-Pacheco, J. C. and Domingo, M. C.},
	journal = {Neural Computing and Applications},
	number = {10},
	pages = {7691-7709},
	title = {Deep learning and Internet of Things for tourist attraction recommendations in smart cities},
	volume = {34},
	year = {2022}}

@inproceedings{wang2018will,
	author = {Wang, D. and Zhang, J. and Cao, W. and Li, J. and Zheng, Y.},
	booktitle = {AAAI conference on artificial intelligence},
	title = {When will you arrive? estimating travel time based on deep neural networks},
	  pages        = {2500--2507},
    year = {2018}}

@inproceedings{siripanpornchana2016travel,
	author = {Siripanpornchana, C. and Panichpapiboon, S. and Chaovalit, P.},
	booktitle = {IEEE Region 10 International Conference TENCON},
	title = {Travel-time prediction with deep learning},
	year = {2016}}

@article{boysen2021last,
	author = {Boysen, N. and Fedtke, S. and Schwerdfeger, S.},
	journal = {Or Spectrum},
	number = {1},
	pages = {1-58},
	title = {Last-mile delivery concepts: a survey from an operational research perspective},
	volume = {43},
	year = {2021}}

@inproceedings{avatefipour2018traffic,
	author = {Avatefipour, O. and Sadry, F.},
	booktitle = {IEEE International Conference on Electro/Information Technology},
	title = {Traffic management system using IoT technology-A comparative review},
	year = {2018}}

@article{lombardi2021model,
	author = {Lombardi, C. and Picado-Santos, L. and Annaswamy, A. M.},
	journal = {Applied Sciences},
	number = {11},
	pages = {4778},
	title = {Model-based dynamic toll pricing: An overview},
	volume = {11},
	year = {2021}}

@article{sarabia2019seaport,
	author = {Sarabia-Jacome, D. and Palau, C. E. and Esteve, M. and Boronat, F.},
	journal = {IEEE Access},
	pages = {4372-4382},
	title = {Seaport data space for improving logistic maritime operations},
	volume = {8},
	year = {2019}}

@article{marchet2014environmental,
	author = {Marchet, G. and Melacini, M. and Perotti, S.},
	journal = {Journal of Manufacturing Technology Management},
	number = {6},
	pages = {775-811},
	title = {Environmental sustainability in logistics and freight transportation: A literature review and research agenda},
	volume = {25},
	year = {2014}}

@article{qi2020framework,
	author = {Qi, B. and Costin, A. and Jia, M.},
	journal = {Travel behaviour and society},
	pages = {10-23},
	title = {A framework with efficient extraction and analysis of Twitter data for evaluating public opinions on transportation services},
	volume = {21},
	year = {2020}}

@article{ruiz2016social,
	author = {Ruiz, T. and Mars, L. and Arroyo, R. and Serna, A.},
	journal = {Transportation research procedia},
	pages = {446-452},
	title = {Social networks, big data and transport planning},
	volume = {18},
	year = {2016}, month={February}}

@article{pandian2019artificial,
	author = {Pandian, A. P.},
	journal = {Journal of Artificial Intelligence},
	number = {02},
	pages = {63-72},
	title = {Artificial intelligence application in smart warehousing environment for automated logistics},
	volume = {1},
	year = {2019}}

@inproceedings{borgi2017big,
	author = {Borgi, T. and Zoghlami, N. and Abed, M. and Naceur, M. S.},
	booktitle = {IEEE International conference on Advanced Logistics and Transport},
	title = {Big data for operational efficiency of transport and logistics: a review},
	year = {2017}}

@article{de2017traffic,
	author = {De Souza, A. M. and Brennand, C. and Yokoyama, R. S. and Donato, E. A. and Madeira, E. R. M. and Villas, L. A.},
	journal = {International Journal of Distributed Sensor Networks},
	number = {4},
	pages = {1550147716683612},
	title = {Traffic management systems: A classification, review, challenges, and future perspectives},
	volume = {13},
	year = {2017}}

@article{aringhieri2022fairness,
	author = {Aringhieri, R. and Bigharaz, S. and Duma, D. and Guastalla, A.},
	journal = {Central European journal of operations research},
	pages = {1-23},
	title = {Fairness in ambulance routing for post disaster management},
	year = {2022}}

@article{jiang2019emergency,
	author = {Jiang, Y. and Yuan, Y.},
	journal = {International journal of environmental research and public health},
	number = {5},
	pages = {779},
	title = {Emergency logistics in a large-scale disaster context: Achievements and challenges},
	volume = {16},
	year = {2019}}

@article{gao2023towards,
	author = {Gao, J. and Peng, P. and Lu, F. and Claramunt, C. and Xu, Y.},
	journal = {Information Processing \& Management},
	number = {4},
	pages = {103369},
	title = {Towards travel recommendation interpretability: Disentangling tourist decision-making process via knowledge graph},
	volume = {60},
	year = {2023}}

@article{asaithambi2023thematic,
	author = {Asaithambi, S. P. R. and Venkatraman, R. and Venkatraman, S.},
	journal = {Technologies},
	number = {1},
	pages = {28},
	title = {A thematic travel recommendation system using an augmented big data analytical model},
	volume = {11},
	year = {2023}}

@inproceedings{chi2021design,
	author = {Chi, D. and Tang, C. and Yin, C.},
	booktitle = {Journal of Physics: Conference Series},
	title = {Design and Implementation of Hotel Big Data Analysis Platform Based on Hadoop and Spark},
	year = {2021}}

@inproceedings{lyu2022towards,
	author = {Lyu, W. and Zhang, K. and Guo, B. and Hong, Z. and Yang, G. and Wang, G. and Yang, Y. and Liu, Y. and Zhang, D.},
	booktitle = {ACM International Conference on Information \& Knowledge Management},
	title = {Towards fair workload assessment via homogeneous order grouping in last-mile delivery},
	year = {2022}}

@article{guo2020optimized,
	author = {Guo, K. and Hu, Y. and Qian, Z. and Liu, H. and Zhang, K. and Sun, Y. and Gao, J. and Yin, B.},
	journal = {IEEE Transactions on Intelligent Transportation Systems},
	number = {2},
	pages = {1138-1149},
	title = {Optimized graph convolution recurrent neural network for traffic prediction},
	volume = {22},
	year = {2020}}

@article{ye2020cyber,
	author = {Ye, J. and Guo, L. and Yang, B. and Li, F. and Du, L. and Guan, L. and Song, W.},
	journal = {IEEE Journal of Emerging and Selected Topics in Power Electronics},
	number = {4},
	pages = {4639-4657},
	title = {Cyber-physical security of powertrain systems in modern electric vehicles: Vulnerabilities, challenges, and future visions},
	volume = {9},
	year = {2020}}

@article{holmstrom2024stalled,
	author = {Holmstr{\"o}m, J. and Kostis, A. and Galariotis, E. and Roubaud, D. and Zopounidis, C.},
	journal = {Industrial Marketing Management},
	pages = {16-26},
	title = {Stalled data flows in digital innovation networks: Underlying mechanisms and the role of related variety},
	volume = {121},
	year = {2024}}

@article{rathore2021cyber,
	author = {Rathore, M. M. and Attique Shah, S. and Awad, A. and Shukla, D. and Vimal, S. and Paul, A.},
	journal = {Sustainability},
	number = {14},
	pages = {7606},
	title = {A cyber-physical system and graph-based approach for transportation management in smart cities},
	volume = {13},
	year = {2021}}

@inproceedings{tarneberg2016experiences,
	author = {T{\"a}rneberg, W. and Chandrasekaran, V. and Humphrey, M.},
	booktitle = {International Conference on Utility and Cloud Computing},
	title = {Experiences creating a framework for smart traffic control using aws iot},
	year = {2016}}

@inproceedings{pandhare2017real,
	author = {Pandhare, K. R. and Shah, M. A.},
	booktitle = {International conference on inventive communication and computational technologies},
	title = {Real time road traffic event detection using Twitter and spark},
	year = {2017}}

@article{alomari2021iktishaf,
	author = {Alomari, E. and Katib, I. and Albeshri, A. and Yigitcanlar, T. and Mehmood, R.},
	journal = {Sensors},
	number = {9},
	pages = {2993},
	title = {Iktishaf+: a big data tool with automatic labeling for road traffic social sensing and event detection using distributed machine learning},
	volume = {21},
	year = {2021}}

@article{kovalchuk2018distributed,
	author = {Kovalchuk, S. V. and Krotov, E. and Smirnov, P. A. and Nasonov, D. A. and Yakovlev, A. N.},
	journal = {Future Generation Computer Systems},
	pages = {144-154},
	title = {Distributed data-driven platform for urgent decision making in cardiological ambulance control},
	volume = {79},
	year = {2018}}

@inproceedings{seal2019real,
	author = {Seal, A. and Mukherjee, A.},
	booktitle = {IEEE SoutheastCon},
	title = {Real time accident prediction and related congestion control using spark streaming in an AWS EMR cluster},
	year = {2019}}

@article{alotaibi2020sehaa,
	author = {Alotaibi, S. and Mehmood, R. and Katib, I. and Rana, O. and Albeshri, A.},
	journal = {Applied Sciences},
	number = {4},
	pages = {1398},
	title = {Sehaa: A big data analytics tool for healthcare symptoms and diseases detection using Twitter, Apache Spark, and Machine Learning},
	volume = {10},
	year = {2020}}

@article{rathore2016real,
	author = {Rathore, M. M. and Ahmad, A. and Paul, A. and Wan, J. and Zhang, D.},
	journal = {Journal of medical systems},
	pages = {1-10},
	title = {Real-time medical emergency response system: exploiting IoT and big data for public health},
	volume = {40},
	year = {2016}}

@inproceedings{noori2020fire,
	author = {Noori, M. A. R. and Mehra, R.},
	booktitle = {IEEE International Conference on Industrial and Information Systems},
	title = {Fire emergency detection from twitter using supervised principal},
	year = {2020}}

@article{silva2021advancing,
	author = {Silva, N. and Barros, J. and Santos, M. Y. and Costa, C. and Cortez, P. and Carvalho, M. S. and Goncalves, J. N. C.},
	journal = {Electronics},
	number = {18},
	pages = {2221},
	title = {Advancing logistics 4.0 with the implementation of a big data warehouse: a demonstration case for the automotive industry},
	volume = {10},
	year = {2021}}

@article{figueiras2018data,
	author = {Figueiras, P. and Guerreiro, G. and Silva, R. and Costa, R. and Jardim-Gon{\c c}alves, R.},
	journal = {Learning Systems: From Theory to Practice},
	pages = {281-301},
	title = {Data processing and harmonization for intelligent transportation systems: an application scenario on highway traffic flows},
	year = {2018}}

@inproceedings{daaif2019efficient,
	author = {Daaif, A. and Bouattane, O. and Youssfi, M. and Snineh, S. M.},
	booktitle = {Lecture Notes in Real-Time Intelligent Systems},
	title = {An Efficient Traffic Monitoring Model Using a Stream Processing Platform Based on Smart Highways Events Generator},publisher="Springer International Publishing", address="Cham",
	year = {2019}}

@inproceedings{khazaei2016realtime,
	author = {Khazaei, H. and Veleda, R. and Litoiu, M. and Tizghadam, A.},
	booktitle = {IEEE World Forum on Internet of Things},
	title = {Realtime big data analytics for event detection in highways},
      pages={472-477},
	year = {2016}}

@article{tseng2018congestion,
	author = {Tseng, F.-H. and Hsueh, J.-H. and Tseng, C.-W. and Yang, Y.-T. and Chao, H.-C. and Chou, L.-D.},
	journal = {IEEE Access},
	pages = {57311-57323},
	title = {Congestion prediction with big data for real-time highway traffic},
	volume = {6},
	year = {2018}}

@article{sundareswaran2020real,
	author = {Sundareswaran, A. and Lavanya, K.},
	journal = {International Journal of Intelligent Information Technologies},
	number = {4},
	pages = {19-36},
	title = {Real-time vehicle traffic prediction in apache spark using ensemble learning for deep neural networks},
	volume = {16},
	year = {2020}}

@article{yu2020dissecting,
	author = {Yu, J. and Fu, Z. and Sarwat, M.},
	journal = {Distributed and Parallel Databases},
	number = {4},
	pages = {963-994},
	title = {Dissecting geosparksim: a scalable microscopic road network traffic simulator in apache spark},
	volume = {38},
	year = {2020}}

@inproceedings{fu2019demonstrating,
	author = {Fu, Z. and Yu, J. and Sarwat, M.},
	booktitle = {International Symposium on Spatial and Temporal Databases},
	title = {Demonstrating geosparksim: A scalable microscopic road network traffic simulator based on apache spark},
	year = {2019}}

@inproceedings{rathore2016exploiting,
	author = {Rathore, M. M. and Ahmad, A. and Paul, A. and Thikshaja, U. K.},
	booktitle = {IEEE Region 10 Symposium},
	title = {Exploiting real-time big data to empower smart transportation using big graphs},
	year = {2016}}

@inproceedings{castellanos2021smart,
	author = {Castellanos, C. and Perez, B. and Correal, D.},
	booktitle = {Smart Cities: A Data Analytics Perspective},
	title = {Smart transportation: A reference architecture for big data analytics},publisher="Springer International Publishing", address="Cham",
	year = {2021}}

@article{chen2017modeling,
	author = {Chen, J. and Cao, Y. and Sun, D.},
	journal = {Journal of Aerospace Information Systems},
	number = {9},
	pages = {504-516},
	title = {Modeling, optimization, and operation of large-scale air traffic flow management on spark},
	volume = {14},
	year = {2017}}

@inproceedings{vonitsanos2021forecasting,
	author = {Vonitsanos, G. and Panagiotakopoulos, T. and Kanavos, A. and Tsakalidis, A.},
	booktitle = {IFIP International conference on artificial intelligence applications and innovations},
	title = {Forecasting air flight delays and enabling smart airport services in apache spark},
    pages="407--417",
	year = {2021}}

@article{oneto2018train,
	author = {Oneto, L. and Fumeo, E. and Clerico, G. and Canepa, R. and Papa, F. and Dambra, C. and Mazzino, N. and Anguita, D.},
	journal = {Big data research},
	pages = {54-64},
	title = {Train delay prediction systems: a big data analytics perspective},
	volume = {11},
	year = {2018}}

@inproceedings{vo2022practical,
	author = {Vo, M.-T. and Tran, T.-V. and Pham, D.-T. and Do, T.-H.},
	booktitle = {IEEE International Conference on Communication, Networks and Satellite},
	title = {A Practical Real-Time Flight Delay Prediction System using Big Data Technology},
	year = {2022}}

@inproceedings{antunes2019big,
	author = {Antunes, H. and Figueiras, P. and Costa, R. and Teixeira, J. and Jardim-Gon{\c c}alves, R.},
	booktitle = {ICIST International Conference on Information Society and Techology},
	title = {Big data analytics for extracting mobility patterns in a large urban center},
	pages={124-129},
    year = {2019}}

@article{luckner2020iot,
	author = {Luckner, M. and Grzenda, M. and Kunicki, R. and Legierski, J.},
	journal = {ACM Transactions on Internet Technology},
	number = {3},
	pages = {1-30},
	title = {IoT architecture for urban data-centric services and applications},
	volume = {20},
	year = {2020}}

@article{mulerikkal2021jp,
	author = {Mulerikkal, J. and Thandassery, S. and Rejathalal, V. and Ayyappan, B.},
	journal = {IEEE Transactions on Intelligent Transportation Systems},
	number = {7},
	pages = {9146-9156},
	title = {JP-DAP: An intelligent data analytics platform for metro rail transport systems},
	volume = {23},
	year = {2021}}

@inproceedings{dimokas2021cloud,
	author = {Dimokas, N. and Margaritis, D. and Gaetani, M. and Favenza, A.},
	booktitle = {Mobility-as-a-Service Systems},
	title = {A cloud-based big data architecture for an intelligent green truck}, publisher="Springer International Publishing", address="Cham",
	year = {2021}}

@article{kumar2018design,
	author = {Kumar, R. S. and Rani, C. and Kumar, P. G.},
	journal = {International Journal of Logistics Systems and Management},
	number = {2},
	pages = {249-266},
	title = {Design of smart logistics transportation system using MapReduce intelligent water drops algorithm in Hadoop environment},
	volume = {31},
	year = {2018}}

@article{belfadel2023conceptual,
	author = {Belfadel, A. and H{\"o}rl, S. and Tapia, R. J. and Politaki, D. and Kureshi, I. and Tavasszy, L. and Puchinger, J.},
	journal = {Computers, Environment and Urban Systems},
	pages = {101989},
	title = {A conceptual digital twin framework for city logistics},
	volume = {103},
	year = {2023}}

@article{selvaraj2020offline,
	author = {Selvaraj, S. and Kim, H. and Choi, E.},
	journal = {Journal of Information Processing Systems},
	number = {2},
	title = {Offline-to-Online Service and Big Data Analysis for End-to-end Freight Management System.},
      pages={377},
	volume = {16},
	year = {2020}}

@inproceedings{provatas2020selis,
	author = {Provatas, N. and Kassela, E. and Chalvantzis, N. and Bakogiannis, A. and Giannakopoulos, I. and Koziris, N. and Konstantinou, I.},
	booktitle = {IEEE International Conference on Big Data},
	title = {Selis bda: Big data analytics for the logistics domain},
	year = {2020}}

@inproceedings{mhand2019scalable,
	author = {Mhand, M. A. and Boulmakoul, A. and Badir, H.},
	booktitle = {New Challenges in Data Sciences: Acts of the Second Conference of the Moroccan Classification Society},
	title = {Scalable and distributed architecture based on Apache Spark Streaming and PROM6 for processing RoRo terminals logs},
	year = {2019}}

@article{alexakis2022distributed,
	author = {Alexakis, T. and Peppes, N. and Demestichas, K. and Adamopoulou, E.},
	journal = {Sensors},
	number = {1},
	pages = {357},
	title = {A distributed big data analytics architecture for vehicle sensor data},
	volume = {23},
	year = {2022}}

@article{zhang2017carstream,
	author = {Zhang, M. and Wo, T. and Xie, T. and Lin, X. and Liu, Y.},
	journal = {Proceedings of the VLDB Endowment},
	number = {12},
	pages = {1766-1777},
	title = {Carstream: an industrial system of big data processing for internet-of-vehicles},
	volume = {10},
	year = {2017}}

@inproceedings{xu2016supporting,
	author = {Xu, W. and Juri, N. R. and Gupta, A. and Deering, A. and Bhat, C. and Kuhr, J. and Archer, J.},
	booktitle = {IEEE International Conference on Big Data},
	title = {Supporting large scale connected vehicle data analysis using HIVE},
	year = {2016}}

@article{nair2016real,
	author = {Nair, Y. C. and Neethu, P. V. and Menon, V. K. and Soman, K. P.},
	journal = {Indian Journal of Science and Technology},
	number = {30},
	title = {Real time vehicular data analytics utilising bigdata platforms and cost effective ECU networks},
    pages={1-5},
	volume = {9},
	year = {2016}}

@article{kahveci2022end,
	author = {Kahveci, S. and Alkan, B. and Mus'ab H, A. and Ahmad, B. and Harrison, R.},
	journal = {Journal of Manufacturing Systems},
	pages = {214-223},
	title = {An end-to-end big data analytics platform for IoT-enabled smart factories: A case study of battery module assembly system for electric vehicles},
	volume = {63},
	year = {2022}}

@inproceedings{triguero2017vehicle,
	author = {Triguero, I. and Figueredo, G. P. and Mesgarpour, M. and Garibaldi, J. M. and John, R. I.},
	booktitle = {IEEE Trustcom/BigDataSE/ICESS},
	title = {Vehicle incident hot spots identification: An approach for big data},
	year = {2017}}

@article{yoo2020implementation,
	author = {Yoo, A. and Shin, S. and Lee, J. and Moon, C.},
	journal = {Applied Sciences},
	number = {21},
	pages = {7858},
	title = {Implementation of a sensor big data processing system for autonomous vehicles in the C-ITS environment},
	volume = {10},
	year = {2020}}

@article{priyanka2021real,
	author = {Priyanka, E. B. and Shankar, M. G. and Tharun, S. and Ravisankar, S. and Saravanan, S. N. and Kumar, B. B. and Pugazhenthi, C.},
	journal = {International Journal of Computing and Digital System},
	pages = {1337-1357},
	title = {Real-time performance analysis of multiple parameters of automotive sensor's can data to predict vehicle driving efficiency},
	year = {2021}}

@article{dinten2023fleet,
	author = {Dint{\'e}n, R. and Garc{\'\i}a, S. and Zorrilla, M.},
	journal = {Procedia Computer Science},
	pages = {806-815},
	title = {Fleet management systems in Logistics 4.0 era: a real time distributed and scalable architectural proposal},
	volume = {217},
	year = {2023}, month={January}}

@article{sbai2020real,
	author = {Sbai, I. and Krichen, S.},
	journal = {Procedia Computer Science},
	pages = {938-947},
	title = {A real-time decision support system for big data analytic: A case of dynamic vehicle routing problems},
	volume = {176},
	year = {2020}, month={October}}

@inproceedings{ademola2022improving,
	author = {Ademola, O. F. and Misra, S. and Agrawal, A.},
	booktitle = {International Conference on Recent Innovations in Computing},
	title = {Improving Real-Time Intelligent Transportation Systems in Predicting Road Accident},
    pages="225--239",
	year = {2022}}

@article{lapatta2022ecotourism,
	author = {Lapatta, N. T.},
	journal = {Journal of Social Science},
	number = {3},
	pages = {534-546},
	title = {Ecotourism recommendations based on sentiments using skyline query and apache-spark},
	volume = {3},
	year = {2022}}

@inproceedings{ambolkar2022hotel,
	author = {Ambolkar, R. and Bhagat, A. and Buga, B. and Gharat, S.},
	booktitle = {International Conference on Artificial Intelligence and Smart Energy},
	title = {Hotel Recommendation System using advanced efficiency and accuracy with modified BERT technique},
	year = {2022}}

@inproceedings{malkawi2023real,
	author = {Malkawi, M. and Alhajj, R.},
	booktitle = {IEEE International Conference on Information Reuse and Integration for Data Science},
	title = {Real-time web-based International Flight Tickets Recommendation System via Apache Spark},
	year = {2023}}

@inproceedings{petrozziello2017data,
	author = {Petrozziello, A. and Jordanov, I.},
	booktitle = {International Conference Modelling, Identification and Control},
	title = {Data analytics for online travelling recommendation system: a case study},
	year = {2017}}

@inproceedings{bellandi2019designing,
	author = {Bellandi, V. and Ceravolo, P. and Damiani, E. and Tacchini, E.},
	booktitle = {Innovations in Big Data Mining and Embedded Knowledge},
	title = {Designing a recommender system for touristic activities in a big data as a service platform},publisher="Springer International Publishing", address="Cham",
	year = {2019}}

@article{suanpang2020integration,
	author = {Suanpang, P. and Jamjuntr, P.},
	journal = {International Journal of Machine Learning and Computing},
	number = {6},
	pages = {714-722},
	title = {The integration of a big data framework and a mobile application on the iOS platform to support smart tourism},
	volume = {10},
	year = {2020}}

@article{lee2018design,
	author = {Lee, C. K. M. and Lv, Y. and Ng, K. K. H. and Ho, W. and Choy, K. L.},
	journal = {International Journal of Production Research},
	number = {8},
	pages = {2753-2768},
	title = {Design and application of Internet of things-based warehouse management system for smart logistics},
	volume = {56},
	year = {2018}}

@inproceedings{sugathan2021espy,
	author = {Sugathan, D. and Revanth, M. and Kumar, V. P. and Desai, S. and Keshava, L.},
	booktitle = {International Conference on Design Innovations for 3Cs Compute Communicate Control},
	title = {eSPY Carting Systems: The Real-Time Tracking Mechanism Architecture},
	year = {2021}}

@article{ruan2021filling,
	author = {Ruan, S. and Fu, X. and Long, C. and Xiong, Z. and Bao, J. and Li, R. and Chen, Y. and Wu, S. and Zheng, Y.},
	journal = {IEEE Transactions on Knowledge and Data Engineering},
	number = {2},
	pages = {1528-1540},
	title = {Filling delivery time automatically based on couriers' trajectories},
	volume = {35},
	year = {2021}}

@article{hong2019crowdsourcing,
	author = {Hong, H. and Li, X. and He, D. and Zhang, Y. and Wang, M.},
	journal = {IEEE Access},
	pages = {26268-26277},
	title = {Crowdsourcing incentives for multi-hop urban parcel delivery network},
	volume = {7},
	year = {2019}}

@inproceedings{morariu2024analysing,
	author = {Morariu, A.-R. and Ahmad, T. and Iancu, B. and Poikonen, J. and Bj{\"o}rkqvist, J.},
	booktitle = {IEEE International Conference on Industrial Cyber-Physical Systems},
	title = {Analysing MLOps and its Applicability in the Maritime Domain through a Systematic Mapping Study},
	year = {2024}}

@inproceedings{zeller2023safety,
	author = {Zeller, M. and Waschulzik, T. and Rothfelder, M. and Klein, C.},
	booktitle = {IEEE International Symposium on Software Reliability Engineering Workshops},
	title = {Safety Assurance of a Driverless Regional Train-Insight in the safe. trAIn Project},
	year = {2023}}

@article{faubel2024mlops,
	author = {Faubel, L. and Schmid, K.},
	journal = {arXiv preprint arXiv:2407.09107},
	title = {MLOps: A Multiple Case Study in Industry 4.0},
	year = {2024}}

@article{urias2023mlops,
	author = {Urias, I. B. and Rossi, R.},
	journal = {EAI Endorsed Transactions on Smart Cities},
	number = {3},
	title = {MLOPS and Microservices Frameworks in the Perspective of Smart Cities.},
	volume = {7},
    pages="1--11",
	year = {2023}}

@inproceedings{sengul2022application,
	author = {Sengul, M. K. and Tarhan, C. and Tecim, V.},
	booktitle = {Innovations in Intelligent Systems and Applications Conference},
	title = {Application of intelligent transportation system data using big data technologies},
	year = {2022}}

@article{remil2024aiops,
	author = {Remil, Y. and Bendimerad, A. and Mathonat, R. and Kaytoue, M.},
	journal = {arXiv preprint arXiv:2404.01363},
	title = {Aiops solutions for incident management: Technical guidelines and a comprehensive literature review},
	year = {2024}}

@article{jana2023framework,
	author = {Jana, A. K.},
	journal = {Journal of Artificial Intelligence, Machine Learning and Data Science},
	number = {3},
	pages = {575-579},
	title = {Framework for Automated Machine Learning Workflows: Building End-to-End MLOps Tools for Scalable Systems on AWS},
	volume = {1},
	year = {2023}}

@inproceedings{angarita2020general,
	author = {Angarita-Zapata, J. S. and Masegosa, A. D. and Triguero, I.},
	booktitle = {International Conference on Information Processing and Management of Uncertainty in Knowledge-Based Systems},
	title = {General-purpose automated machine learning for transportation: A case study of auto-sklearn for traffic forecasting},
    pages={728–744},
	year = {2020}}

@inproceedings{conlan2023real,
	author = {Conlan, C. and Oakley, J. and Demirci, G. V. and Sfyridis, A. and Ferhatosmanoglu, H.},
	booktitle = {CEUR Workshop Proceedings},
	title = {Real-time spatio-temporal forecasting with dynamic urban event and vehicle-level flow information},
	volume = {3379},
	year = {2023}}

@inproceedings{cha2023development,
	author = {Cha, J. and Jeong, H. and Han, S. and Kim, D. and Oh, J. and Hwang, S. and Park, B.},
	booktitle = {International Conference on Human-Computer Interaction},
	title = {Development of MLOps Platform Based on Power Source Analysis for Considering Manufacturing Environment Changes in Real-Time Processes},
    pages="224--236",
	year = {2023}}

@article{khan2024optimizing,
	author = {Khan, A. and Ahmed, A.},
	journal = {Emerging Trends in Machine Intelligence and Big Data},
	number = {1},
	pages = {18-37},
	title = {Optimizing Retail Operations, Inventory Management and Sales Forecasting with Big Data and AI in China},
	volume = {16},
	year = {2024}}

@article{karmaker2021automl,
	author = {Karmaker, S. K. and Hassan, M. M. and Smith, M. J. and Xu, L. and Zhai, C. and Veeramachaneni, K.},
	journal = {ACM Computing Surveys},
	number = {8},
	pages = {1-36},
	title = {Automl to date and beyond: Challenges and opportunities},
	volume = {54},
	year = {2021}}

@article{oakley2023foresight,
	author = {Oakley, J. and Conlan, C. and Demirci, G. V. and Sfyridis, A. and Ferhatosmanoglu, H.},
	journal = {GeoInformatica},
	title = {Foresight Plus: Scalable Serverless Real-Time Spatio-Temporal Traffic Forecasting},
    volume={28}, pages={649–677}, number={4},
	year = {2023}}

@article{mishra2023autonomous,
	author = {Mishra, S. and Palanisamy, P.},
	journal = {IEEE Access},
	pages = {136318-136349},
	title = {Autonomous advanced aerial mobility-An end-to-end autonomy framework for UAVs and beyond},
	volume = {11},
	year = {2023}}

@inproceedings{peramo2022intelligent,
	author = {Peramo, E. and Aborot, J. and Bensig, J. R. and Deiparine, P. J. and Others},
	booktitle = {IEEE International Conference on Computing},
	title = {Intelligent Flood Detection using Traffic Surveillance Images based on Convolutional Neural Network and Image Parsing},
	year = {2022}}

\begin{IEEEbiography}
[{\includegraphics[width=1in,height=1.25in,clip,keepaspectratio]{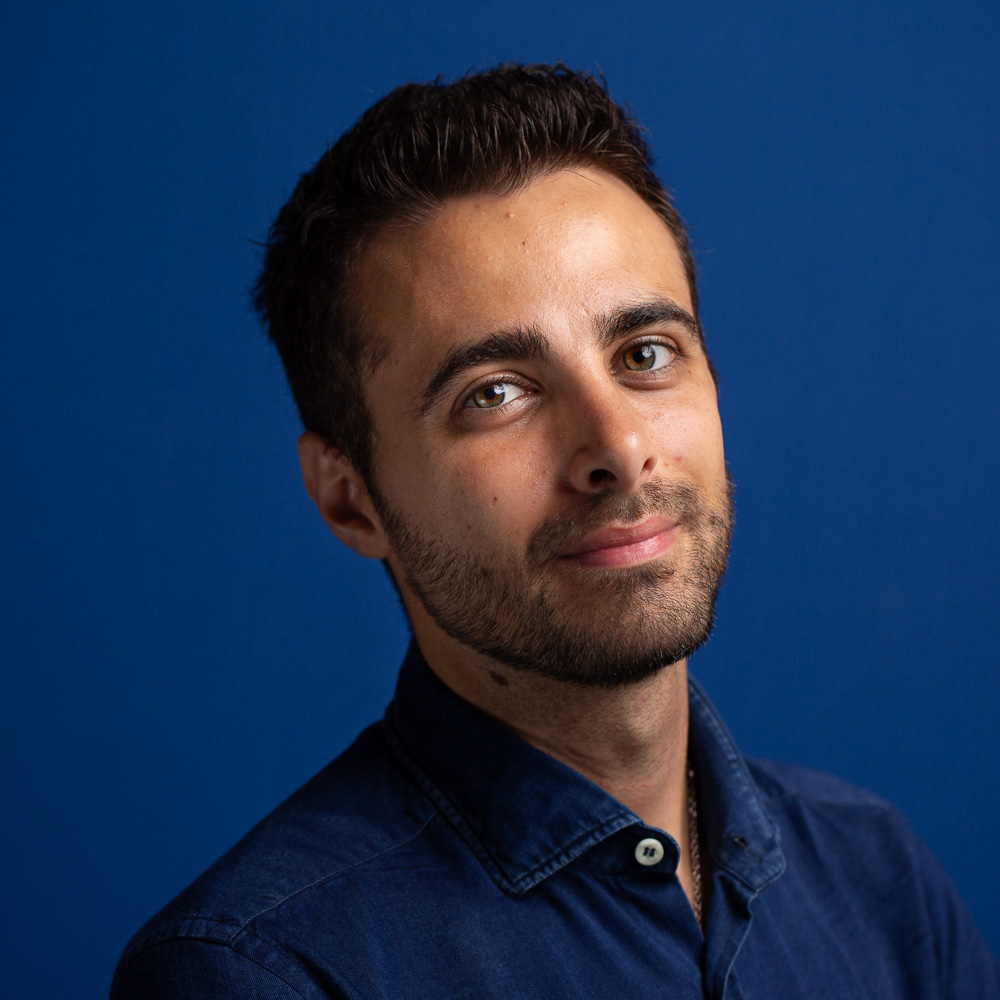}}]{Antonio Emanuele Cinà} (Member, IEEE) is an Assistant Professor at the University of Genoa, Italy. He received his Ph.D. (cum laude) in Computer Science from Ca' Foscari University of Venice in 2023, defending a thesis on the vulnerabilities and emerging risks arising from the malicious use of training data in AI. His research interests encompass all aspects of AI system security and the study of their trustworthiness, with primary expertise in training (poisoning) and inference-time (evasion) attacks. Antonio has also been investigating the capabilities of AI for cybercrime detection and prevention.
\end{IEEEbiography}

\begin{IEEEbiography}
[{\includegraphics[width=1in,height=1.25in,clip,keepaspectratio]{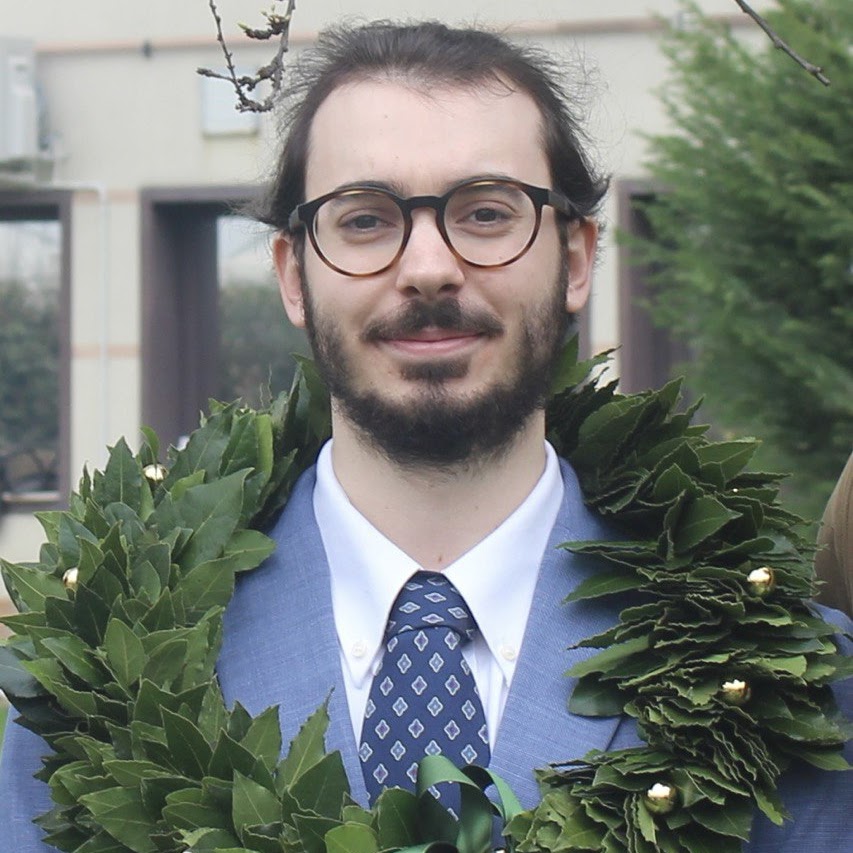}}]{Giovanni Scodeller} 
is a Ph.D. student at the University of Genoa and a member of the sAIfer Lab, a joint lab between the University of Genoa and the University of Cagliari focused on machine learning security. He obtained his Ph.D. in 2023 from Ca’ Foscari University in Venice, where he also completed his Master’s degree in Computer Science. His research interests include uncertainty quantification for machine learning and machine learning security.
\end{IEEEbiography}

\begin{IEEEbiography}[{\includegraphics[width=1in,height=1.25in,clip,keepaspectratio]{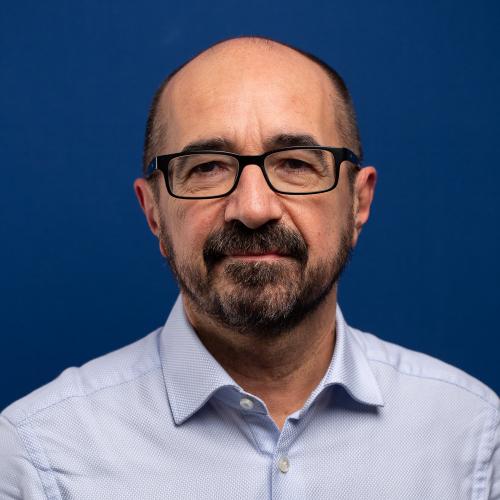}}]{Davide Anguita}
received the ``Laurea” degree in Electronic Engineering and a Ph.D. degree in Computer Science and Electronic Engineering from the University of Genoa, Genoa, Italy, in 1989 and 1993, respectively. After working as
a Research Associate at the International Computer Science Institute, Berkeley, CA, on special-purpose processors for neurocomputing, he returned to the University of Genoa. He is currently Full Professor of Computer Engineering
with the Department of Informatics, Bio-Engineering, Robotics, and Systems Engineering (DIBRIS). His current research focuses on the theory and application of kernel methods and artificial neural networks.
\end{IEEEbiography}

\begin{IEEEbiography}
[{\includegraphics[width=1in,height=1.25in,clip,keepaspectratio]{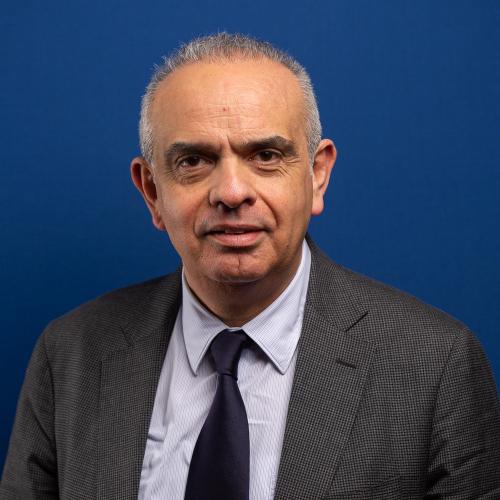}}]{Fabio Roli}
is a Full Professor of Computer Engineering at the University of Genova and the
University of Cagliari, Italy, and Founding Director of the Pattern Recognition and Applications laboratory at the University of Cagliari. He has been
doing research on the design of pattern recognition and machine learning systems for thirty years. He has provided seminal contributions to the fields of multiple classifier systems and adversarial machine learning, and he has played a leading role
in the establishment and advancement of these research themes. He has been appointed Fellow of the IEEE and Fellow of the International Association for Pattern Recognition. He is a recipient of the Pierre Devijver Award for his contributions to statistical pattern recognition.
\end{IEEEbiography}

\begin{IEEEbiography}
[{\includegraphics[width=1in,height=1.25in,clip,keepaspectratio]{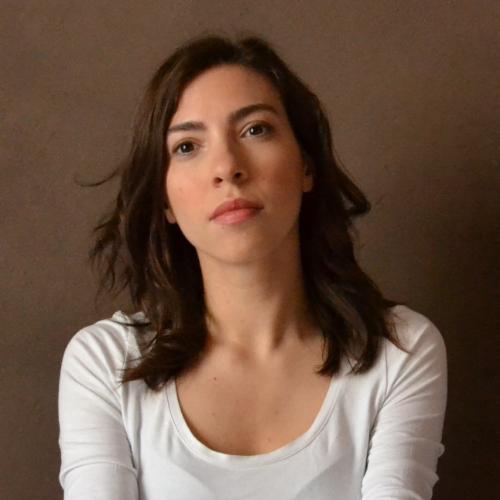}}]{Cecilia Caterina Pasquale} (Member, IEEE) received the bachelor’s degree in civil engineering from the Polytechnic University of Torino, Turin, Italy, in 2009, the master’s degree in transports and logistics engineering, and the Ph.D. degree in monitoring of systems and environmental risk management from the University of Genova, Genoa, Italy, in 2012 and 2016, respectively. She is currently an Assistant Professor with the University of Genova. Her research interests include modelling, optimization, and control methods applied to the field of transportation systems. She is also an Associate Editor for Control Engineering Practice. She is the Chair of the Technical Committee on Planning and Control of Transportation Networks of the IEEE Intelligent Transportation Systems Society.
\end{IEEEbiography}

\begin{IEEEbiography}
[{\includegraphics[width=1in,height=1.25in,clip,keepaspectratio]{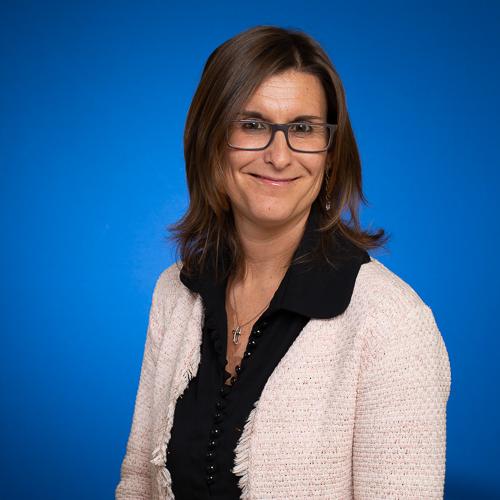}}]{Silvia Siri} (Senior Member, IEEE) received the Ph.D. degree in information and communication technologies from the University of Genova, Genoa, Italy, in 2006. She is currently an Associate Professor of automatic control with the University of Genova. She is the coauthor of more than 150 papers in international journals, chapters, and conference proceedings. Her research interests include modelling and control of freeway traffic systems, optimal management of electric mobility in smart cities, optimization and planning of freight logistics. Presently she is Senior Editor of the IEEE Transactions on Intelligent Transportation Systems and Associate Editor of IEEE Control Systems Letters. In the triennium 2019-2021 she served as a member of the Board of Governors of the IEEE Intelligent Transportation Systems Society. Since 2021 she has been Chair of the IEEE ITSS Italian Chapter.
\end{IEEEbiography}

\begin{IEEEbiography}
[{\includegraphics[width=1in,height=1.25in,clip,keepaspectratio]{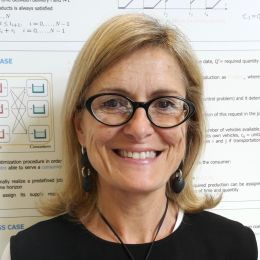}}]{Simona Sacone} 
(Senior Member, IEEE) is currently a Professor of automatic control with the Department of Informatics, Bioengineering, Robotics and Systems Engineering, University of Genova, Genoa, Italy, where she teaches systems theory, identification and estimation techniques, and sustainable systems modelling and where she also acts as the Coordinator of the Ph.D. degree Course on systems engineering. Her research interests include the optimization and control of complex physical processes by means of discreteevent and hybrid modelling, control approaches mainly focusing on freeway traffic control and logistic networks planning. She has authored and coauthored more than 150 papers published in international journals, international books, and international conference proceedings. She is also the Editor-in-Chief of the IEEE Transactions on Intelligent Transportation Systems. From 2019 to 2024, she was a member of the Board of Governors of the IEEE Intelligent Transportation Systems Society.
\end{IEEEbiography}

\begin{IEEEbiography}[{\includegraphics[width=1in,height=1.25in,clip,keepaspectratio]{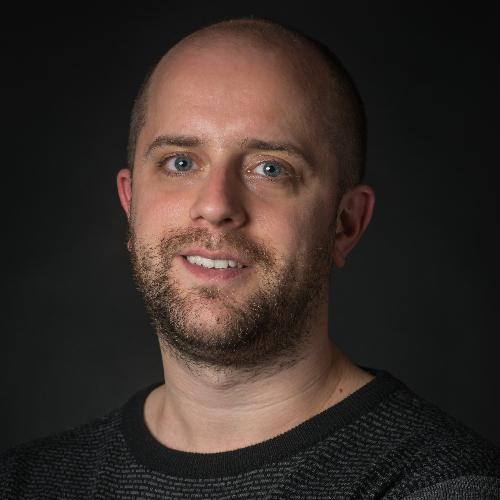}}]{Luca Oneto} born in 1986 in Rapallo, Italy, completed his BSc and MSc in Electronic Engineering at the University of Genoa in 2008 and 2010, respectively. In 2014, he earned his PhD in Computer Engineering from the same institution. From 2014 to 2016, he worked as a Postdoc in Computer Engineering at the University of Genoa, where he then served as an Assistant Professor from 2016 to 2019. Luca co-founded the company ZenaByte s.r.l. in 2018. In 2019, he became an Associate Professor in Computer Science at the University of Pisa, and from 2019 to 2024, he held the position of Associate Professor in Computer Engineering at the University of Genoa. Currently, he is a Full Professor in Computer Engineering at the University of Genoa. He has been coordinator and local responsible in numerous industrial, H2020, and Horizon Europe projects. He has received prestigious recognitions, including the Amazon AWS Machine Learning Award and the Somalvico Award for the best young AI researcher in Italy. His primary research interests lie in Statistical Learning Theory and Trustworthy AI. Additionally, he focuses on data science, utilizing and improving cutting-edge machine learning and AI algorithms to tackle real-world problems.
\end{IEEEbiography}

\end{document}